%% file: main.tex
\documentclass[11pt]{chili}

\usepackage{multirow}
\usepackage{makecell}
\usepackage{array}
\usepackage{wrapfig}
\usepackage{float}
\usepackage{subcaption}
\usepackage[section]{placeins}   
\input{math_commands.tex}         

\input{qiushi_custom}

\usepackage[nameinlink,capitalise]{cleveref}
\crefname{section}{\S\hspace{-2.5pt}}{\S\S\hspace{-2.5pt}}
\Crefname{section}{\S\hspace{-2.5pt}}{\S\S\hspace{-2.5pt}}
\crefname{subsection}{\S\hspace{-2.5pt}}{\S\S\hspace{-2.5pt}}
\Crefname{subsection}{\S\hspace{-2.5pt}}{\S\S\hspace{-2.5pt}}
\crefname{subsubsection}{\S\hspace{-2.5pt}}{\S\S\hspace{-2.5pt}}
\Crefname{subsubsection}{\S\hspace{-2.5pt}}{\S\S\hspace{-2.5pt}}
\crefname{appendix}{\S\hspace{-2.5pt}}{\S\S\hspace{-2.5pt}}
\Crefname{appendix}{\S\hspace{-2.5pt}}{\S\S\hspace{-2.5pt}}
\crefname{figure}{Fig.}{Figs.}
\Crefname{figure}{Fig.}{Figs.}
\crefname{table}{Tab.}{Tabs.}
\Crefname{table}{Tab.}{Tabs.}

\title{%
  \begin{minipage}[c]{20mm}%
    \centering\raisebox{-2mm}{\includegraphics[height=12mm]{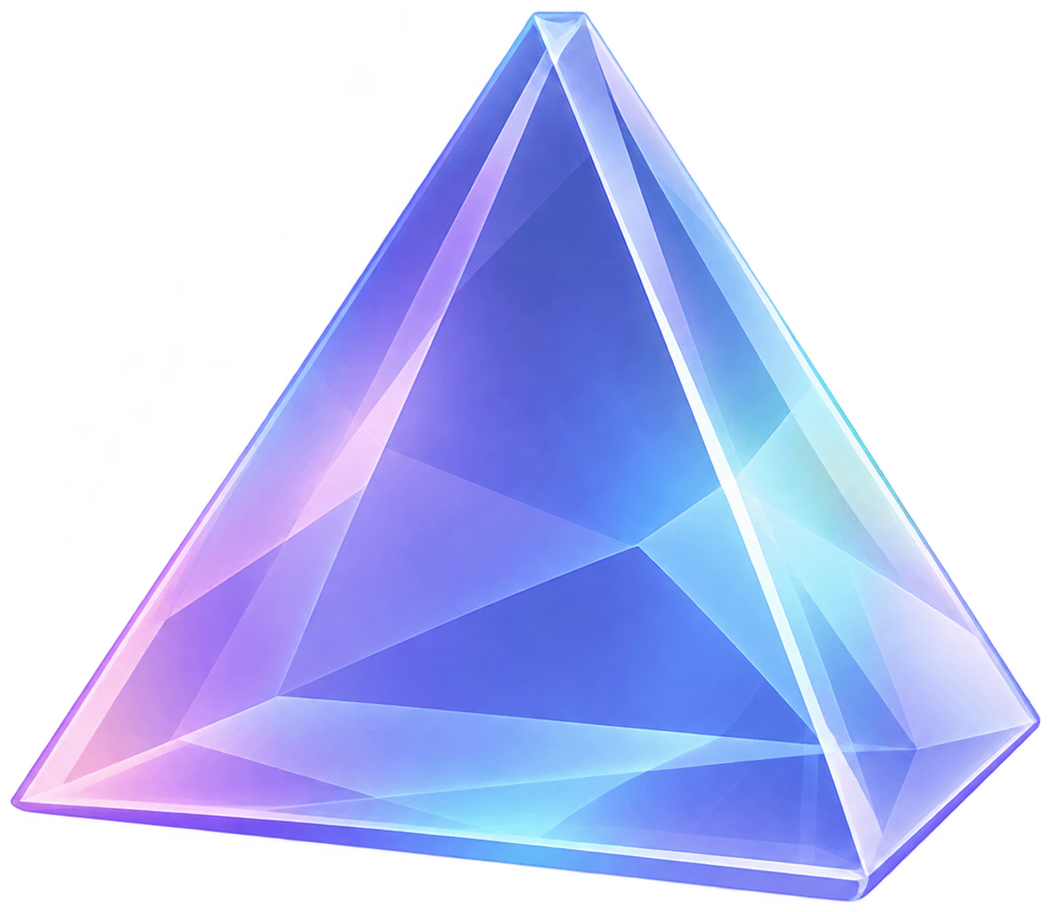}}%
  \end{minipage}\hspace{2mm}%
  \begin{minipage}[c]{\dimexpr\linewidth-22.5mm\relax}%
    \raggedright PrismaCoder: Pushing the Boundaries of Open Multimodal\\Code Intelligence through Decomposed Rubric Synthesis%
  \end{minipage}%
}

\ifanonauthors
\author{PrismaCoder Contributors}
\else
\makeatletter
\renewcommand\@author{%
  \begin{minipage}{\textwidth}
    {\centering\Authfont
      \mbox{Qiushi Sun\textsuperscript{1\equalmark}}, \mbox{Jingyang Gong\textsuperscript{1\equalmark}}, \mbox{Yang Liu\textsuperscript{2\equalmark}}, \mbox{Qiaosheng Chen\textsuperscript{2\equalmark}}, \mbox{Zhangyue Yin\textsuperscript{3\equalmark}},\\[1pt]
      \mbox{Zixian Huang\textsuperscript{2}}, \mbox{Hanxu Hu\textsuperscript{5}}, \mbox{Jing Xiong\textsuperscript{1}}, \mbox{Fangzhi Xu}, \mbox{Haochen Tan\textsuperscript{4\corrmark}}, \mbox{Haoli Bai\textsuperscript{4}},\\[1pt]
      \mbox{Lifeng Shang\textsuperscript{4}}, \mbox{Xiaoguang Li\textsuperscript{4}}, \mbox{Ben Kao\textsuperscript{1}}, \mbox{Qipeng Guo\textsuperscript{3,6}}, \mbox{Lingpeng Kong\textsuperscript{1}} and \mbox{Jiahui Gao\textsuperscript{1\corrmark}}\par}
    \vspace{3pt}
    {\centering\Affilfont
      \mbox{\textsuperscript{1}The University of Hong Kong}\hspace{1em}\mbox{\textsuperscript{2}Nanjing University}\hspace{1em}\mbox{\textsuperscript{3}Fudan University}\\[1pt]
      \mbox{\textsuperscript{4}Huawei Foundation Model Department}\hspace{1em}\mbox{\textsuperscript{5}University of Zurich}\hspace{1em}\mbox{\textsuperscript{6}Shanghai Innovation Institute}\par}
  \end{minipage}}
\makeatother

\correspondingauthor={qiushisun@connect.hku.hk, jingyang.gong@connect.hku.hk, ggaojiahui@gmail.com \\ \authorlegend}
\fi

\begin{document}
\maketitle

\input{Sections/0-abstract}
\input{Sections/1-intro}
\input{Sections/2-related}
\input{Sections/3-method}
\input{Sections/4-experiments}
\input{Sections/5-analysis}
\input{Sections/6-conclusion}
\input{Appendices/acknowledgments}

\bibliographystyle{iclr2027_conference}   
\bibliography{must_cite}

\clearpage
\appendix
\appendixtoc
\input{Appendices/appendix_content}

\end{document}

%% file: math_commands.tex
\usepackage{amsmath,amsfonts,bm}

\def\eqref#1{equation~\ref{#1}}

\def\1{\bm{1}}

\DeclareMathAlphabet{\mathsfit}{\encodingdefault}{\sfdefault}{m}{sl}
\SetMathAlphabet{\mathsfit}{bold}{\encodingdefault}{\sfdefault}{bx}{n}



%% file: qiushi_custom.tex
\newif\ifshowabstracttitle  \showabstracttitletrue  
\newif\ifshowlogo            \showlogofalse          
\newif\ifnumericcite         \numericcitefalse       
\newif\ifappendixtoc         \appendixtoctrue        
\newif\ifanonauthors         \anonauthorsfalse       

\usepackage{xspace}         
\usepackage{etoolbox}       
\usepackage{fontawesome5}   

\ifnumericcite
  \usepackage[numbers,square,sort&compress]{natbib}  
\else
  \usepackage[round,authoryear]{natbib}              
  \makeatletter
  \renewcommand\NAT@biblabel[1]{[#1]}
  \let\NAT@bibsetup\NAT@bibsetnum
  \makeatother
\fi

\usepackage[most]{tcolorbox}   

\definecolor{deepred}{HTML}{A51C30}         
\definecolor{AbstractRed}{HTML}{FBEDEF}     
\definecolor{boxredlight}{HTML}{FBE4E7}     
\definecolor{darkblue}{rgb}{0, 0, 0.5}
\definecolor{veronica-red}{RGB}{196,30,58}  
\definecolor{ForestGreen}{RGB}{34,139,34}
\definecolor{BrickRed}{rgb}{.72,0,0}
\definecolor{LakeBlue}{RGB}{0,61,153}
\definecolor{ultramarineblue}{RGB}{18,10,143}   
\definecolor{todored}{RGB}{222,30,30}           
\definecolor{boxgreen}{rgb}{0.9,1.0,0.9}
\definecolor{boxblue}{rgb}{0.9,0.9,1.0}
\definecolor{boxred}{rgb}{1.0,0.85,0.95}
\colorlet{headingred}{veronica-red!90!black}

\colorlet{BrandPrimary}{deepred}            
\colorlet{BrandRule}{deepred}               
\colorlet{BrandLink}{headingred}            
\hypersetup{linkcolor=headingred, urlcolor=headingred, citecolor=headingred}

\colorlet{MorandiAbstract}{AbstractRed}

\newcommand{\abstractheadingfont}{\normalfont\bfseries\fontsize{14}{16}\selectfont\color{headingred}}
\makeatletter
\pretocmd{\archer@renderabstract}{%
  \ifshowabstracttitle
    {\centering\abstractheadingfont\abstractname\par}\vspace{4pt}%
  \fi
}{}{}
\makeatother

\newcommand{\qsheadingfont}{\color{headingred}\bfseries\fontsize{13}{14}\selectfont}
\titleformat{\section}
  {\large\bfseries\qsheadingfont}{\thesection.}{0.5em}{#1}[]
\titleformat{name=\section,numberless}
  {\large\bfseries\qsheadingfont}{}{0em}{#1}[]
\titleformat{\subsection}
  {\color{headingred}\bfseries}{\thesubsection.}{0.5em}{#1}[]
\titleformat{\subsubsection}
  {\color{headingred}\bfseries\itshape}{\thesubsubsection.}{0.5em}{#1}[]
\titleformat{\paragraph}[runin]
  {\bfseries\color{headingred}}{\theparagraph.}{0.5em}{#1}
\titlespacing*{\paragraph}{0pt}{1.0ex plus 0.5ex minus .2ex}{1em}

\newcommand{\logoblock}{%
  \fcolorbox{BrandRule}{BrandLight}{\rule[-0.45em]{0pt}{1.5em}\scriptsize\itshape\,logo 1\,}\hspace{4pt}%
  \fcolorbox{BrandRule}{BrandLight}{\rule[-0.45em]{0pt}{1.5em}\scriptsize\itshape\,logo 2\,}%
}
\makeatletter
\let\ps@firststyleorig\ps@firststyle
\@namedef{f@nch@ps@firststyleorig-is-fancyhdr}{}
\makeatother
\fancypagestyle{firststyle}[firststyleorig]{\fancyhead[L]{\ifshowlogo\logoblock\fi}}

\renewcommand{\titlefont}{\color{BrandPrimary}\normalfont\bfseries\fontsize{15}{19}\selectfont}

\makeatletter
\renewcommand\AB@affilsepx{\protect\hspace{1em}\protect\Affilfont}
\patchcmd{\archer@maketitle@onecolumn}{\begin{adjustwidth}{0pt}{24pt}}{\begin{adjustwidth}{0pt}{0pt}}{}%
  {\PackageWarning{qiushi}{title width patch failed; width unchanged}}
\patchcmd{\archer@maketitle@onecolumn}{\vskip11pt}{\vskip2pt}{}%
  {\PackageWarning{qiushi}{title bottom-skip patch failed; spacing unchanged}}
\makeatother

\newcommand{\runningtitle}{PrismaCoder: Open Multimodal Code Intelligence through Decomposed Rubric Synthesis}
\hypersetup{pdftitle={PrismaCoder: Pushing the Boundaries of Open Multimodal Code Intelligence through Decomposed Rubric Synthesis}}

\newcommand{\equalmark}{\ensuremath{\ast}}        
\newcommand{\corrmark}{\,\faEnvelope[regular]}    

\newcommand{\authorlegend}{\upshape\footnotesize
  \textsuperscript{\ensuremath{\ast}}\,Equal contribution\quad
  \textsuperscript{\faEnvelope[regular]}\,Corresponding author}

\newtcolorbox{notebox}[1][]{%
  enhanced, breakable,
  colback=boxredlight, colframe=veronica-red, boxrule=0.5pt, arc=2pt,
  left=6pt, right=6pt, top=4pt, bottom=4pt,
  fonttitle=\bfseries, coltitle=white,
  #1
}

\makeatletter
\newcommand{\appendixtocname}{Appendix Contents}
\newcommand{\appendixtoc}{%
  \ifappendixtoc
    \section*{\appendixtocname}%
    \@starttoc{apc}%
    \let\qs@oldaddcontentsline\addcontentsline
    \renewcommand\addcontentsline[3]{%
      \def\qs@tmp{##1}\def\qs@toc{toc}%
      \ifx\qs@tmp\qs@toc \qs@oldaddcontentsline{apc}{##2}{##3}\fi}%
  \fi
}
\makeatother

\newcommand{\cmark}{\textcolor{ForestGreen}{\ding{51}}}
\newcommand{\xmark}{\textcolor{BrickRed}{\ding{55}}}

\definecolor{pievision}{HTML}{DA8A9C}
\definecolor{pietext}{HTML}{8FA8C8}
\newcommand{\swatch}[1]{\textcolor{#1}{\rule{0.85em}{0.85em}}\,}

\newcommand{\methodname}{\textsc{PrismaCoder}\xspace}
\newcommand{\pipename}{\textsc{PrismaForge}\xspace}
\newcommand{\modelnine}{\textsc{PrismaCoder}-9B\xspace}
\newcommand{\modelthirtyfive}{\textsc{PrismaCoder}-35B\xspace}
\newcommand{\modelthirtyfiveA}{\textsc{PrismaCoder}-35B-A3B\xspace}
\newcommand{\ourdata}{\textsc{PrismaCoder}-2M\xspace}

\newcommand{\geminipro}{\texttt{Gemini-3-Pro}\xspace}
\newcommand{\claudeopus}{\texttt{Claude-Opus-4.6}\xspace}
\newcommand{\gptfouro}{\texttt{GPT-4o}\xspace}

\newcommand{\gemmaicon}{\raisebox{-0.15em}{\includegraphics[height=0.9em]{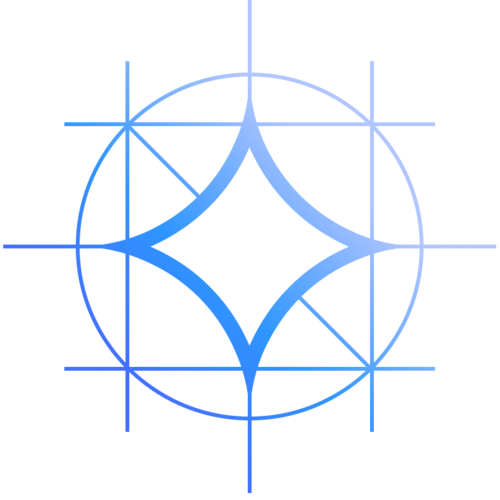}}\,}

\newcommand{\casepanel}[4][\footnotesize]{\begin{minipage}[t]{#4}\centering
\includegraphics[width=\linewidth]{Figures/case_studies/#2}\\[2pt]
{#1 #3}\end{minipage}}
\newcommand{\casepanelh}[5][\footnotesize]{\begin{minipage}[t]{#4}\centering
\includegraphics[width=\linewidth,height=#5,keepaspectratio]{Figures/case_studies/#2}\\[2pt]
{#1 #3}\end{minipage}}

\newcommand{\casetriptych}[5]{%
\begin{figure}[htbp]
\centering
\begin{minipage}[t]{0.32\linewidth}\centering
\includegraphics[width=\linewidth]{Figures/case_studies/#1_gold.png}\\[2pt]
{\footnotesize Gold}
\end{minipage}\hfill
\begin{minipage}[t]{0.32\linewidth}\centering
\includegraphics[width=\linewidth]{Figures/case_studies/#1_baseline.png}\\[2pt]
{\footnotesize #2}
\end{minipage}\hfill
\begin{minipage}[t]{0.32\linewidth}\centering
\includegraphics[width=\linewidth]{Figures/case_studies/#1_ours.png}\\[2pt]
{\footnotesize #3}
\end{minipage}
\caption{#4}
\label{#5}
\end{figure}}

\definecolor{figCurLine}{HTML}{6D8ABE}
\definecolor{figCurFill}{HTML}{EEF3FA}
\definecolor{figRlLine}{HTML}{D78C6E}
\definecolor{figRlFill}{HTML}{FBF2EC}
\definecolor{figRubLine}{HTML}{8E6AA8}
\definecolor{figRubFill}{HTML}{F5F0F7}
\definecolor{figNodeLine}{HTML}{B5B9C4}
\definecolor{figDataLine}{HTML}{4E9C87}
\definecolor{figDataFill}{HTML}{DDF3EC}
\definecolor{figCodeLine}{HTML}{9DB4DA}
\definecolor{figCodeFill}{HTML}{DCE8F8}
\definecolor{figVisLine}{HTML}{8FC9BB}
\definecolor{figVisFill}{HTML}{D6EFE8}

%% file: Sections/0-abstract.tex
\begin{abstract}
Multimodal code intelligence connects the symbolic world of programs with the
visual world their execution creates. Interest in this capability is
burgeoning, as models are increasingly asked to produce charts, webpages,
animations, and semi-structured documents. Recently, proprietary models have
pushed far ahead, producing interactive artifacts even from a text description
or a screenshot alone. Yet their data and training remain closed, and
open progress is held back by the quantity and quality of available
data and by the pipelines and methods for producing it. Quality is the harder
of the two, since a sample must be correct as a program, well built, and above
all visually right once rendered. No existing signal covers all three.
Execution often returns just a pass or fail, code similarity does not map to the
visual output, and task-specific verifiers do not generalize. To push open
progress further, we start from the data side with \pipename{}, a hierarchical
data pipeline centered on a synthesized rubric that decomposes these factors
into checkable criteria. The same rubric serves as both
filter and reward: it curates the supervised corpus, then drives
reinforcement learning on visual-coding tasks. Leveraging \pipename{} for data
synthesis and filtering, we build \ourdata{}, the most diverse open dataset for
multimodal code intelligence and the largest to date. On the model side, we train \modelnine{} and
\modelthirtyfive{} on this data and release both. They lead the open state of
the art by a wide margin and rival leading proprietary systems. 
Our analyses further show that the proposed rubric-based pipeline improves multimodal coding performance across model backbones.
We release our code, model, and data at the
\href{https://qiushisun.github.io/PrismaCoder-Home/}{PrismaCoder Homepage}.
\end{abstract}

%% file: Sections/1-intro.tex
\section{Introduction}
\label{sec:intro}

\input{Figures/task_overview}

From symbols to pixels, code intelligence now extends beyond textual source code
into the visual world that programs produce~\citep{zan2023large,sun2024survey, zhao2026beyond}. A growing range of applications
specifies intent through images and judges a result by what its code renders, from charts and web pages to
diagrams, typeset documents, and animations. Leading
proprietary systems now produce interactive artifacts from a text description or
a screenshot alone.

Prior open efforts have mostly addressed one corner of this space at a time, from
plotting~\citep{ni2025viscoder, pandasplotbench2025timur, yang2024matplotagent}
and chart-to-code~\citep{yang2025chartmimic, zhao2025chartcoder} to
screenshot-to-page generation~\citep{gui2025webcode2m, si2025design2code}. Each
comes with its own data and recipe,
making it difficult for the resulting models to generalize across task domains and output formats.
Recently, open work such as
JanusCoder~\citep{sun2026januscoder} has brought many of these scenarios under a
shared visual-programmatic interface, with follow-up refinements building on the
same view~\citep{zhao2025vincicoderunifyingmultimodalcode, jiang2026viscodex, ni2026viscoder}.
Despite these advances, publicly available training data remains limited in scale and uneven in quality, with only a handful of open releases and resources to date~\citep{sun2026januscoder, jiang2026viscodex}.
The pipelines used to build these datasets are rarely released, while proprietary systems keep their data and training closed. 
Moreover, existing public data are typically curated with either a single per-sample score, which provides limited diagnostic information about errors, or a task-specific verifier, which does not generalize across tasks.

To address this gap, we introduce \pipename{}, a rubric-based framework for multimodal code data curation and policy optimization. Rather than relying solely on a holistic judgment or a fixed checklist, \pipename{} applies a common evaluation procedure across data sources and task domains while tailoring its criteria to each instruction. Each rubric combines reusable domain-level checks with instruction-specific requirements, with the number of criteria adjusted to the estimated task difficulty. Code-level and visual-level criteria are evaluated separately on the generated program and its rendered artifact, respectively. 
These scores guide data curation and serve as reinforcement learning rewards.

Using \pipename{}, we curate \ourdata{},  the broadest open dataset to date for advancing multimodal code intelligence, spanning more than 12 categories and
10 programming languages with both text-centric and vision-centric data. To build this coverage, we draw seeds from complementary sources: public visual-coding corpora cover common plotting and graphics tasks; research figures and open-source repositories contribute long-tail scientific examples; and web, interactive-demo, and chemical-structure sources broaden the range of artifacts. We then convert the seeds into self-contained tasks, check candidate programs through format-specific execution and rendering, and use instruction-specific rubric scores to filter or refine the resulting examples. This gives the dataset both breadth of coverage and a consistent quality-control process.

To assess whether the breadth of \ourdata{} translates into broad model capability, we develop \modelnine{} and \modelthirtyfive{}.  
Evaluated on a wide range of representative benchmarks, \methodname{} leads the existing open state of the art by a wide margin and matches the proprietary systems to a considerable degree (\cref{sec:experiments}).
Through extensive analyses (\cref{sec:analysis}), we show that the data and pipeline generalize across base models beyond our release. We further show that rubric-guided curation outperforms direct-judge selection and that rubric scores align closely with human judgments.

Our main contributions are summarized as follows:
\begin{itemize}[leftmargin=2em, itemsep=2pt, topsep=0pt, parsep=0pt]
    \item \textbf{Decomposed Rubric Synthesis for Visual-coding Data.} We introduce
    \pipename{}, a principled pipeline whose hierarchically synthesized rubrics
    serve as both filter and reward. Based on it, we curate \ourdata{}, the broadest and largest open
    dataset to date.
    \item \textbf{Open Models that Rival Proprietary Systems.} We build \modelnine{}/35B on
    \ourdata{}, which lead the open state of the art by a wide margin and match or surpass
    leading proprietary systems on charts, front-end pages, and semi-structured documents.
    \item \textbf{Empirical Insights across Data, Modeling, and Training.}
    Beyond data and model contributions, we further share insights through extensive analysis. It shows that our pipeline generalizes across scenarios and backbones and yields artifacts better aligned with human preference.
\end{itemize}

%% file: Figures/task_overview.tex
\begin{figure}[t]
\centering
\includegraphics[width=0.95\linewidth]{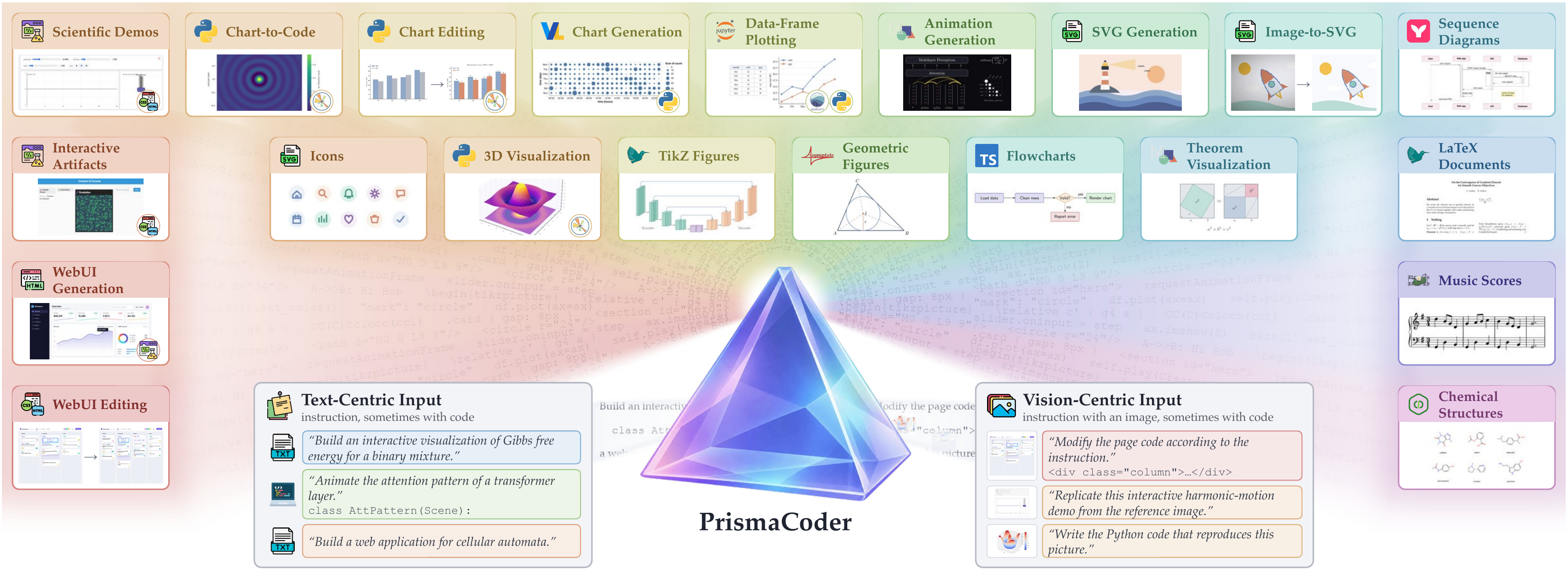}
\caption{Scenarios and formats covered by \ourdata{} and \methodname{}.}
\label{fig:task-overview}
\vspace{-7pt}  
\end{figure}

%% file: Sections/2-related.tex
\section{Related Work}
\label{sec:related}

\paragraph{Multimodal Code Intelligence.}
Code intelligence now extends beyond text-only programming to visual-programmatic tasks, where intent may be visual~\citep{li2024mmcode, zhang2024humanevalv, wang2025codevision} and outputs can be judged only after execution or rendering~\citep{zhao2026beyond}.
On the output side, syntactic validity is insufficient; correctness also depends on layout, semantics, interaction, and editability.
On the input side, reading values and layout off renders remains a perception bottleneck~\citep{masry2022chartqa, wang2024charxiv, xia2025chartx} across vision-language models~\citep{zhang2024tinychart, liu2025perceptionbottleneckvlmschart}.
The tasks run from screenshot-to-code~\citep{gui2025webcode2m, si2025design2code, yun2024webcode} to front-end engineering~\citep{ScreenCoder, zhu2025frontendbench, sun2025fullfront} and generative UI~\citep{zheng2026code2world}; chart generation~\citep{yang2025chartmimic, zhao2025chartcoder, wu2025plot2code} and editing~\citep{zhang2024academicvis, zhao2025chartedit}; SVG and symbolic graphics~\citep{yang2025omnisvg, nishina2024svgeditbench, xu2024symbol} with their structural probes~\citep{qiu2025can, chen2025symbolicgraphics}; and interactive artifacts and code-driven demonstrations~\citep{zhang2025artifactsbench, ku2025theoremexplainagent, sun2025osgenesis}.
Recent generalists unify these domains via JanusCoder's visual-programmatic interface~\citep{sun2026januscoder}, VisCodex's vision--code integration~\citep{jiang2026viscodex}, VisCoder2's agentic visualization coding~\citep{ni2026viscoder}, and VinciCoder's visual reinforcement learning~\citep{zhao2025vincicoderunifyingmultimodalcode}.
Yet a shared modeling interface does not by itself make heterogeneous notions of correctness consistently measurable or usable for training~\citep{zhao2026beyond}.
Complementing this progress, we contribute on the data side, with \ourdata{} spanning the broadest scenarios and open data types.

\paragraph{Learning Signals for Code Generation.}
Conventional code tasks~\citep{zhang2023repocoder} draw training data from instruction synthesis and evolution~\citep{wei2024magicoder,luo2024wizard,luo2025tree}.
For many such tasks, executable tests provide a cheap, reproducible signal for data filtering~\citep{xu2025kodcode, sun2026codeevo} and reinforcement learning~\citep{shao2024deepseekmath, xu2025interactive, zhuo2025bigcodearena}.
Visual outputs break this assumption: successful execution confirms only that the code runs, not that its render is faithful, well composed, or semantically correct~\citep{zhao2026beyond, liu2026visual, guo2025mammothvl}.
Visual benchmarks instead use task-specific render checks for charts~\citep{yang2025chartmimic}, visual and structural similarity for web pages~\citep{si2025design2code}, or judge protocols for interactive artifacts~\citep{zhang2025artifactsbench}; effective individually, these measures do not compose across artifacts.
Recent work moves toward finer supervision, from JanusCoder's single-pass rubric over fixed dimensions~\citep{sun2026januscoder} to VinciCoder's coarse-to-fine rewards~\citep{zhao2025vincicoderunifyingmultimodalcode}. Neither signal is yet both fine enough to localize failures and general enough to cover every task, at curation and at training alike. In this work, one synthesized rubric per instruction serves both roles, filtering the training data and rewarding reinforcement learning across every format we cover.

%% file: Sections/3-method.tex
\section{PrismaForge}
\label{sec:method}
In this section, we present \pipename{}, a rubric-based scoring framework that lies in the center of our data curation and training pipeline.

\input{Figures/pipeline}

\subsection{Elastic Rubric Scorer}
\label{subsec:rubric}

For a large-scale coding dataset prepared for visualization or demonstration, the key challenge is to determine which candidates are sufficiently correct and useful to retain. Unlike problems with easily verifiable outputs, visualization quality is often specified only implicitly~\citep{wu2025plot2code, yang2025chartmimic}. A single holistic score is therefore hard to verify and offers little insight into why a candidate succeeds or fails~\citep{ye2024iclrw-flask, sun2026januscoder}.

A natural alternative is to evaluate candidates using explicit criteria, but a fixed rubric is too coarse for diverse visualization tasks. Different instructions may prioritize numerical accuracy and axis scales, visual similarity, layout, or artifact editability. A universal rubric must stay broad and may miss the properties most important for a specific example. At the other extreme, manually designing separate rubrics for every visualization instruction would require substantial human effort and is hard to scale.

To balance these two extremes, we introduce a sample-specific rubric system that combines reusable domain-level criteria with criteria synthesized specifically for each instruction.

Formally, let $i \in \{1,\ldots,N\}$ index dataset instances, and let $d_i \in \mathcal{D}$ denote the visualization domain associated with instance $i$. We represent candidate sample $i$ as a tuple $(x_i, c_i, v_i)$ where $x_i$ denotes the instruction, which may contain text and optional visual references; $c_i$ denotes the response code produced for that instruction; and $v_i = \operatorname{Exec}(c_i)$ denotes the visual artifact obtained by executing $c_i$. 

We evaluate each candidate using two complementary sets of rubric criteria: code rubrics $\mathcal{R}_i^{(c)}$, which assess properties that can be verified from $c_i$, and visual rubrics $\mathcal{R}_i^{(v)}$, which assess properties directly observable from $v_i$.

\input{Figures/rubric_example}

\Cref{fig:rubric-example} shows why the two views are kept apart. The synthesized page
carries the reference's content, colors, and layout in its code, so every code check passes.
Its footer, however, is drawn over the Services section and clips the third button. A holistic
judge still rates the page 78 out of 100, above about four fifths of the pool, whereas the
vision rubric scores it 0.18. The checks and judgments behind both scores are listed in
Appendix~\cref{app:prompts}.

Within each modality, we further distinguish between domain-shared and sample-specific criteria. Domain-shared rubrics encode requirements that recur across samples within the same visualization domain, whereas sample-specific rubrics are derived from the instruction $x_i$ and capture requirements unique to the current sample. Thus, for modality $m \in \{c,v\}$,
\begin{equation}
\label{eq:rubric-decomposition}
\mathcal{R}_i^{(m)}
=
\mathcal{R}_{\mathrm{shared}}^{(m)}(d_i)
\cup
\mathcal{R}_{\mathrm{spec}}^{(m)}(x_i).
\end{equation}

Because individual criteria may differ in importance, each rubric item $r \in \mathcal{R}_i^{(m)}$ is assigned a nonnegative, sample-dependent weight $w_{i,r}^{(m)}$.

To score a candidate,
let $f^{(m)}(r,y) \in [0,1]$ denote the judge score indicating how well artifact $y$ satisfies rubric criterion $r$ under modality $m$. The normalized score for modality $m$ is then
\begin{equation}
\label{eq:modality-score}
S_i^{(m)}
=
\frac{1}{
\sum_{r \in \mathcal{R}_i^{(m)}}
w_{i,r}^{(m)}
} \sum_{r \in \mathcal{R}_i^{(m)}}
w_{i,r}^{(m)}
f^{(m)}(r,y_i^{(m)})
\end{equation}

The two modality scores are then combined into an overall sample score with appropriate weighting
\begin{equation}
\label{eq:overall-score}
S_i
=
\lambda_c S_i^{(c)}
+
\lambda_v S_i^{(v)},
\end{equation}
where $\lambda_c,\lambda_v \geq 0$ and $\lambda_c+\lambda_v=1$.

In practice, candidate samples vary substantially in difficulty and complexity, and therefore should not be evaluated with the same number of rubric items. We first assign each instruction $x_i$ a difficulty level $q_i \in \mathcal{Q}$ and use this level to guide rubric generation. Each difficulty band is associated with a bounded range for the number of generated criteria. This allows more complex samples to receive finer-grained evaluation while keeping the rubric compact enough to control judging cost.

\subsection{Rubric-Based Policy Optimization}
\label{subsec:rl}

The same rubric framework can also be used to construct rewards for reinforcement learning. Because rubric scores are normalized within each modality and aggregated using a common scoring rule, they provide a consistent reward scale across samples with different numbers and types of criteria. We therefore use the rubric-based evaluator as a structured reward function during policy optimization.

For each prompt $x_i$, we sample $K$ rollout responses from the behavior policy
\begin{equation}
{c_i^{(1)},\cdots,c_i^{(K)}}
\sim
\pi_{\theta_{\mathrm{old}}}(\cdot \mid x_i).
\end{equation}
Each rollout $c_i^{(k)}$ is executed to obtain the visual artifacts $v_i^{(k)}=\textrm{Exec}\left(c_i^{(k)}\right)$. Rollouts that failed to execute or render receive reward $0$.

The judge model then scores these rollouts. For each successful rollout trajectory $(x_i, c_i^{(k)}, v_i^{(k)})$, we apply its accompany rubrics set $(\mathcal{R}_i^{(c)}, \mathcal{R}_i^{(v)})$ and obtain the aggregated score $s_i^{(k)}$ via Eq.~\ref{eq:modality-score}--Eq.\ref{eq:overall-score}. These scores are then leveraged to compute the within-group normalized advantage
\begin{equation}
\hat{A}_i^{(k)}
=
\frac{
s_i^{(k)}-\mu_i
}{
\sigma_i+\epsilon
}
\end{equation}
where $\mu_i=\frac{1}{K} \sum_{k=1}^{K} s_i^{(k)}$ and $\sigma_i$ denotes the standard deviation of ${s_i^{(k)}}$.

With this reward formulation, the policy can then be optimized with the standard GRPO
clipped objective~\citep{shao2024deepseekmath}. In practice, we also used GSPO objective~\citep{gspo} with Rollout Routing Replay techniques~\citep{ma2025stabilizing} for the MoE models to stabilize experts training.

\section{PrismaCoder-2M}
\label{sec:prisma-data}
\input{Figures/data_pie}

We build \ourdata{} with \pipename{} by collecting seeds from complementary sources, gating
them by rubric score into a supervised corpus, and holding out a subset for reinforcement
learning, where the same rubrics supply the reward.

\subsection{Data Sourcing}
\label{subsec:data-gather-gate}
Multimodal code generation needs several capabilities at once, so we collect four kinds of
seeds that each cover one of them and train on them together (\cref{tab:data-stats,fig:data-pie}).

\paragraph{Visualization Fundamentals.} Visual coding starts from a core of plotting skills,
such as mapping data to axes and marks and composing legends and layered plots. Public plotting
datasets build this foundation with instruction--code and visualization--code pairs in
Python~\citep{hunter2007matplotlib} and in SVG, Mermaid, Vega-Lite, and other visualization
languages~\citep{ni2025viscoder, ni2026viscoder, jiang2026viscodex}.

\paragraph{Diverse Scientific Artifacts.} Scientific work produces more varied visuals than
standard charts, from multi-panel figures to domain-specific encodings. Plotting code from
open-source GitHub repositories and figures from arXiv papers bring this variety into the data,
and an AST-based pass splits long or monolithic scripts into self-contained units.

\paragraph{Front-End Layout and Interaction.} Front-end development asks a model to arrange
many elements on a page and to make them respond to users. Web pages from
WebCode2M~\citep{gui2025webcode2m} and the MCD dataset~\citep{jiang2026viscodex} supply the
layout, while interactive demos supply the behavior. The demos are adapted from prior
releases~\citep{sun2026januscoder} or rewritten from static pages~\citep{laurenccon2024websight}
and complex generation prompts~\citep{luo2026mmmg, ni2025proimage} into tasks of varying
visual, functional, and reasoning complexity.

\paragraph{Symbolic Notation.} Science, mathematics, and music communicate through symbolic
notation, where a single wrong symbol breaks the result. Chemical-structure code seeded from
PubChem~\citep{kim2025pubchem} must describe a valid molecule, and \LaTeX{}, Asymptote, LilyPond, and Manim programs must compile into well-formed
documents, figures, scores, and scenes. The first three formats draw on the same sources as the
plots, and the Manim seeds come from the 3Blue1Brown video code.
See Appendix~\cref{app:sources}.

\subsection{Iterative Data Gating}
\label{subsec:data-gating}

Every seed enters \pipename{} as a candidate $(x_i, c_i, v_i)$. Where a seed lacks working code, a curation model $\mathcal{L}^{C}$ generates it, $c_i \sim \mathcal{L}^{C}(\cdot \mid x_i)$, through a code-data synthesis pipeline. \pipename{} works with any such pipeline, and we use a typical open-source one~\citep{sun2026januscoder}. Each candidate must then compile or render through a per-format gatekeeper, $v_i = \operatorname{Exec}(c_i)$, and candidates that still fail after regeneration are set aside in a repair pool.

A rubric generator $\mathcal{L}^{R}$ assigns each candidate its difficulty level and its code and visual rubrics (\cref{subsec:rubric}), and a separate judge $\mathcal{L}^{J}$ scores the code against $\mathcal{R}_i^{(c)}$ and the render against $\mathcal{R}_i^{(v)}$. The modality scores and the overall score $S_i$ follow Eqs.~\ref{eq:modality-score} and~\ref{eq:overall-score}. We use distinct models for rubric generation and judging so that biases in rubric construction are not reproduced during scoring.

We then gate candidates by their overall scores. Those with $S_i < \tau_1$ are discarded, those with $\tau_1 \leq S_i < \tau_2$ are sent back for re-synthesis, and those with $S_i \geq \tau_2$ are retained in the curated pool.

The synthesis pipeline serves this gate in two ways. It re-rolls the code of a mid-band candidate with the failed rubric items as feedback, and it synthesizes new samples seeded by retained candidates. Both kinds of output re-enter the pipeline at the execution gate and are scored and routed like any other candidate. For the final supervised corpus, we retain a sample only if its score meets the threshold for its domain, $S_i \geq \tau_{d_i}$.

\subsection{RL Data Curation}
\label{subsec:rl-data}
Reinforcement learning draws on a compact set of instructions held out from the supervised
pool, which SFT never sees. Among them we keep the medium or hard instructions on which the
SFT policy sometimes succeeds and sometimes fails. Such instructions sit near the policy's
capability boundary, and their rollouts differ in quality enough to give group-relative
optimization a contrast to learn from, whereas instructions that every rollout passes or
fails provide none.

Formally, let $\pi_{\mathrm{SFT}}$ denote the SFT-tuned policy. For each candidate instance $i$, we draw $K$ independent responses
$c_i^{(k)} \sim \pi_{\mathrm{SFT}}(\cdot \mid x_i)$ and evaluate each rollout using the rubric-based scoring function defined in \cref{subsec:rubric}.
We then measure the informativeness of instance $i$ by the empirical dispersion of its rubric
scores
\begin{equation}
\label{eq:rl-score-variance}
U_i=
\frac{1}{K}
\sum_{k=1}^{K}
\left(S_i^{(k)}-\bar S_i\right)^2,
\quad
\bar S_i
=\frac{1}{K}\sum_{k=1}^{K}S_i^{(k)}.
\end{equation}
A large $U_i$ means that the policy sometimes succeeds and sometimes fails on the same
instruction.

In practice, we run RL on all Python, HTML, and SVG tasks, and for each domain we hold out
20K \emph{medium} or \emph{hard} candidates before SFT. We then draw $K=5$ rollouts per instance from $\pi_{\mathrm{SFT}}$, score
each rollout against its code and visual rubrics, and retain the 5K instances with the
largest $U_i$. RL thus concentrates on the policy's capability frontier at a fraction
of the rollout cost of a larger candidate pool.

%% file: Figures/pipeline.tex
\begin{figure}[t]
\centering
\includegraphics[width=0.95\linewidth]{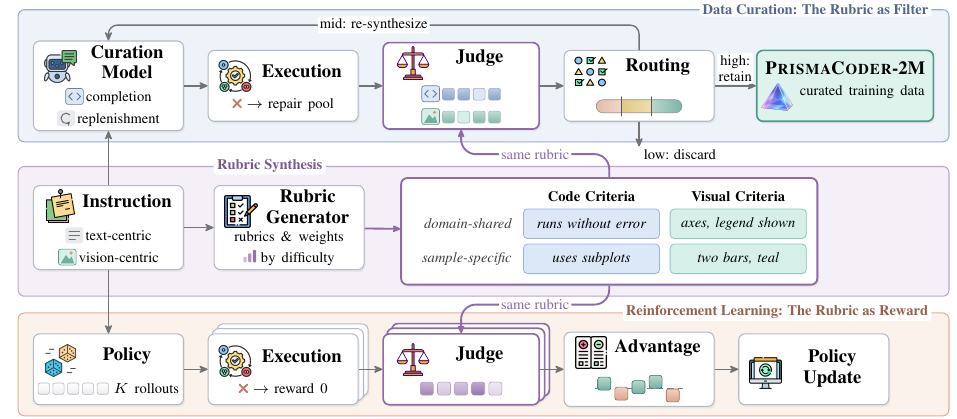}
\caption{\pipename{} writes one rubric per instruction and uses it twice, to filter
candidate data into \ourdata{} and to reward rollouts in reinforcement learning.}
\label{fig:pipeline}
\end{figure}

%% file: Figures/rubric_example.tex
\begin{figure}[t]
\centering
\setlength{\fboxsep}{0pt}\setlength{\fboxrule}{0.4pt}
\begin{minipage}[c]{0.36\linewidth}\centering
\fcolorbox{black!30}{white}{\includegraphics[height=1.36in]{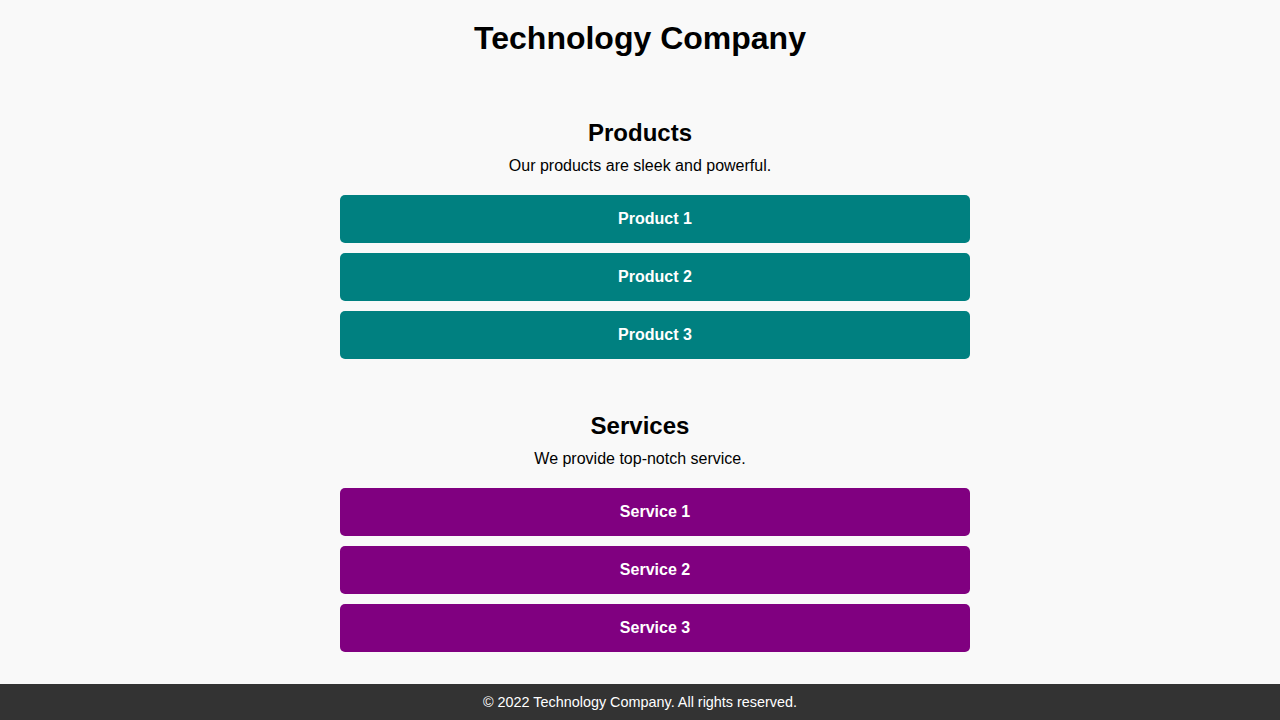}}\\[3pt]
{\footnotesize Reference}
\end{minipage}\hfill
\begin{minipage}[c]{0.31\linewidth}\centering
\fcolorbox{black!30}{white}{\includegraphics[height=1.36in]{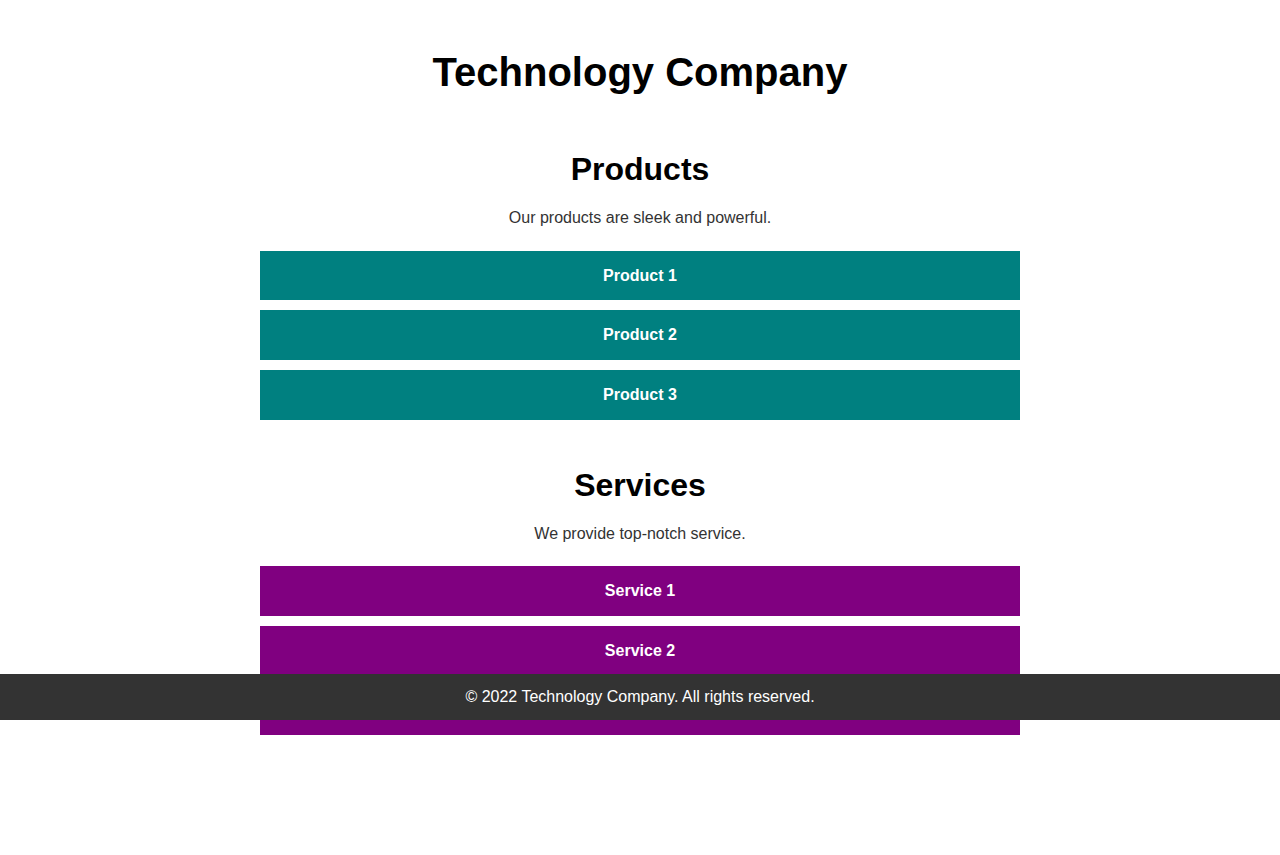}}\\[3pt]
{\footnotesize Synthesized}
\end{minipage}\hfill
\begin{minipage}[c]{0.29\linewidth}\centering
\footnotesize
\setlength{\tabcolsep}{4pt}
\renewcommand{\arraystretch}{1.3}
\begin{tabular}{@{}lcr@{}}
\toprule
Signal & Checks & Score \\
\midrule
Holistic judge & -- & 78\,/\,100 \\
\rowcolor{figCodeFill} Code rubric & 5\,/\,5 pass & 1.00 \\
\rowcolor{figVisFill} Vision rubric & 1\,/\,5 pass & 0.18 \\
\bottomrule
\end{tabular}
\end{minipage}
\caption{Scored separately, the vision rubric catches a misplaced footer that every code
check and a holistic judge let through.}
\label{fig:rubric-example}
\end{figure}

%% file: Figures/data_pie.tex
\begin{figure}[t]
\begin{minipage}[c]{0.47\linewidth}
\centering
\captionof{table}{Statistics of \ourdata{}.}
\label{tab:data-stats}
\footnotesize
\setlength{\tabcolsep}{5pt}
\renewcommand{\arraystretch}{0.9}
\begin{tabular}{l r r}
\toprule
Task family & \#Samples & Share \\
\midrule
\multicolumn{3}{l}{\swatch{pievision}\textit{Vision-centric}} \\
WebUI Generation      & 357.1K & 17.8\% \\
Chart-to-Code         & 196.5K &  9.8\% \\
Image-to-SVG          & 181.0K &  9.0\% \\
WebUI Editing         & 175.0K &  8.7\% \\
Demo / Artifacts      & 163.3K &  8.1\% \\
Chemical Structures   & 108.0K &  5.4\% \\
Interactive Artifacts &   9.1K &  0.5\% \\
\midrule
\multicolumn{3}{l}{\swatch{pietext}\textit{Text-centric}} \\
Chart Generation      & 181.1K &  9.0\% \\
Demo / Artifacts      & 163.0K &  8.1\% \\
Text-to-WebUI         & 135.2K &  6.7\% \\
\LaTeX{}              & 133.1K &  6.6\% \\
SVG Generation        &  53.3K &  2.7\% \\
Animation (Manim)     &  35.9K &  1.8\% \\
JavaScript            &  28.5K &  1.4\% \\
Mermaid               &  22.7K &  1.1\% \\
LilyPond              &  21.9K &  1.1\% \\
Vega-Lite             &  20.0K &  1.0\% \\
Asymptote             &  19.4K &  1.0\% \\
\midrule
Total                 & 2M & 100\% \\
\bottomrule
\end{tabular}
\end{minipage}\hfill
\begin{minipage}[c]{0.50\linewidth}
\centering
\includegraphics[width=\linewidth]{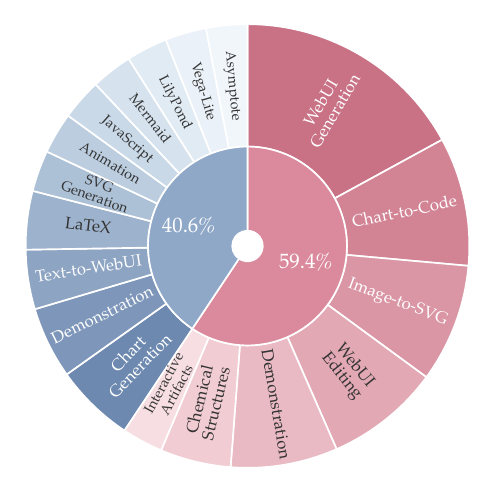}
\caption{Composition of \ourdata{} by input modality and task family.}
\label{fig:data-pie}
\end{minipage}
\vspace{-1em}
\end{figure}

%% file: Sections/4-experiments.tex
\section{Experiments}
\label{sec:experiments}

\subsection{Experimental Setup}
\label{subsec:settings}

\paragraph{Models.} We release \modelnine{} and \modelthirtyfiveA{} (abbreviated as \modelthirtyfive{}), post-trained from
Qwen3.5-9B and Qwen3.5-35B-A3B~\citep{qwen3_5}. We also train
Gemma-4-31B-it~\citep{google2026gemma} and Qwen3-VL-30B-A3B-Instruct~\citep{bai2025qwen3}.
More training details are in Appendix~\cref{app:training}.

\paragraph{Data Pipeline Details.} 
We reprocess data from prior open releases~\citep{sun2026januscoder, jiang2026viscodex, ni2026viscoder}. \pipename{} treats its
models as pluggable, and in this paper we use \claudeopus{}~\citep{anthropic2026opus46} as the
curation model and \geminipro{}~\citep{gemini2025gemini3} as the rubric generator. Both data gating
and reinforcement learning use Gemini 3 Flash as the judge.
The sources are listed per category in Appendix~\cref{app:sources}.

\paragraph{Baselines.} Besides Qwen3.5, we compare against open multimodal code models, namely JanusCoderV~\citep{sun2026januscoder}, VinciCoder~\citep{zhao2025vincicoderunifyingmultimodalcode}, and VisCoder2-7B~\citep{ni2026viscoder}.

\subsection{Benchmarking}
\label{subsec:benchmarks}

\paragraph{Vision-Centric Settings.} The vision-centric suite comprises
ChartMimic~\citep{yang2025chartmimic}, WebCode2M~\citep{gui2025webcode2m},
DesignBench~\citep{xiao2025designbench}, Design2Code~\citep{si2025design2code}, and
ChemDraw~\citep{zhao2025vincicoderunifyingmultimodalcode}, together with the image-conditioned
lenses of InteractScience~\citep{chen2025interact} (VQT) and UniSVG~\citep{li2025unisvg} (ISVGen).

\paragraph{Text-Centric Settings.} The text-centric suite comprises
ArtifactsBench~\citep{zhang2025artifactsbench}, PandasPlotBench~\citep{pandasplotbench2025timur},
and VisPlotBench~\citep{ni2026viscoder} in eight visualization languages, together with
InteractScience's programmatic lens (PFT) and UniSVG's text-conditioned lens (TSVGen).

\paragraph{Evaluation Protocol.} Evaluation follows each benchmark's official protocol, with
deterministic metrics reported alongside judge scores. Since some designated judges share a
model family with our pipeline, we check that the gains hold across judges
(\cref{subsec:reliability}). Details are presented in \cref{app:benchmarks}.

\subsection{Vision-Centric Results}
\label{subsec:results-vision}

\paragraph{Web Interfaces.} 
\modelnine{} and \modelthirtyfive{} lead open baseline on front-end generation, achieving judge scores more than 20 points higher than those of task-specific open baselines (\cref{tab:main-vision}). 
\modelnine{} also scores above proprietary model on DesignBench generation and WebCode2M CLIP score. DesignBench editing is the only web-interface metric on which both models score below all Claude 4.6 and Gemini 3 variants.
\input{Tables/main_vision}

\paragraph{Data Visualization.} On ChartMimic, both of our models lead all open baselines. Specifically, \modelthirtyfive outperforms most proprietary models under both settings. Notably, our 9B model demonstrates the most substantial improvement over its base (the lowest among all evaluated models) and even surpasses most proprietary models. 
This clearly demonstrates the quality and utility of our data.
In addition, the low-level score evaluates rendered text, layout, chart type, and color fidelity, which means our model is more capable of generating those details.

\paragraph{Scientific and Vector Graphics.} On chemical-structure code, \modelthirtyfive{} parses
every output and reaches the highest Tanimoto similarity in the table (97.56). On
InteractScience's visual lens both models lead the open baselines, while
image-conditioned SVG is the one vision-centric column led by an open baseline, VinciCoder-8B. Full per-benchmark metrics are presented in
Appendix~\cref{app:results}, and a chart-to-code case across training stages in Appendix~\cref{app:cases}.

\subsection{Text-Centric Results}
\input{Tables/main_text}

\paragraph{Interactive Applications and Plots.} 
On ArtifactsBench, both models exceed all open baselines and \gptfouro{} on the overall score (\cref{tab:main-text}). On InteractScience's programmatic lens, \modelthirtyfive{} scores 28.00, above the best open baseline and comparable to Gemini 3 Flash. Scaling from 9B to 35B raises the InteractScience PFT score by 3.20 points.

\paragraph{Long-Tail Visual Languages.} 
\modelthirtyfive{} leads open baselines on all three VisPlotBench metrics (\cref{tab:main-text}). Its Mean score is 11.02 points above their best, comparable to Claude Sonnet 4.6.

\label{subsec:results-text}

%% file: Tables/main_vision.tex
\begin{table}[ht]
\centering
\caption{Vision-centric results. \textbf{Best}\,/\,\underline{second} among open
models.}
\label{tab:main-vision}
\setlength{\tabcolsep}{3pt}
\resizebox{\linewidth}{!}{%
\begin{tabular}{l cc ccc cc cc c c cc}
\toprule
& \multicolumn{2}{c}{ChartMimic} & \multicolumn{3}{c}{WebCode2M}
& \multicolumn{2}{c}{DesignBench} & \multicolumn{2}{c}{Design2Code}
& \multicolumn{1}{c}{\makecell{Inter-\\actSci.}} & \multicolumn{1}{c}{UniSVG}
& \multicolumn{2}{c}{ChemDraw} \\
\cmidrule(lr){2-3}\cmidrule(lr){4-6}\cmidrule(lr){7-8}\cmidrule(lr){9-10}\cmidrule(lr){11-11}\cmidrule(lr){12-12}\cmidrule(lr){13-14}
Model
& \makecell{Low-Level} & High-Level
& \makecell{LLM\\Score} & Visual & CLIP
& Gen. & Edit.
& \makecell{LLM\\Score} & Visual
& \makecell{VQT}
& \makecell{ISVGen}
& Validity & \makecell{Tani.\\Sim.} \\
\midrule
\multicolumn{14}{l}{\textit{Open models}} \\
Qwen3-VL-8B-Instruct & 65.92 & 62.81 & 53.67 & 73.55 & 79.16 & 55.12 & 8.38 & 57.47 & 82.81 & 34.06 & 66.27 & 78.91\% & 54.79 \\
Qwen3-VL-30B-A3B-Instruct & 71.04 & 67.30 & 54.80 & 77.40 & 79.42 & 51.58 & 8.07 & 57.94 & 85.11 & 37.67 & 74.30 & 82.03\% & 54.55 \\
Qwen3.5-9B & 58.35 & 55.55 & 51.74 & 81.60 & 79.93 & 50.73 & 8.19 & 53.12 & 85.60 & 27.28 & 63.09 & 88.28\% & 66.79 \\
Qwen3.5-35B-A3B & 70.22 & 63.90 & 65.32 & 88.64 & 85.94 & 64.70 & 8.66 & 67.49 & 90.98 & 29.33 & 62.91 & 92.97\% & 79.86 \\
JanusCoderV-7B & 75.12 & 68.97 & 40.92 & 72.90 & 76.00 & 40.37 & 7.44 & 36.49 & 70.81 & 27.67 & 61.60 & 84.75\% & 72.81 \\
JanusCoderV-8B & 76.54 & 69.23 & 35.17 & 65.95 & 73.04 & 34.30 & 7.26 & 34.53 & 68.61 & 33.32 & 62.47 & 86.10\% & 76.91 \\
VinciCoder-7B & 78.30 & 66.09 & 43.11 & 72.42 & 78.59 & 43.24 & 1.40 & 42.30 & 76.64 & 0.00 & \underline{85.72} & 93.18\% & 92.15 \\
VinciCoder-8B & 77.59 & 66.44 & 46.16 & 78.26 & 80.79 & 48.09 & 1.71 & 45.72 & 81.89 & 2.33 & \textbf{87.18} & 94.21\% & 91.02 \\
\rowcolor{boxblue} \textbf{\modelnine} & \underline{81.57} & \underline{73.82} & \textbf{69.72} & \textbf{90.41} & \textbf{88.46} & \textbf{69.95} & \textbf{8.93} & \textbf{70.51} & \textbf{91.54} & \underline{51.87} & 69.64 & \underline{96.88\%} & \underline{95.94} \\
\rowcolor{boxblue} \textbf{\modelthirtyfiveA} & \textbf{83.42} & \textbf{80.69} & \underline{67.78} & \underline{89.66} & \underline{87.78} & \underline{68.98} & \underline{8.89} & \underline{67.55} & \underline{91.22} & \textbf{53.54} & 84.14 & \textbf{100.00\%} & \textbf{97.56} \\
\midrule
\multicolumn{14}{l}{\textit{Proprietary}} \\
\rowcolor[gray]{0.92} \gptfouro & 76.38 & 72.02 & 54.94 & 83.45 & 80.22 & 52.64 & 8.89 & 58.48 & 88.23 & 46.01 & 70.98 & 92.19\% & 57.80 \\
\rowcolor[gray]{0.92} Gemini 3 Flash & 85.12 & 83.04 & 69.63 & 90.41 & 87.34 & 69.28 & 9.45 & 70.12 & 92.18 & 65.25 & 82.24 & 63.28\% & 81.39 \\
\rowcolor[gray]{0.92} Gemini 3.1 Pro & 77.96 & 81.73 & 70.97 & 89.96 & 87.64 & 69.47 & 9.57 & 70.90 & 90.70 & 71.05 & 76.85 & 93.12\% & 92.45 \\
\rowcolor[gray]{0.92} Claude Sonnet 4.6 & 80.23 & 80.73 & 68.19 & 87.76 & 86.80 & 66.53 & 9.24 & 67.99 & 90.83 & 77.36 & 80.38 & 88.28\% & 85.43 \\
\rowcolor[gray]{0.92} Claude Opus 4.6 & 81.11 & 79.68 & 66.38 & 88.25 & 87.11 & 65.69 & 9.55 & 66.75 & 90.19 & 76.25 & 78.70 & 96.88\% & 77.71 \\
\bottomrule
\end{tabular}}
\vspace{-0.5em}
\end{table}

%% file: Tables/main_text.tex
\begin{table}[t]
\centering
\footnotesize
\setlength{\tabcolsep}{4.5pt}
\caption{Text-centric results. \textbf{Best}\,/\,\underline{second} among open
models.} 
\label{tab:main-text}
\resizebox{0.92\linewidth}{!}{%
\begin{tabular}{l c ccc c c ccc}
\toprule
& \multicolumn{1}{c}{\makecell{Artifacts\\Bench}} & \multicolumn{3}{c}{PandasPlotBench}
& \multicolumn{1}{c}{\makecell{Inter-\\actSci.}} & \multicolumn{1}{c}{UniSVG}
& \multicolumn{3}{c}{VisPlotBench} \\
\cmidrule(lr){2-2}\cmidrule(lr){3-5}\cmidrule(lr){6-6}\cmidrule(lr){7-7}\cmidrule(lr){8-10}
Model
&
& \makecell{Incorr.\\Code$\downarrow$} & Visual & Task
& \makecell{PFT}
& \makecell{TSVGen}
& ExecPass & Mean & \makecell{Good\\$\geq$75} \\
\midrule
\multicolumn{10}{l}{\textit{Open models}} \\
Qwen3-VL-8B-Instruct & 34.71 & 16.00\% & 60.01 & 73.67 & 15.47\% & 68.13 & 56.52\% & 40.25 & 33.62 \\
Qwen3-VL-30B-A3B-Instruct & 41.70 & 11.43\% & 60.05 & \textbf{80.58} & 24.40\% & 69.33 & 64.38\% & 49.28 & 42.88 \\
Qwen3.5-9B & 41.24& 24.57\% & 52.98 & 65.47 & 22.40\% & 69.28 & 52.22\% & 41.18 & 36.71 \\
Qwen3.5-35B-A3B & 44.98  & 15.43\% & 60.16 & 75.17 & \underline{26.67\%} & \textbf{72.15} & 63.01\% & 51.45 & 48.51 \\
JanusCoderV-7B & 31.16 & 23.43\% & 53.05 & 65.77 & 17.73\% & 66.43 & 58.42\% & 48.10 & 32.15 \\
JanusCoderV-8B & 32.00 & 26.30\% & 56.30 & 70.91 & 17.60\% & 62.17 & 56.21\% & 47.22 & 30.87 \\
VinciCoder-7B & 11.35 & \textbf{9.14\%} & 25.37 & 51.57 & 0.40\% & 64.06 & 64.10\% & 23.95 & 10.11 \\
VinciCoder-8B & 11.97 & 12.00\% & 25.47 & 60.79 & 0.54\% & 63.45 & 63.92\% & 12.55 & 2.80 \\
VisCoder2-7B & 17.51 & 20.00\% & 59.48 & 67.59 & 1.00\% & 65.44 & \underline{66.73\%} & 49.84 & 42.50 \\
\rowcolor{boxblue} \textbf{\modelnine} & \underline{45.28} & 10.29\% & \underline{60.54} & 77.94 & 24.80\% & 68.49 & 65.30\% & \underline{53.99} & \underline{52.62} \\
\rowcolor{boxblue} \textbf{\modelthirtyfiveA} & \textbf{46.01} & \underline{9.29\%} & \textbf{62.38} & \underline{79.85} & \textbf{28.00\%} & \underline{70.72} & \textbf{71.14\%} & \textbf{62.47} & \textbf{61.49} \\
\midrule
\multicolumn{10}{l}{\textit{Proprietary}} \\
\rowcolor[gray]{0.92} \gptfouro & 33.54 & 8.00\% & 67.63 & 82.46 & 31.07\% & 72.80 & 67.99\% & 56.01 & 52.75 \\
\rowcolor[gray]{0.92} Gemini 3 Flash & 51.42 & 3.43\% & 69.95 & 91.21 & 28.13\% & 74.64 & 76.25\% & 60.92 & 58.54 \\
\rowcolor[gray]{0.92} Gemini 3.1 Pro & 53.04 & 1.71\% & 78.41 & 94.70 & 44.13\% & 74.52 & 79.57\% & 69.47 & 69.15 \\
\rowcolor[gray]{0.92} Claude Sonnet 4.6 & 64.65 & 3.43\% & 74.59 & 91.93 & 37.33\% & 74.93 & 73.00\% & 62.20 & 61.06 \\
\rowcolor[gray]{0.92} Claude Opus 4.6 & 62.47 & 1.71\% & 76.03 & 94.97 & 35.47\% & 75.47 & 73.09\% & 65.94 & 66.03 \\
\bottomrule
\end{tabular}}
\end{table}

%% file: Sections/5-analysis.tex
\needspace{6\baselineskip}  
\section{Analysis}
\label{sec:analysis}

\subsection{Rubric Efficacy}
\label{subsec:efficacy}

To intuitively demonstrate the advantages of the rubric-based method over direct judging, we fine-tune Qwen3.5-9B on two equal-sized subsets drawn from the same data pool.
\begin{figure}[!ht]
\centering
\includegraphics[width=\linewidth]{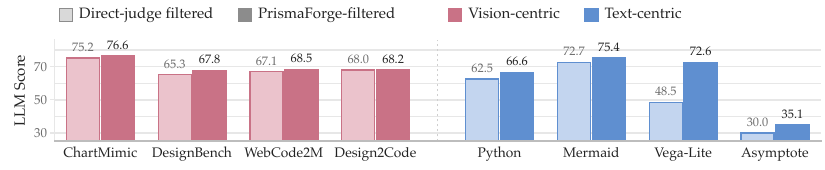}
\caption{\pipename{} filtering outperforms direct-judge filtering on every benchmark.}
\label{fig:filter-ablation}
\end{figure}

One is filtered by direct scoring, where a judge assigns each
sample a single holistic score as in a typical pipeline~\citep{sun2026januscoder}, and the
other by the \pipename{} rubric. The model trained on the rubric-filtered subset scores higher on
every benchmark, by as much as 24 points on Vega-Lite (\cref{fig:filter-ablation}). The
gains are largest on the text-centric formats and small on the vision-centric benchmarks,
where Design2Code moves by a fraction of a point. Since both subsets have the same size, the gap
reflects which samples each judge keeps.

\subsection{Backbone Generalization}
\label{subsec:backbone}
\begin{figure}[t]
\centering
\includegraphics[width=\linewidth]{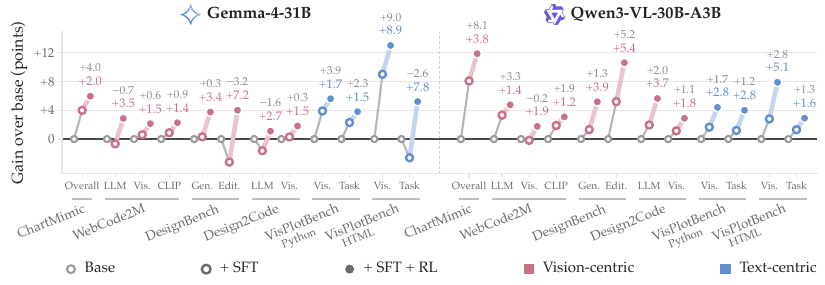}
\caption{Per-stage gains of our SFT and RL pipeline over two base models on every metric.}
\label{fig:backbone-stages}
\end{figure}
Beyond the released Qwen3.5 backbones, we apply the same data and method to
Gemma-4-31B-it~\citep{google2026gemma} and Qwen3-VL-30B-A3B-Instruct~\citep{bai2025qwen3} on
Python plotting and HTML, the two most represented domains. On both backbones all twelve metrics
end above the base model, and RL adds to SFT on every one (\cref{fig:backbone-stages}). SFT
carries most of the ChartMimic gain, whereas RL drives most of the HTML gains and recovers every
metric that dips after SFT alone. More results on Gemma are presented in Appendix~\cref{app:backbone}.

\needspace{17\baselineskip}  
\subsection{Rubric Agreement with Humans}
\label{subsec:reliability}
\begin{wrapfigure}{r}{0.48\linewidth}
\centering
\includegraphics[width=\linewidth]{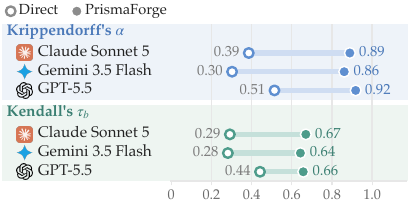}
\caption{\pipename{} scoring aligns judges with humans better than direct
scoring.}
\label{fig:judge-agreement}
\end{wrapfigure}
Because multimodal code artifacts are ultimately created for humans, consistency with human judgment provides a compelling test to verify that the quality of data yielded by our rubric-based PrismaForge aligns more closely with human preferences.
At the same time, preserving the relative ordering induced by scores is particularly important for reward-based policy optimization, where inconsistent rankings can provide conflicting training signals.
To evaluate our method along both dimensions, we conducted a human study. We sampled 100 examples from our training data and asked human annotators, as well as three proprietary model judges, to evaluate them using direct scoring and the \pipename{} rubrics, respectively. We then measured the agreement between each model judge and the human ratings in terms of both absolute score levels and pairwise ordering (\cref{fig:judge-agreement}). The rubric-based judging method achieves substantially higher agreement with human judgments and greater consistency in the induced partial ordering across all three model judges. The full study protocol and detailed statistics are provided in Appendix~\cref{app:adjudication}.
We further show that the gains of our pipeline hold across judge models, with the
experiments presented in Appendix~\cref{app:backbone}.

%% file: Sections/6-conclusion.tex
\section{Conclusion}
\label{sec:conclusion}

In this work, we push open multimodal code intelligence forward by jointly
contributing and fully releasing data, pipeline, and models. \pipename{} treats
multimodal code data under one lens with rubrics synthesized from coarse to fine
and decomposed into code-level and visual-level criteria, a signal that both
generalizes and localizes what went wrong.
We then curate and release \ourdata{}, the most broadly covering open dataset
for multimodal code intelligence, which enables principled selection and stable reinforcement learning on
visual-coding tasks. On this data we build \modelnine{}/35B, which lead the open state of the art
by a wide margin and rival proprietary systems. Extensive analyses across base models and task recipes show that the data and
the pipeline generalize beyond our release, and we share these insights and resources
to broaden access to visual-coding capabilities.

%% file: Appendices/acknowledgments.tex
\section*{Acknowledgments}
\label{sec:ack}

This research is supported by the WYNG Foundation (HKU 25AG100407).
We thank Xiao Wang for advice on molecular generation and Xuanle Zhao for
guidance on reproducing the baselines.

%% file: Appendices/appendix_content.tex
\section{Data Toolkit Details}
\label{app:toolkit}

\subsection{AST-Based Decomposition}
\label{app:toolkit:ast}
Plotting scripts collected from GitHub and Manim scene files are often long, and one file may
draw several figures. 
For such data, we parse each script into an abstract syntax tree without executing it and split it into self-contained units. A unit is the code that produces one figure in a plotting script, or one scene in a Manim file (a class inheriting from \texttt{Scene}). Each unit keeps the imports and statements it depends on,
drops project-private helpers that exist only in the source repository, and carries its embedded
text and a short excerpt as features. The units are stored as structured entries that the
synthesis pipeline and the gatekeeper of Section~\cref{subsec:data-gating} read. All of these processing pipelines will be made public.

\subsection{Per-Format Gatekeepers}
\label{app:toolkit:gatekeeper}
Every candidate runs in a sandbox matched to its language before it is judged. Python plotting
code executes headlessly and must save a figure, and HTML pages render in a headless browser to a
screenshot. SVG, \LaTeX{}, Mermaid, Asymptote, LilyPond, and Vega-Lite sources go through their
standard renderers or compilers, Manim scenes render to their frames, and chemical-structure code
must parse into a valid molecule. A candidate passes when the run finishes within a fixed time
limit and produces its artifact, and it fails on an error, a timeout, or an empty artifact. The
same gatekeeper serves data gating (\cref{subsec:data-gating}) and reinforcement learning
(\cref{subsec:rl}), where a failed rollout receives reward $0$.

\subsection{Decontamination}
\label{app:toolkit:decontamination}
Part of \ourdata{} is synthesized by frontier models that may have seen public benchmarks, so we
decontaminate the released data against the test sets of every benchmark in \cref{subsec:benchmarks}.
A training sample is removed when its instruction shares a long $n$-gram with a test instruction.

\section{Data Sources}
\label{app:sources}

\input{Tables/appendix/data-source}

Table~\ref{tab:data_source_stat} provides overall statistics on the samples used from each data source, as well as the number of samples processed at each stage of the gating procedure. For datasets that are higher in quality and more closely aligned with our target tasks, more than half of the original responses can be retained. In contrast, broader and less task-specific data sources, such as GitHub and arXiv papers, require more extensive processing, with a larger fraction of samples needing to be regenerated.

\paragraph{Coverage Compared with Prior Open Work.}
\Cref{tab:comparison} sets the data types and the method of \ourdata{} beside the open
multimodal code-intelligence works of \cref{sec:related}, read from their public papers. A
check marks a type present in a work's released training corpus as its paper describes it,
and the rows where our data is absent are listed as well.
\input{Tables/comparison}

\section{Rubric Synthesis Prompts and Examples}
\label{app:prompts}

\newtcolorbox{promptbox}[2][]{enhanced, width=\linewidth, arc=2mm, boxrule=0.6pt,
  colframe=#2Line, colback=#2Fill!45!white, colbacktitle=#2Line, coltitle=black!85,
  fonttitle=\bfseries\small, title={#1},
  left=3.5mm, right=3.5mm, top=1mm, bottom=1.2mm, toptitle=0.6mm, bottomtitle=0.6mm}
\newcommand{\promptsec}[1]{\par\vspace{1.5pt}{\color{black!75}\scshape\bfseries #1}\par\vspace{0pt}}
\newcommand{\promptsub}[1]{{\itshape #1}\par\vspace{-1pt}}
\newcommand{\pvar}[1]{\texttt{\{#1\}}}
\newenvironment{promptlist}{\begin{enumerate}[leftmargin=1.3em, itemsep=0pt, topsep=0pt, parsep=0pt, partopsep=0pt]}{\end{enumerate}}
\newenvironment{jsonbox}{\par\vspace{1.5pt}\begin{tcolorbox}[colback=black!4, colframe=black!18, boxrule=0.4pt, arc=1mm,
  left=2mm, right=2mm, top=0.6mm, bottom=0.6mm, boxsep=0pt]\ttfamily\scriptsize\setlength{\parskip}{0pt}}{\end{tcolorbox}}

Web UI reconstruction must be evaluated from two complementary perspectives: whether the generated HTML/CSS implements the required content and structure, and whether its rendered output reproduces the intended visual appearance. 
A single holistic score often conflates these two aspects and provides little evidence about the source of an error. 
We therefore synthesize two sets of sample-specific binary rubrics from the task instruction and reference screenshot. 
Code rubrics target properties that can be verified directly from the implementation, whereas vision rubrics target properties observable in the rendered page. 
\Cref{fig:code-rubric-prompt,fig:vision-rubric-prompt} show the corresponding prompt templates. 
Both prompts organize checks into shared quality criteria and sample-dependent requirements, and assign each check a weight according to its importance. 
The difficulty level assigned to the instruction enters both prompts as an item budget, so harder samples receive more criteria (\cref{subsec:rubric}). 
The templates are abridged to remove repetitive wording.

\begin{figure}[!ht]
\centering
\begin{promptbox}[Code Rubric Prompt (Web UI Generation)]{figCode}
\scriptsize\linespread{0.9}\selectfont\setlength{\parskip}{1.5pt}\setlength{\parindent}{0pt}
You are an expert evaluator of Web UI generation. Your task is to generate a sample-specific binary rubric for evaluating the generated HTML/CSS from code evidence.
\promptsec{Sample Inputs}
\begin{promptlist}
\item Problem statement: \pvar{prompt}
\item Input screenshot: [the input screenshot is attached to this message]
\item Difficulty level: \pvar{difficulty}, with a budget of \pvar{min\_items}--\pvar{max\_items} rubric items in total
\end{promptlist}
\promptsec{Task}
Generate code rubrics that can later be judged from the generated HTML/CSS code alone, while remaining grounded in the problem statement and screenshot. Stay within the item budget and spend it on the properties that matter most for this sample. Every question must be decidable from the code alone; do not ask about the rendered appearance.
Each item contains (1) a yes/no \texttt{question}, where 1 is pass and 0 is fail, and (2) a \texttt{weight} in $[0,1]$: $0.8$--$1.0$ for critical, $0.4$--$0.7$ for important, and $0.1$--$0.3$ for minor requirements.
\promptsec{Rubric Dimensions}
\noindent\begin{minipage}[t]{0.485\linewidth}
\promptsub{Shared rubrics (recur across the domain)}
\begin{promptlist}
\item \textbf{Screenshot faithfulness.} Check whether the code captures the major visible regions and their relationships.
\item \textbf{Render safety and structural integrity.} Check whether the HTML/CSS is complete and can render without structural or layout-breaking errors.
\item \textbf{Placeholder image handling.} Check whether placeholders are present, constrained, and placed only where the reference requires them.
\end{promptlist}
\end{minipage}\hfill
\begin{minipage}[t]{0.485\linewidth}
\promptsub{Task-specific rubrics (unique to this instruction)}
\begin{promptlist}
\item \textbf{Layout mapping and section structure.} Check the order, nesting, alignment, and spatial organization of page sections.
\item \textbf{Content element coverage.} Check whether important text, controls, repeated items, and media regions are represented.
\item \textbf{Self-contained implementation.} Check that the page uses inline CSS and no unpermitted external dependencies or assets.
\end{promptlist}
\end{minipage}
\promptsec{Output Format}
Return only valid JSON following this structure:
\begin{jsonbox}
\{\\
\hspace*{1em}"shared\_rubrics": \{\\
\hspace*{2em}"screenshot\_faithfulness": [ \{"question": "...", "weight": ...\}, ... ], ...\\
\hspace*{1em}\},\\
\hspace*{1em}"task\_specific\_rubrics": \{ "layout\_mapping\_section\_structure": [...], ... \}\\
\}
\end{jsonbox}
\end{promptbox}
\caption{Prompt template for synthesizing code rubrics in Web UI generation.}
\label{fig:code-rubric-prompt}
\end{figure}

\begin{figure}[!ht]
\centering
\begin{promptbox}[Vision Rubric Prompt (Web UI Generation)]{figVis}
\scriptsize\linespread{0.9}\selectfont\setlength{\parskip}{1.5pt}\setlength{\parindent}{0pt}
You are an expert evaluator of Web UI generation. Your task is to generate a sample-specific binary rubric for evaluating the generated webpage from visual evidence in its rendered screenshot.
\promptsec{Sample Inputs}
\begin{promptlist}
\item Problem statement: \pvar{prompt}
\item Input screenshot: [the input screenshot is attached to this message]
\item Difficulty level: \pvar{difficulty}, with a budget of \pvar{min\_items}--\pvar{max\_items} rubric items in total
\end{promptlist}
\promptsec{Task}
Generate vision rubrics that can later be judged from the rendered screenshot, while remaining grounded in the problem statement and input screenshot. Stay within the item budget and spend it on the properties that matter most for this sample. Every question must be decidable from the rendered screenshot alone; do not ask about the source code.
Each item contains (1) a yes/no \texttt{question}, where 1 is pass and 0 is fail, and (2) a \texttt{weight} in $[0,1]$: $0.8$--$1.0$ for critical, $0.4$--$0.7$ for important, and $0.1$--$0.3$ for minor requirements.
\promptsec{Rubric Dimensions}
\noindent\begin{minipage}[t]{0.485\linewidth}
\promptsub{Shared rubrics (recur across the domain)}
\begin{promptlist}
\item \textbf{Screenshot faithfulness.} Check whether the output preserves the reference's major regions, hierarchy, and visual relationships.
\item \textbf{Visual integrity.} Check for overlap, clipping, collapse, distortion, or other visible rendering failures.
\item \textbf{Text legibility.} Check whether visible text remains readable and unobscured.
\end{promptlist}
\end{minipage}\hfill
\begin{minipage}[t]{0.485\linewidth}
\promptsub{Task-specific rubrics (unique to this instruction)}
\begin{promptlist}
\item \textbf{Layout structure and alignment.} Check the placement, spacing, and alignment of the major sections and elements.
\item \textbf{Content element coverage.} Check whether important visible text, controls, repeated items, and media regions appear.
\item \textbf{Placeholder image normality.} Check whether placeholders have the expected count, size, and placement without causing visible artifacts.
\end{promptlist}
\end{minipage}
\promptsec{Output Format}
Return only valid JSON following this structure:
\begin{jsonbox}
\{\\
\hspace*{1em}"shared\_rubrics": \{\\
\hspace*{2em}"screenshot\_faithfulness": [ \{"question": "...", "weight": ...\}, ... ], ...\\
\hspace*{1em}\},\\
\hspace*{1em}"task\_specific\_rubrics": \{ "layout\_structure\_alignment": [...], ... \}\\
\}
\end{jsonbox}
\end{promptbox}
\caption{Prompt template for synthesizing vision rubrics in Web UI generation.}
\label{fig:vision-rubric-prompt}
\end{figure}

The complementary roles of the two rubric views are illustrated by the representative reconstruction in \cref{fig:rubric-example}. 
Code rubrics verify that the expected elements and implementation constraints are present, whereas vision rubrics reveal rendering failures that may remain hidden in
otherwise valid HTML/CSS. 
In this example, the synthesized page preserves the reference's overall content, colors, and organization, yet contains a salient local layout failure. 
It receives a perfect code-rubric score of $1.0$, indicating that the implementation contains the expected structure and content. 
It also obtains a holistic LLM judgment score of $78$, outperforming approximately $78.6\%$ of all samples. 
Nevertheless, our sample-specific vision rubrics assign an aggregate score of only $0.182$. 
They identify that the footer is incorrectly positioned in the middle of the page, where it overlaps and clips the third service button and disrupts the intended vertical hierarchy. 
They also capture that the synthesized buttons have square rather than rounded corners. 
We next examine the individual judgments that produce this discrepancy.

\paragraph{Rubric-level breakdown.}
\Cref{tab:case-rubrics} reports the individual
checks behind these aggregate scores. The code evaluator passes all five
checks because the source contains the expected colors, stacking mechanism,
centering rule, footer styling, and copyright text. In contrast, the rendered
output passes only the horizontal-centering check; the remaining vision
rubrics expose the misplaced footer, clipped service button, and incorrect
button corners.

\begin{table}[H]
\centering
\footnotesize
\renewcommand{\arraystretch}{1.0}
\setlength{\tabcolsep}{4pt}
\begin{tabular}{>{\raggedright\arraybackslash}p{0.37\textwidth} c c >{\raggedright\arraybackslash}p{0.42\textwidth}}
\toprule
Rubric & Weight & Pass & Judgment evidence \\
\midrule
\multicolumn{4}{l}{\textit{Code rubrics} (all pass; weighted score $1.0$)} \\
The three Product blocks use a teal-like background. & 0.8 & \cmark & The buttons explicitly use \texttt{background-color: \#008080}. \\
The three Product items are stacked vertically. & 0.8 & \cmark & The shared button class uses \texttt{display: block}. \\
The ``Technology Company'' heading is horizontally centered. & 0.8 & \cmark & The header uses \texttt{text-align: center}. \\
The footer containing the copyright text has a dark background. & 0.7 & \cmark & The footer explicitly uses \texttt{background-color: \#333}. \\
The footer contains the required copyright statement. & 0.9 & \cmark & The exact statement is present, with \texttt{\&copy;} encoding the copyright symbol. \\
\midrule
\multicolumn{4}{l}{\textit{Vision rubrics} (one pass; weighted score $0.8/(1.0+0.8+0.9+0.9+0.8)=0.182$)} \\
The page preserves the centered vertical hierarchy from the main heading through the footer. & 1.0 & \xmark & The footer overlaps the Services section instead of following it at the bottom. \\
All primary content is horizontally centered without unintended horizontal scrolling. & 0.8 & \cmark & The content is centered and no horizontal scrollbar is visible. \\
The teal and purple buttons form vertically stacked blocks with rounded corners. & 0.9 & \xmark & The buttons are stacked, but their corners are sharp rather than rounded. \\
The footer is a dark, full-width bar at the bottom with centered white text. & 0.9 & \xmark & Although its color and width are correct, the footer is placed over the last service item rather than at the bottom. \\
Exactly three teal Product buttons and three purple Service buttons are visible. & 0.8 & \xmark & The misplaced footer obscures and clips the third Service button. \\
\bottomrule
\end{tabular}
\caption{Code and vision rubrics with their judgments for the example in \cref{fig:rubric-example}.}
\label{tab:case-rubrics}
\end{table}

\paragraph{Holistic judgment.}
The holistic evaluator assigns a score of $78$ with this rationale:
\begin{quote}
\small
The rendered image matches the reference closely in fonts, colors, and general
structure. However, there is a layout error where the footer is positioned too
high, overlapping the ``Service 3'' button and leaving empty space at the
bottom.
\end{quote}
Thus, the holistic evaluator notices the dominant layout error but compresses
it into a relatively high scalar score. The rubric-based evaluation both
amplifies its impact according to sample-specific requirements and identifies
the additional corner-shape and content-visibility failures. These localized
but perceptually important errors are easily diluted in a single holistic
similarity score, whereas our fine-grained rubrics explicitly expose and
penalize each failure. The two rubric
views thus provide more diagnostic and structurally sensitive feedback than
code inspection or holistic visual judgment alone.

In our experiments, we use Gemini-3-Flash for rubrics. For common visualization tasks, such as Python plotting, the average cost of generating rubrics for a single sample is approximately \$0.006 based on the official API pricing, while evaluating the generated code and visual artifacts costs approximately \$0.01 per sample. For HTML-related tasks, the input cost is substantially higher because rich, complex websites typically involve much longer HTML code. In these cases, rubric generation costs approximately \$0.01 per sample, while evaluation costs approximately \$0.04.

\section{RL Reward Specification}
\label{app:reward}

This appendix specifies the reward computation procedure behind \cref{subsec:rl}. 

For each rollout response $c_i^{(k)}$, the procedure first extracts code from the response text. For Python, we extract the longest fenced code block. For HTML, we extract a single complete document spanning from the doctype declaration to the closing tag. For SVG, we extract either the code contained in a fenced block or the first complete \texttt{<svg>} span. A completion that does not contain valid code of the required type is assigned a reward of $0$ without further evaluation.

A dispatcher then routes each rollout to exactly one task family, and the scorer registered for that family returns a reward in $[0,1]$, where $1$ denotes a fully satisfactory solution. An unscoreable completion (due to a missing or malformed answer, unavailable rubric or reference material, execution or rendering failure, or a judge failure that persists after retries) receives a reward of $0$.

The basic reward computation for each task largely follows Eq.~\ref{eq:modality-score} and Eq.~\ref{eq:overall-score}. Specifically, the response code and the rendered visual artifact are scored separately and then aggregated using a weighted average. We use different combinations of code-score and visual-score weights across task families. This is because for tasks such as Python plotting, correct data representation and placement are as important as visual integrity and presentation quality. In contrast, for educational demonstration artifacts, visual readiness and interactivity are often more important than the particular implementation used to produce them.

We found, however, that predefined rubrics alone may fail to capture certain forms of misalignment in a few tasks such as interface reproduction. Instructions for these tasks inherently allow some flexibility in content generation, particularly with respect to styling and layout organization. As a result, some errors or mismatches can only be assessed reliably after the generated page is viewed alongside the reference page. To address this limitation, we introduce a complementary holistic score for task families of this type. This score is combined with the rubric-based scores using a weight denoted by $\lambda_h$. In this way, the evaluation retains the consistency and structure of rubric-based scoring while also capturing qualitative discrepancies that become apparent only when the generated artifact is evaluated as a whole.

The reward logic and the weight combinations used for each task domain are summarized in \Cref{tab:reward-spec}.

For each sample, the rubric-based method requires one API call to score each modality. Because code and visual artifacts are evaluated separately, its API cost is approximately $2\times$ that of directly scoring the visual outputs alone. In our experiments, a full round of GRPO training involves approximately 808K judgment requests. Resource constraints prevented a full GRPO comparison between \pipename{} and direct scoring. In a focused Python plotting experiment, rubric-based scoring improved ChartMimic overall by 6.91 points over direct scoring.

\input{Tables/appendix/reward_spec}

\section{Benchmark Details and Evaluation Settings}
\label{app:benchmarks}

Code-driven video and animation, and theorem and symbolic visualization, are released as
training data only and are not part of the evaluation.

\paragraph{Vision-Centric Benchmarks.} ChartMimic reproduces a chart from its
image; its Direct split regenerates the chart and its Customized split edits an
existing one, scored by low-level element extraction (text, layout, type, color)
and a high-level judge score. WebCode2M, DesignBench, and Design2Code cover
front-end generation from a screenshot: WebCode2M buckets pages by length and
reports a judge score, visual similarity, CLIP, and TreeBLEU; DesignBench scores
generation and editing with a judge score plus CLIP, MAE, SSIM, and CMS;
Design2Code reports element-level block, text, position, and color matches, CLIP,
and a judge score. InteractScience's visual lens (VQT) is a visually grounded
qualitative test judged by a vision--language model. UniSVG's image-conditioned
lens (ISVGen) reproduces an image as SVG. ChemDraw measures execution rate,
molecule validity, exact match, and Tanimoto similarity.

\paragraph{Text-Centric Benchmarks.} ArtifactsBench reports an overall score
over ten artifact domains. PandasPlotBench reports the incorrect-code rate and
mean and Good($\geq$75) scores under a visual and a task lens. InteractScience's
programmatic lens (PFT) is a functional test over the generated demonstration.
UniSVG's text-conditioned lens (TSVGen) generates SVG from text. VisPlotBench
covers eight languages with the same metrics as PandasPlotBench, using its designated
judge GPT-4.1~\citep{openai2025gpt41}.

\paragraph{Evaluation Settings.} Evaluation follows the benchmark's official
protocol, including its designated judge model where one is specified, so that
our numbers stay comparable with published results. 
Due to the availability of legacy models, ChartMimic's high-level score is
judged by GPT-5.5~\citep{openai2026gpt55} instead of the original \gptfouro{}.

\section{Training Details}
\label{app:training}

All experiments were conducted on four interconnected compute nodes, each equipped with 8 NVIDIA H200 GPUs, for a total of 32 GPUs. We used the Megatron backend in the ms-swift framework for distributed SFT and RL. Both stages performed full-parameter fine-tuning with FlashAttention and sequence parallelism enabled. Activation recomputation reduced memory use throughout training.

For SFT, we initialized from Qwen3.5-9B/Qwen3.5-35B-A3B and trained the language model for 5 epochs.  We used a maximum sequence length of 32,768 tokens and enabled sequence packing for improved token utilization. The global batch size was set to be 128. The learning rate peaked at $1\times10^{-5}$, with a warm-up fraction of 5\%, and decayed to a minimum learning rate of $5\times10^{-7}$.

For RL, we initialized from the SFT checkpoint and performed one epoch of GRPO/GSPO. We implemented a rubric-based reward plugin following the methodology described in Section~\ref{subsec:rl} and Appendix~\ref{app:reward}. For the judge model $L^J$ used to evaluate online rollouts, we selected Gemini-3-Flash to balance evaluation quality and cost. The global training batch size was 512 in terms of rollouts (equivalent to 64 samples per batch) and 8 candidate responses were generated per prompt at each rollout step. Rollout generation was performed with colocated vLLM inference using tensor parallelism of size 8 and a maximum model length of 36,000 tokens. Sampling used a temperature of 1.0 and top-$p$ of 0.95.

The GRPO/GSPO learning rate was set to $3\times10^{-6}$. We used a KL regularization coefficient of $\beta=0.04$, and asymmetric clipping thresholds of $\epsilon=0.20$ and $\epsilon_{\mathrm{high}}=0.28$. Dynamic sampling and overlong-response filtering were enabled during rollout collection. Training was conducted in BF16 precision, with model and optimizer offloading enabled to reduce GPU memory usage.

\section{Full Per-Benchmark Results}
\label{app:results}
Each benchmark reports several metrics; the main text shows a compact headline set
per benchmark, and the complete metric sets are here, one table per benchmark,
mirroring our evaluation workbook.

\subsection{Visualization}

\paragraph{ChartMimic.}

The complete results of the direct-1800 split of the ChartMimic benchmark is shown in Table~\ref{tab:chartmimic-full}. We can see that PrismaCoder shows strong performance against both open and proprietary models. PrismaCoder-9B achieves an overall score of 77.70, substantially outperforming other similarly sized open models (around 72–73) and even surpassing GPT-4o (74.20). Scaling to PrismaCoder-35B further improves the overall score to 82.05, with particularly strong execution rate (96.16), text fidelity (90.41), layout (94.00), and GPT-based similarity (80.69). Its overall performance is competitive with much larger proprietary models, exceeding Claude Sonnet 4.6 and Claude Opus 4.6 in this evaluation, while approaching Gemini 3 Flash. These results suggest that PrismaCoder is especially effective at faithfully reconstructing chart structure and visual details while maintaining reliable executable code.
\begin{table}[htp]
\centering
\footnotesize
\input{Tables/appendix/chartmimic}
\end{table}

\paragraph{PandasPlotBench.}

Table~\ref{tab:pandasplot-full} illustrates a more detailed results of PandasPlotBench. On this benchmark, PrismaCoder demonstrates competitive visualization-code generation performance across both model scales. PrismaCoder-9B achieves a mean visual/task score of 60.54/77.94 with 14.29\% incorrect code, comparing favorably with other models in a similar size range, while PrismaCoder-35B further reduces the incorrect-code rate to 9.29\% and reaches 63.38/79.85, with 60.6\% of examples attaining a visual score above 75. These gains are relatively modest compared with some strong baselines, which is unsurprising given the nature of the benchmark. Introduced in 2024, the benchmark contains 175 human-curated, largely Matplotlib-gallery-derived tasks and evaluates plotting from controlled natural-language descriptions, including synthetically shortened variants. In contrast, much of PrismaCoder's training data is curated toward more realistic and diverse visualization workflows. We include this classic Python visualization benchmark to assess our models on established plotting tasks.

\begin{table}[htp]
\centering
\footnotesize
\begin{minipage}[t]{0.72\linewidth}
\centering
\input{Tables/appendix/pandasplotbench}
\end{minipage}
\end{table}

\paragraph{VisPlotBench.}
Table~\ref{tab:visplot-avg} averages the three metrics over the eight languages, and
Table~\ref{tab:visplot-full} lists the incorrect-code rate per language. \modelthirtyfive{} has
the lowest incorrect-code rate of every model in the table (28.86\%) and the highest mean and
good-plot scores under both lenses. The mean task score of
\modelthirtyfive{} is 62.47, against 23.95 for the best VinciCoder model, and \modelnine{}
leads both VinciCoder models on every column. Per language, the remaining failures concentrate
on LilyPond, \LaTeX{}, and Asymptote, where both models still break more programs than the
specialized open baselines. On the web and diagram formats the picture reverses:
\modelthirtyfive{} has the lowest incorrect-code rate of every model on Vega-Lite and HTML and
produces no broken program on SVG, and \modelnine{} is the strongest open model on HTML.
\begin{table}[htp]
\centering
\footnotesize
\begin{minipage}[t]{0.5\linewidth}
\centering
\input{Tables/appendix/visplotbench_avg}
\end{minipage}\hfill
\begin{minipage}[t]{0.46\linewidth}
\centering
\input{Tables/appendix/unisvg}
\end{minipage}
\end{table}
\begin{table}[htp]
\centering
\footnotesize
\input{Tables/appendix/visplotbench_languages}
\end{table}

\paragraph{UniSVG.}
Table~\ref{tab:unisvg} reports the image-conditioned lens (ISVGen), the text-conditioned lens
(TSVGen), and the benchmark's final score. \modelthirtyfive{} has the highest final score in the
table (72.96), and both of our models lead every open baseline on TSVGen. VinciCoder's strength sits in ISVGen alone, the lens its training
targets, and its text-conditioned scores are the lowest among the open models.
\modelthirtyfive{} comes within about three points of it on ISVGen and leads it by seven on
TSVGen, so the final score favors the balanced profile.

\paragraph{ChemDraw.}
Table~\ref{tab:chemdraw} adds exact match and two Tanimoto thresholds to the two columns of the
main table. \modelthirtyfive{} leads every column. All 128 of its outputs parse, 92.19\% match
the reference molecule exactly, and 95.31\% reach a Tanimoto similarity of 1. \modelnine{} is
second on every column and above every proprietary model on exact match and Tanimoto
similarity. Both are large gains over the Qwen3.5 bases, whose exact-match rates are 33.59\% and
53.12\%. The mean Tanimoto similarity is averaged over RDKit-valid outputs only, so the 81.39 of
Gemini 3 Flash covers the 63\% of its outputs that parse, whereas ours cover every sample.
\begin{table}[htp]
\centering
\footnotesize
\input{Tables/appendix/chemdraw}
\end{table}

\subsection{WebDev}

\paragraph{WebCode2M.}
Table~\ref{tab:webcode2m-full} shows that \methodname{}'s advantage is consistent across webpages of different code lengths.
\modelnine obtains Visual scores of 90.06, 90.02, and 91.15 on the short, mid, and long subsets, respectively, while its LLM score remains similarly stable across the three groups.
CLIP similarity decreases modestly with page length, from 90.29 to 87.32, but remains well above the prior open models throughout.
The main exception is TreeBLEU: both \methodname{} models score below JanusCoder and VinciCoder despite producing substantially more faithful renderings according to the visual metrics.
This difference arises because TreeBLEU measures similarity to the reference HTML tree rather than the rendered webpage itself.
JanusCoder and VinciCoder are trained on the original WebCode2M code, which naturally favors reproducing structures similar to the benchmark references.
\methodname{} instead uses regenerated code paired with webpages rendered from that code, rather than directly using the original WebCode2M source code.
This regeneration is intentional.
The original WebCode2M code comes from diverse real-world webpages and can differ substantially from the code distribution naturally produced by LLMs.
Using regenerated code therefore better aligns the training targets with inference-time generations, while re-rendering the regenerated programs ensures exact consistency between the code and its visual target.
Its generated webpages can consequently closely match the target appearance without following the particular HTML or DOM structure of the reference implementation.
The similarly low TreeBLEU scores of the proprietary models further illustrate that reference-code similarity does not necessarily track rendering fidelity.
\begin{table}[htp]
\centering
\footnotesize
\input{Tables/appendix/webcode2m}
\end{table}

\paragraph{DesignBench.}
The detailed results in Table~\ref{tab:designbench-full} show that the generation gains are consistent across multiple visual metrics.
\modelnine reaches a Judge score of 69.95 and a CLIP score of 84.77, while reducing MAE to 42.07 and increasing SSIM to 0.6956.
\modelthirtyfive exhibits a nearly identical profile and achieves the lowest MAE and highest SSIM among the evaluated models.
The agreement across Judge, CLIP, MAE, and SSIM indicates that the improvement is not tied to a particular evaluator, but is consistently reflected across multiple complementary measures of visual similarity.
Editing shows a more nuanced pattern: both \methodname{} models achieve an Editing-Judge score of about 8.9, substantially exceeding the prior open baselines and matching GPT-4o, while remaining slightly behind Gemini 3.1 Pro.
In contrast, their CMS and image-based similarity metrics are less competitive with the strongest proprietary models.
This difference can be partly attributed to the reference-dependent nature of these metrics.
CMS measures the overlap between the code lines modified by the model and those modified in the reference implementation, so a correct edit may still receive a lower score when it is realized through different code locations.
Similarly, MAE and SSIM compare the resulting rendering against a specific reference output and can penalize visually different but still valid implementations.
The strong Editing-Judge scores therefore suggest that \methodname{} reliably completes the requested modifications, even when its implementation and rendering differ from the reference.
\begin{table}[htp]
\centering
\footnotesize
\input{Tables/appendix/designbench}
\end{table}

\paragraph{Design2Code.}
Table~\ref{tab:design2code} provides a finer-grained view of the visual gains on Design2Code.
\modelnine reaches 90.09 in Block, 95.95 in Text, 89.30 in Position, 90.00 in Color, and 92.37 in CLIP, yielding an average Visual score of 91.54.
Relative to VinciCoder-8B, the largest gains appear in Position, Color, and Block, which improve by 15.40, 13.00, and 10.68 points, respectively, whereas the improvements in Text and CLIP are smaller.
Under the Design2Code evaluation, these dimensions respectively capture spatial alignment of matched elements, perceptual color fidelity, and the coverage of visual blocks while penalizing missing or hallucinated elements.
The results therefore indicate that \methodname{}'s advantage comes primarily from recovering a more complete set of webpage elements and placing them more accurately, together with improved color matching, rather than simply reproducing the visible text.
\modelthirtyfive exhibits a very similar metric profile, with only minor differences between the two model sizes across the evaluated dimensions.
\begin{table}[htp]
\centering
\footnotesize
\begin{minipage}[t]{0.535\linewidth}
\centering
\input{Tables/appendix/design2code}
\end{minipage}
\end{table}

\subsection{Artifacts and Demos}

\paragraph{ArtifactsBench.}
Table~\ref{tab:artifactsbench-full} presents the overall and domain-level performance on ArtifactsBench. The results show that our models generalize well across the heterogeneous artifact types in ArtifactsBench. Rather than being concentrated in a single domain, \modelnine maintains strong performance across all five categories and achieves the highest overall score among the open models, with particularly strong results on SVG generation and web applications. \modelthirtyfive reaches a similar overall level but shows a different strength profile, with especially strong performance on game development and simulation. This contrast suggests that the two models are capable of handling different forms of front-end complexity, from visual and interface construction to artifacts that require richer interaction and application logic. More importantly, both models consistently outperform the other open baselines and the proprietary baseline at the overall level, despite ArtifactsBench spanning substantially different artifact structures. This cross-domain consistency indicates that their advantage is not tied to a specific artifact format, but reflects stronger general front-end code generation and the ability to turn complex specifications into executable artifacts.
\begin{table}[htp]
\centering
\footnotesize
\input{Tables/appendix/artifactsbench}
\end{table}

\paragraph{InteractScience.}
Table~\ref{tab:interactscience} shows full metrics on InteractScience for both PFT and VQT lenses. The results highlight the complementary strengths of our models captured by the two lenses of InteractScience. PFT measures whether a model can correctly implement the interactive components specified in a long implementation plan, while VQT places greater emphasis on integrating visual understanding, executable code, and scientific reasoning. Our models perform particularly strongly under this setting. \modelnine and \modelthirtyfive both achieve substantially higher VQT scores than the other open models, while also maintaining competitive PFT performance. This gap is important because success on VQT cannot be obtained simply by producing syntactically valid front-end code. The generated science demos must also reflect the intended visual state and the scientific relationships encoded in the task. The results therefore point to a useful property of our models whose capability is not limited to implementing isolated UI interactions, but extends to coordinating code, visual requirements, and scientific content when following a detailed implementation plan.
\begin{table}[htp]
\centering
\footnotesize
\begin{minipage}[t]{0.6\linewidth}
\centering
\input{Tables/appendix/interactscience}
\end{minipage}
\end{table}

\section{Extended Analysis}
\label{app:extended}

\subsection{Backbone Generalization}
\label{app:backbone}
\begin{figure}[H]
\centering
\includegraphics[width=\linewidth]{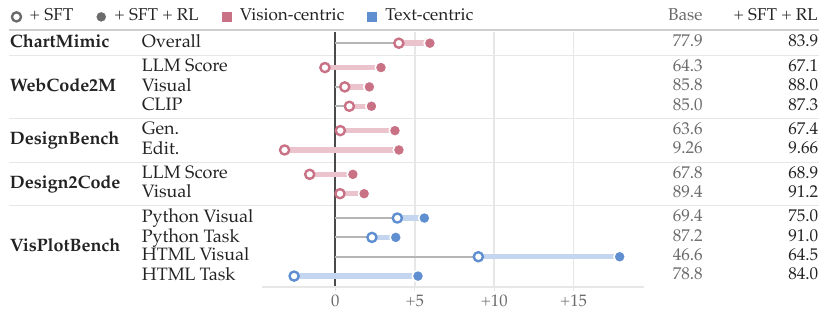}
\caption{SFT and RL gains on \gemmaicon Gemma-4-31B-it, with base and final scores.}
\label{fig:gemma-stages}
\end{figure}

\Cref{fig:gemma-stages} restates the Gemma arm of \cref{fig:backbone-stages} as
dumbbells, with the base and final score of every metric. The dumbbells express every
metric on a percentage scale, so the change in the 0--10 DesignBench Edit score is
plotted $\times 10$ while its score columns keep the native scale.
\Cref{tab:backbone-gemma} scores
the VisPlotBench splits under two judges, the benchmark's designated
GPT-4.1~\citep{openai2025gpt41} and the newer GPT-5.5~\citep{openai2026gpt55}. The two
agree on the direction of every training effect. The one disagreement is the HTML task
score, where GPT-4.1 ranks the RL checkpoint first and GPT-5.5 the SFT one. 
\input{Tables/appendix/backbone_gemma}

\subsection{Cross-Family Adjudication and Human Evaluation}
\label{app:adjudication}
\Cref{tab:human-study} lists the agreement statistics of the human study of
\cref{subsec:reliability}. The 100 samples come from the SFT split of \ourdata{},
sampled by difficulty level with Python and HTML artifacts in equal numbers across
tasks. Twenty annotators each rated ten samples directly and ten others through the
rubric, with disjoint sample sets per annotator, so each sample carries two human
ratings under each mode. Each of the three judges, Claude Sonnet
5~\citep{anthropic2026sonnet5}, Gemini 3.5 Flash~\citep{gemini2026gemini35flash}, and
GPT-5.5~\citep{openai2026gpt55}, scored every sample twice under each mode. The
direct mode asks for a single holistic score, as in a typical
pipeline~\citep{sun2026januscoder}; the rubric mode scores through the \pipename{}
rubric of \cref{subsec:rubric}. Krippendorff's $\alpha$ and Kendall's $\tau_b$ are
computed between each judge's scores and the human ratings under the same mode. Across
the three judges, $\alpha$ rises from 0.30--0.51 under direct scoring to 0.86--0.92
under the rubric, and $\tau_b$ from 0.28--0.44 to 0.64--0.67.
\input{Tables/appendix/human_study}

\section{Case Studies}
\makeatletter\setlength{\@fptop}{0pt}\setlength{\@fpsep}{10pt plus 4pt}\setlength{\@fpbot}{0pt plus 1fil}\makeatother
\label{app:cases}
\raggedbottom

\begin{figure}[htbp]
\centering
\casepanel[\scriptsize]{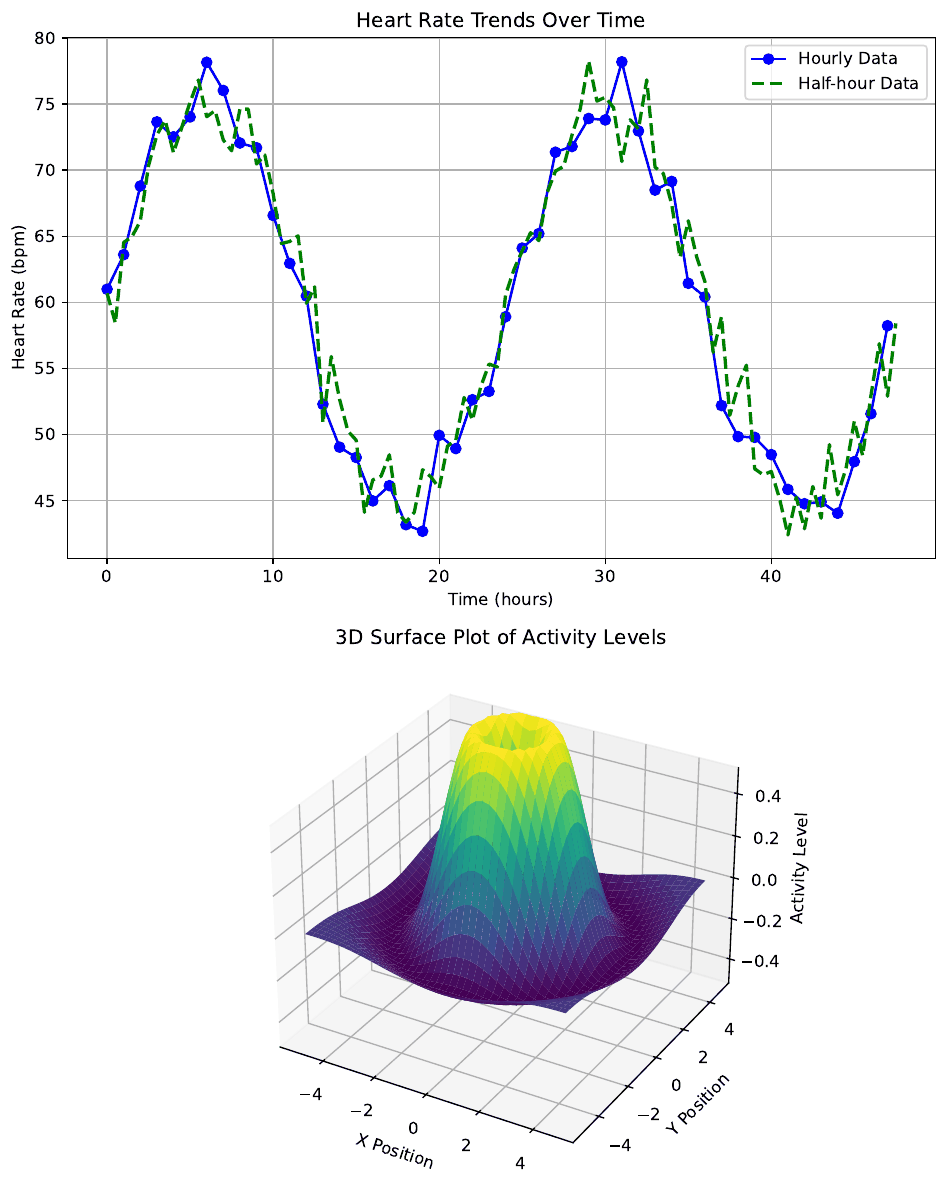}{Gold}{0.235\linewidth}\hfill
\casepanel[\scriptsize]{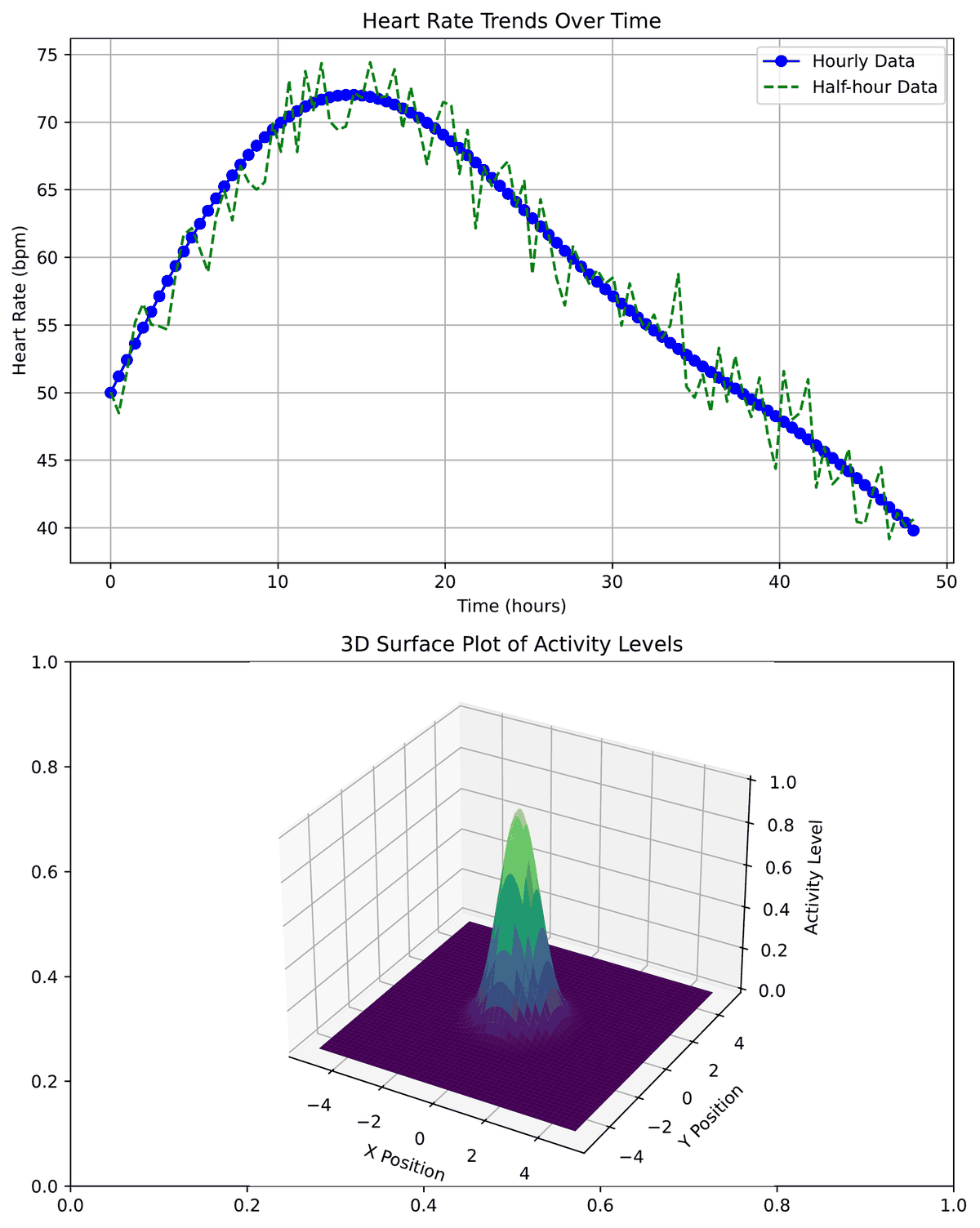}{Qwen3.5-9B}{0.235\linewidth}\hfill
\casepanel[\scriptsize]{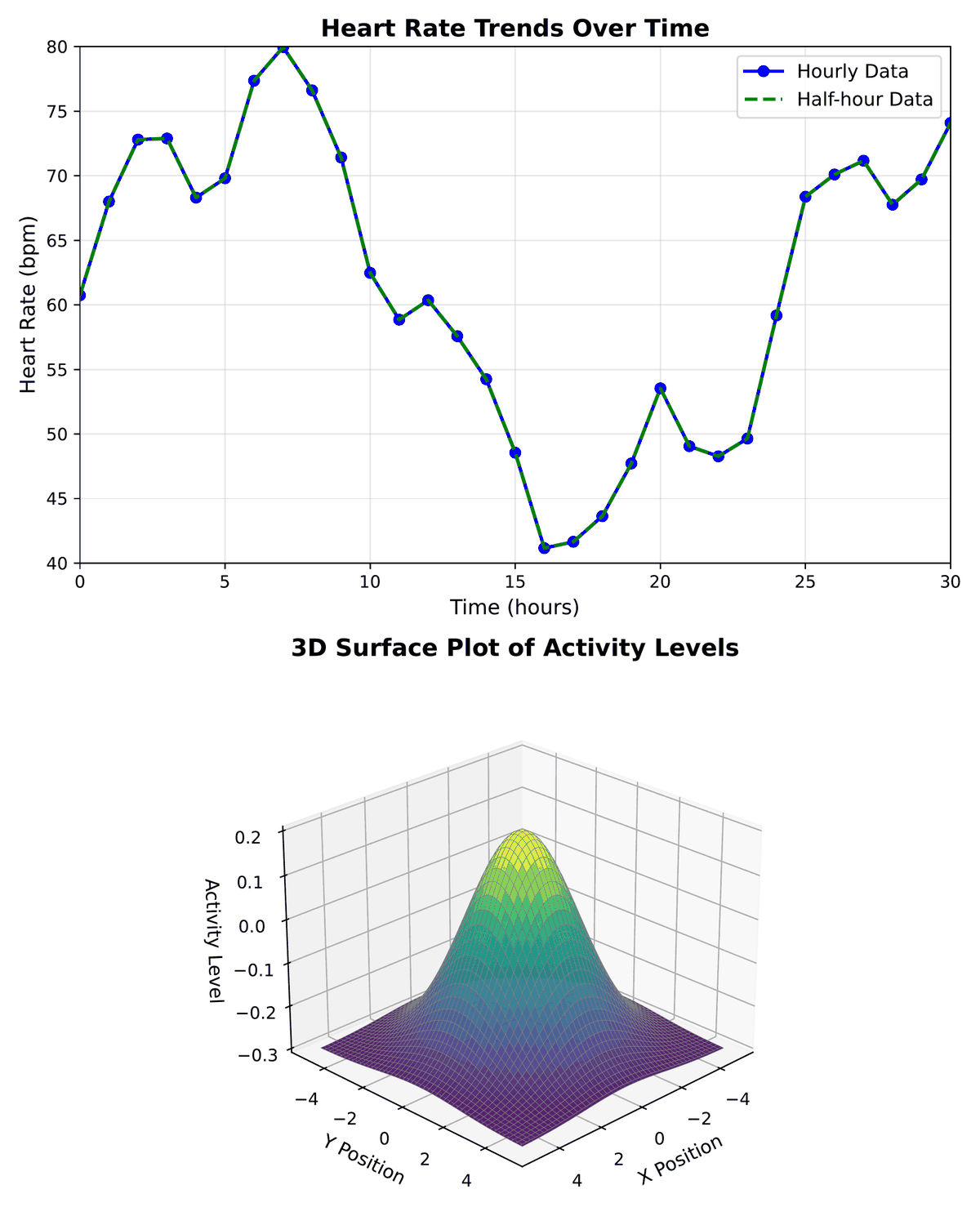}{\modelnine{} w/o RL}{0.235\linewidth}\hfill
\casepanel[\scriptsize]{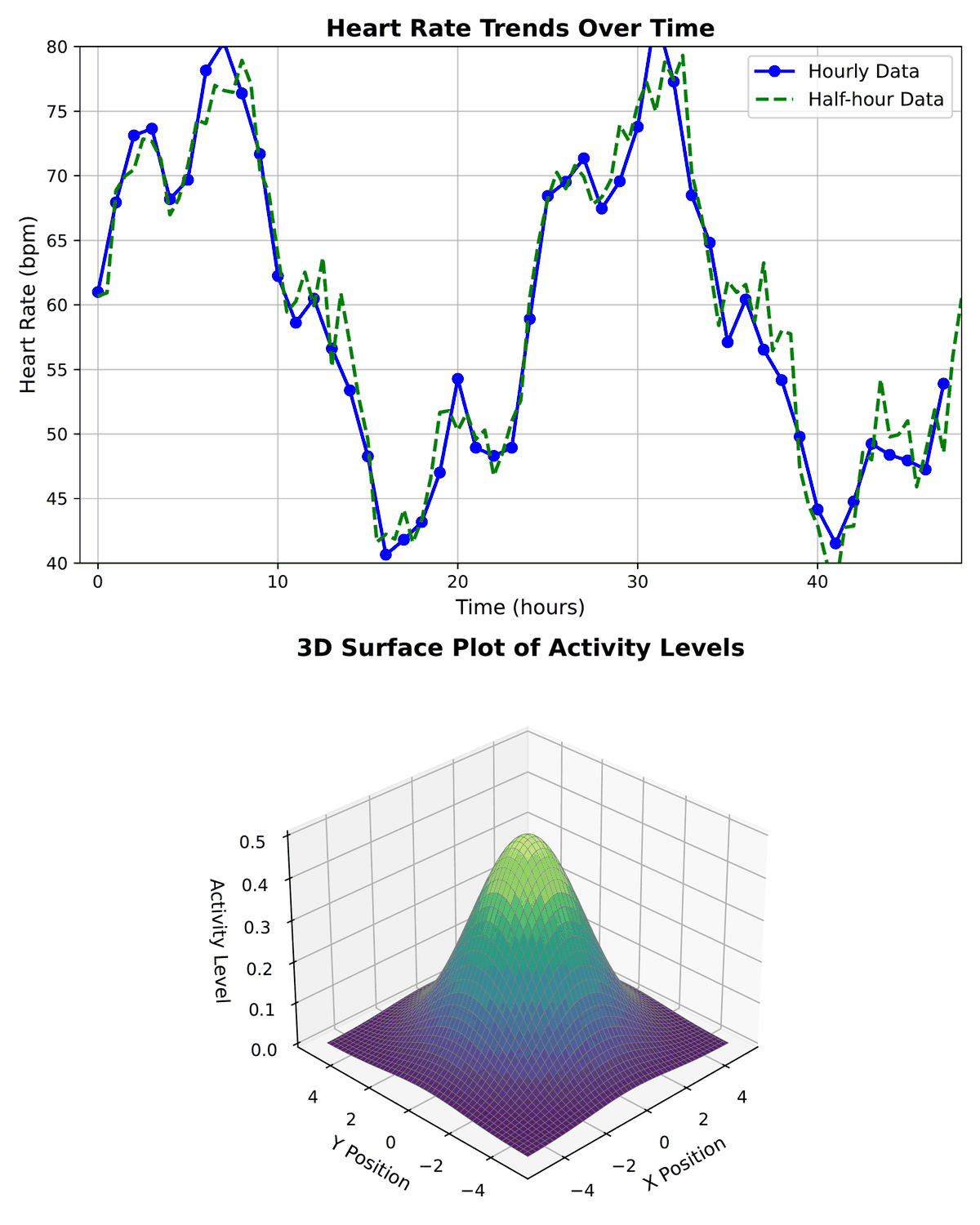}{\modelnine{}}{0.235\linewidth}
\caption{Chart-to-code on ChartMimic across training stages, where SFT recovers one of
the line chart's two cycles and RL restores both.}
\label{fig:case-chartmimic-stages}
\end{figure}

\begin{figure}[htbp]
\centering
\casepanel{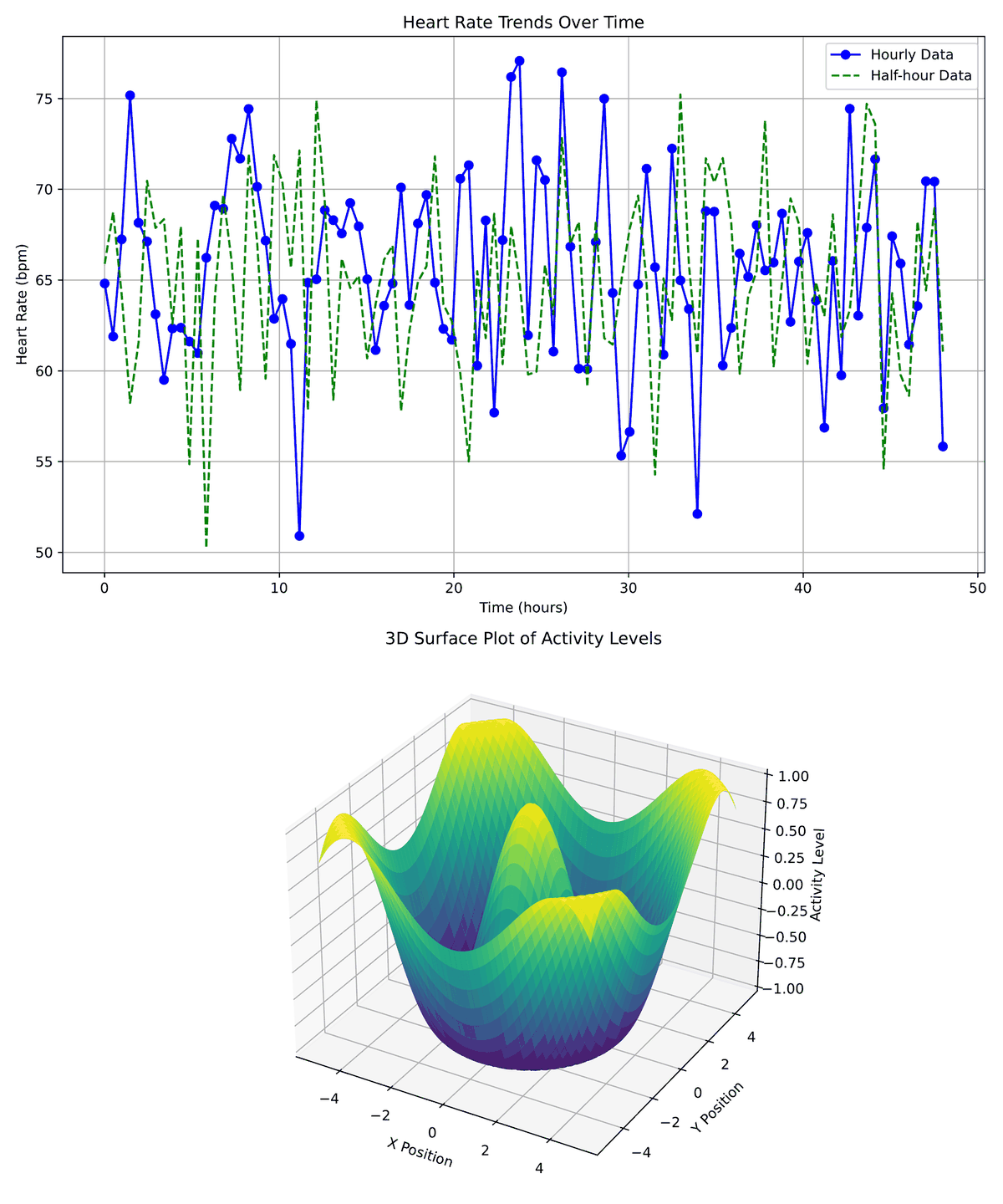}{VinciCoder-7B}{0.32\linewidth}\hfill
\casepanel{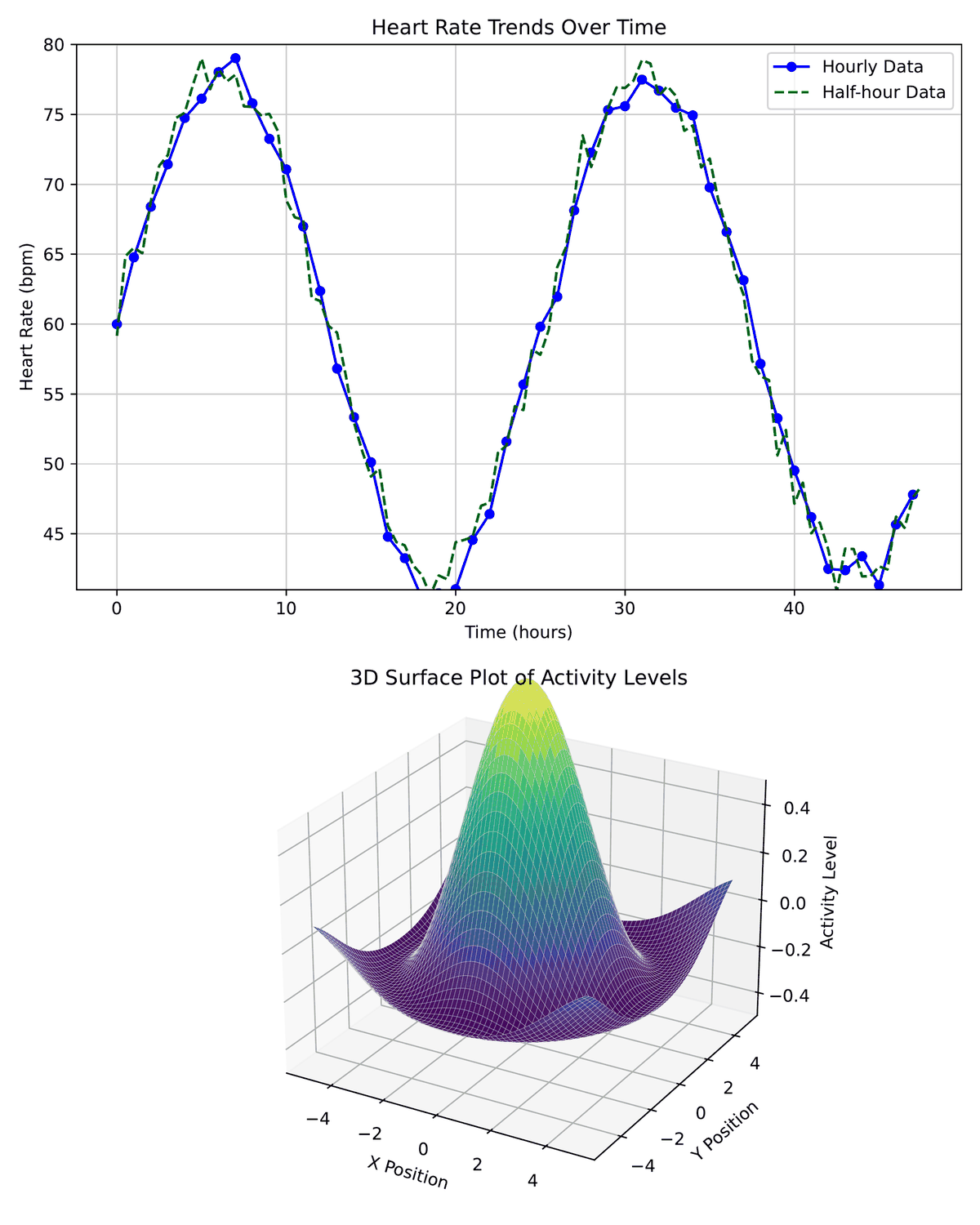}{Claude Opus 5}{0.32\linewidth}\hfill
\casepanel{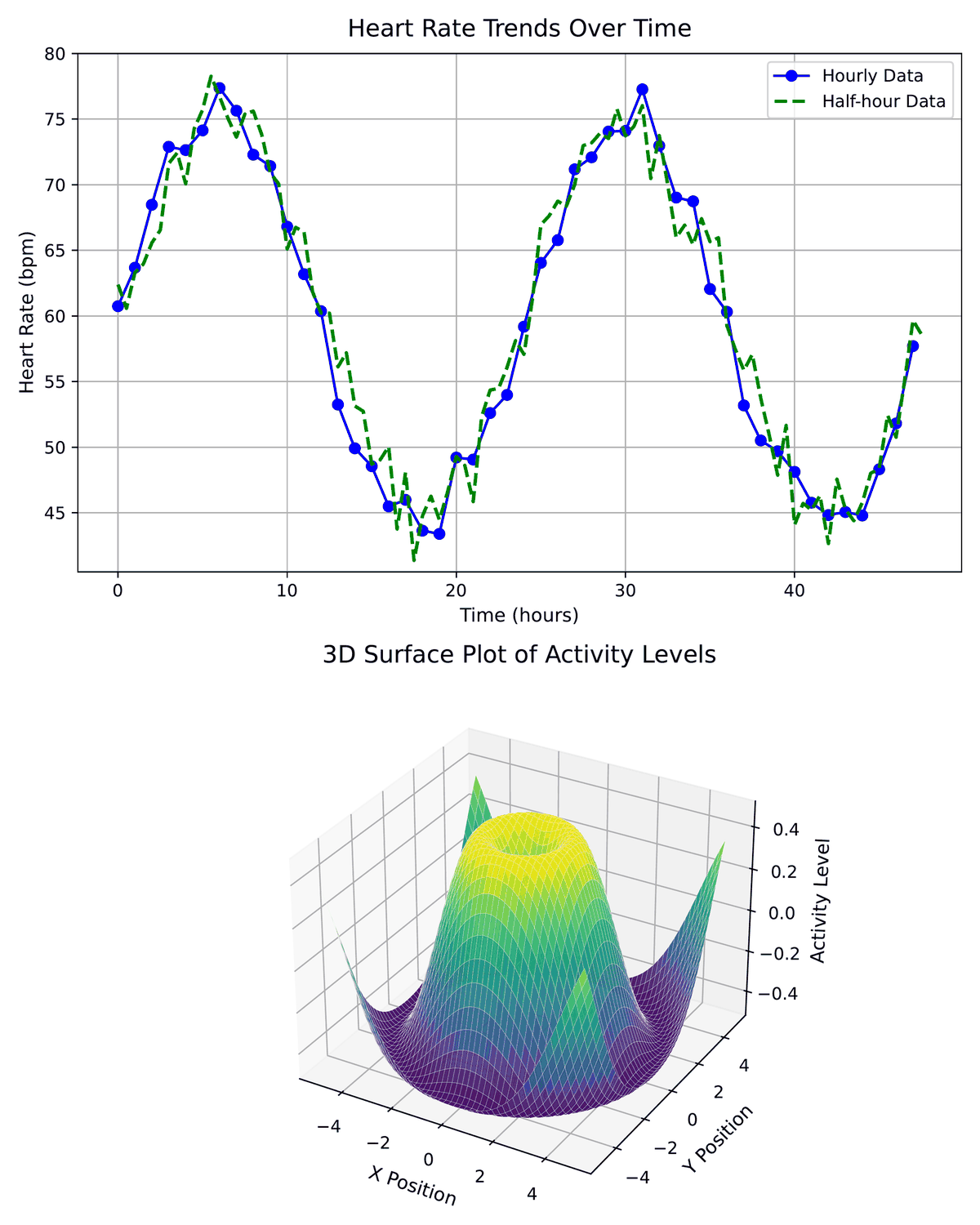}{Gemini 3.5 Flash}{0.32\linewidth}
\caption{The same case for an open and two proprietary references, of which only the
proprietary models reproduce the line chart.}
\label{fig:case-chartmimic-refs}
\end{figure}

\begin{figure}[htbp]
\centering
\casepanel{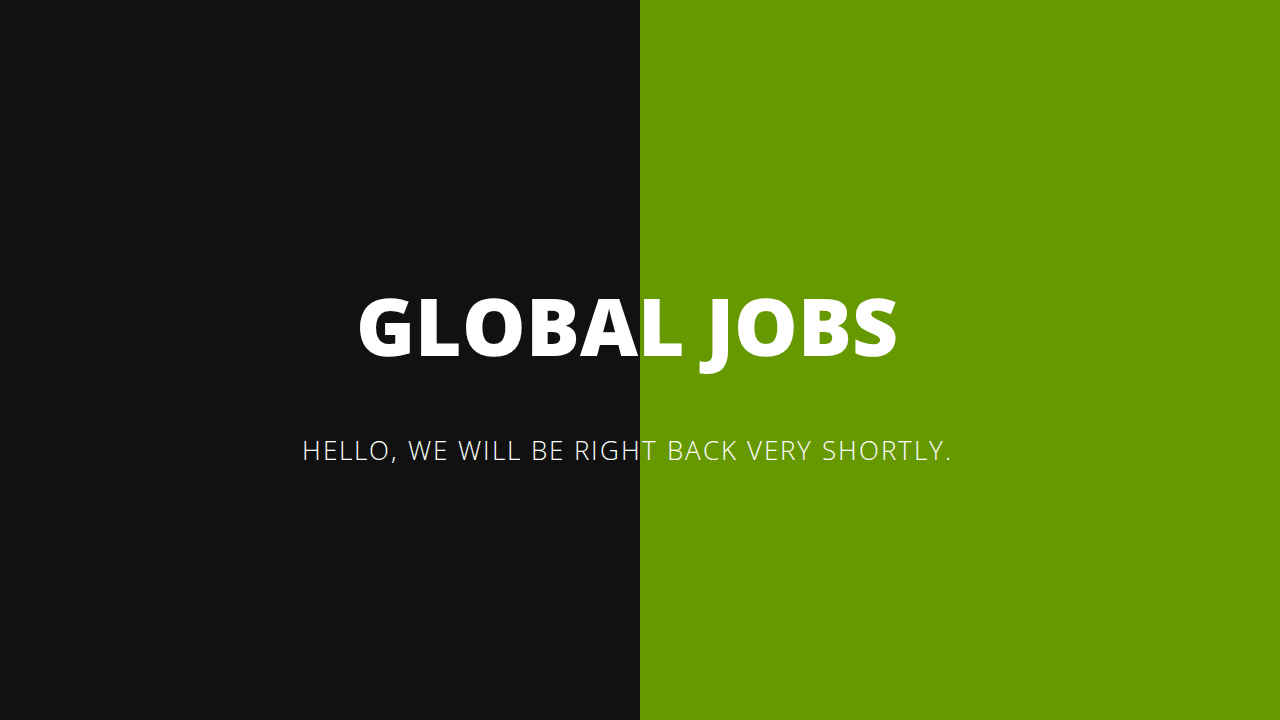}{Gold}{0.48\linewidth}\hfill
\casepanel{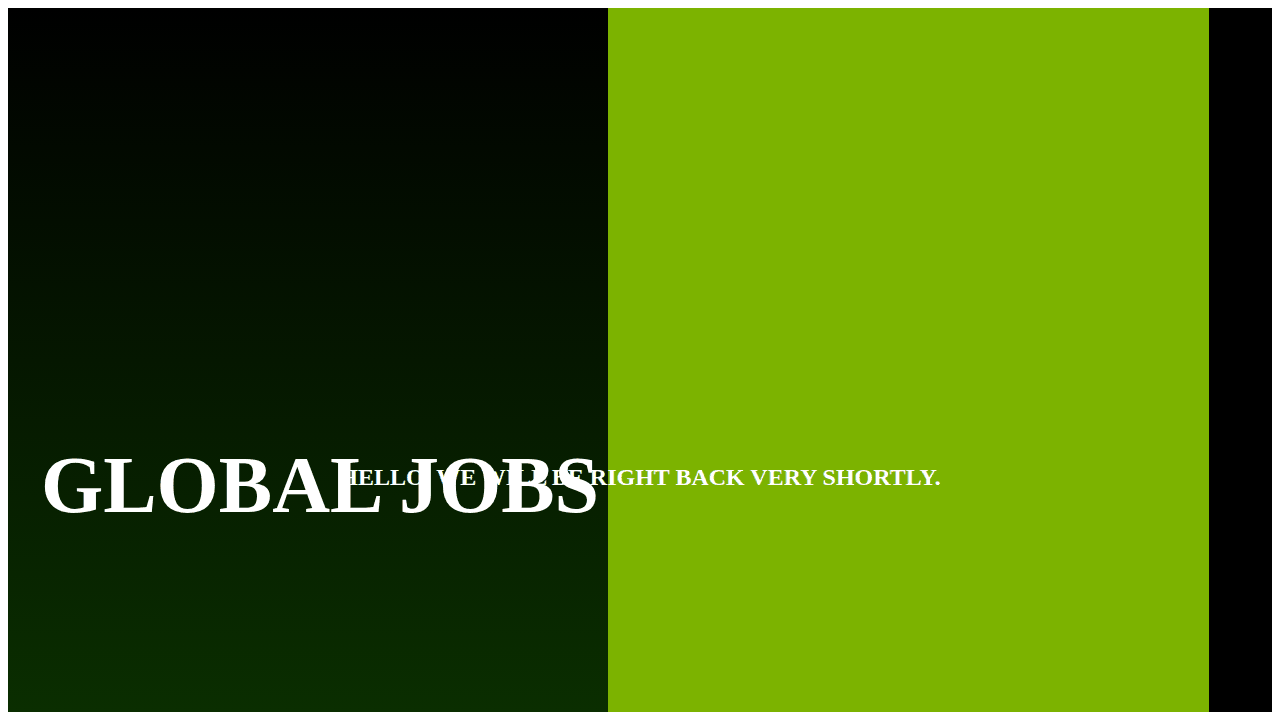}{Qwen3.5-9B}{0.48\linewidth}\\[6pt]
\casepanel{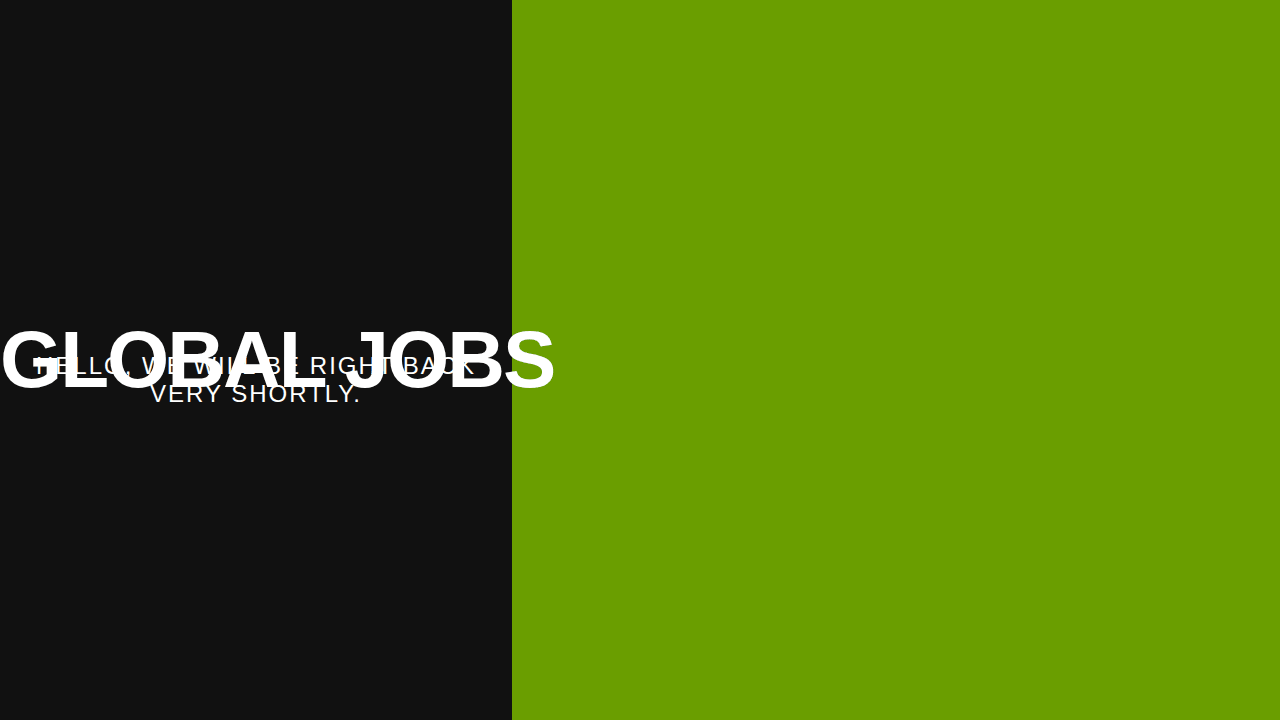}{\modelnine{} w/o RL}{0.48\linewidth}\hfill
\casepanel{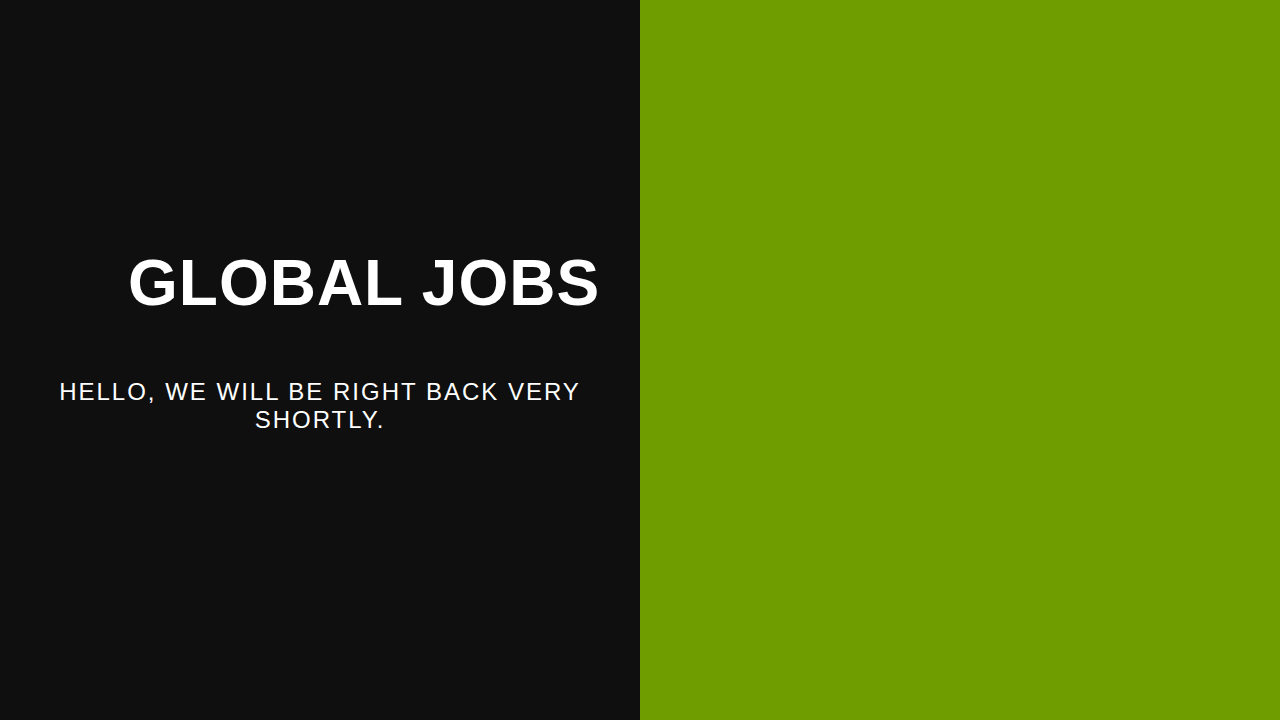}{\modelnine{}}{0.48\linewidth}
\caption{Front-end generation on Design2Code, where the base and SFT-only models break
the title layout and \modelnine{} reproduces the page.}
\label{fig:case-web-gen-9b}
\end{figure}

\begin{figure}[htbp]
\centering
\casepanelh{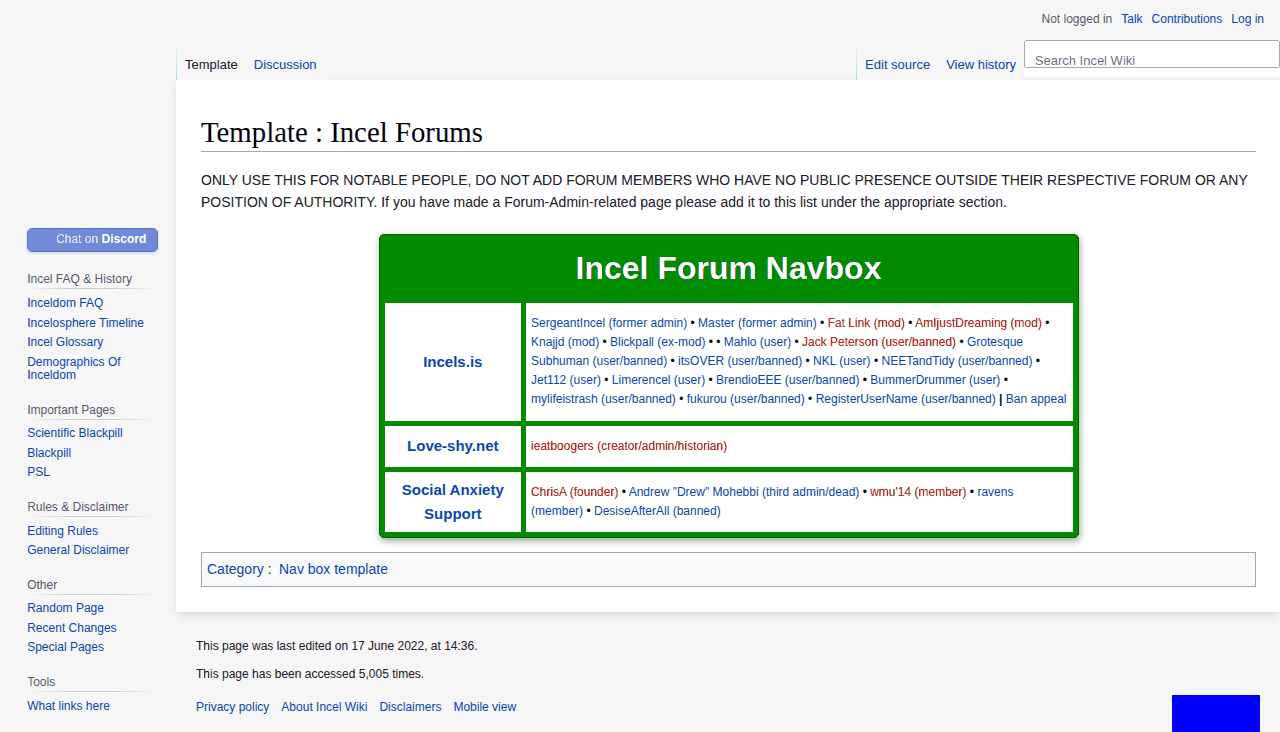}{Gold}{0.48\linewidth}{1.7in}\hfill
\casepanelh{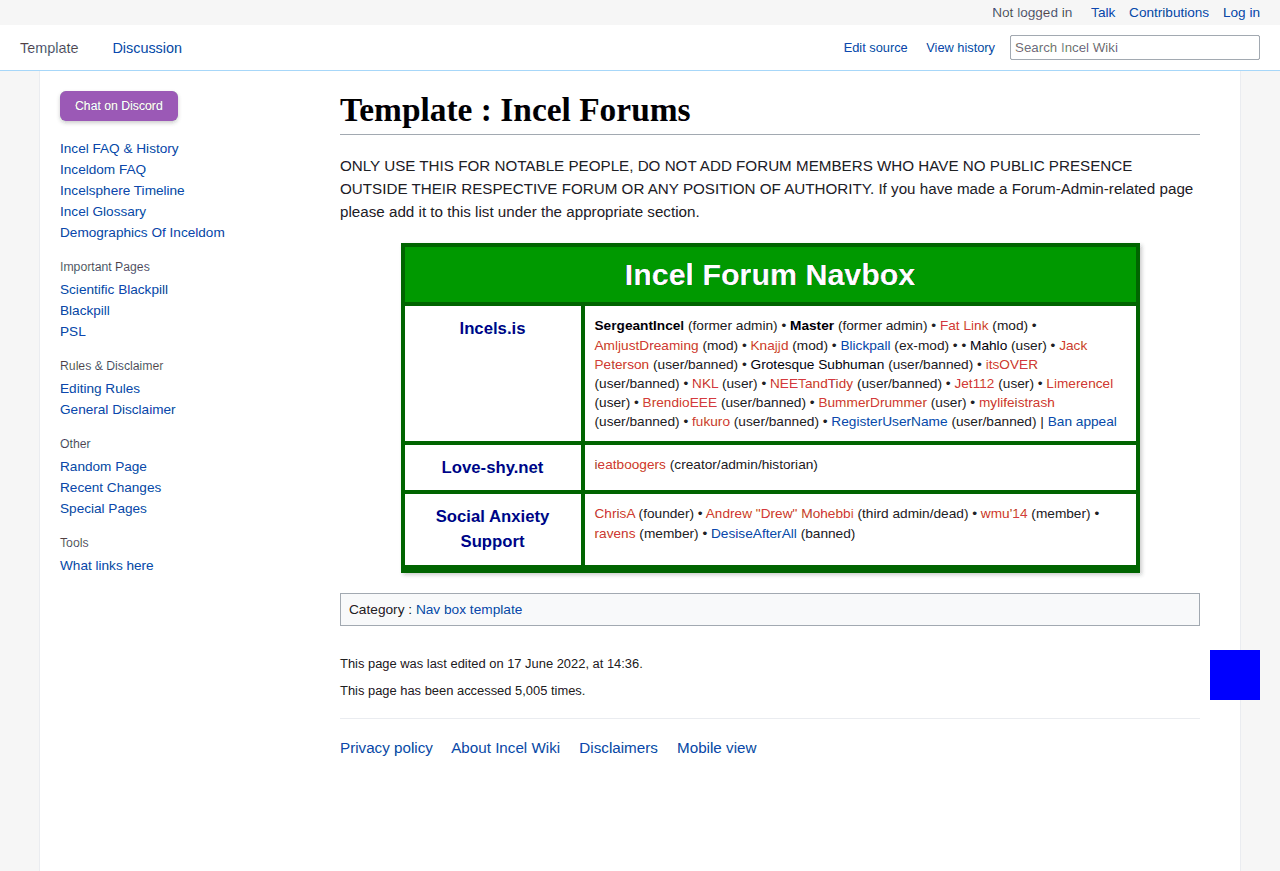}{Qwen3.5-35B-A3B}{0.48\linewidth}{1.7in}\\[6pt]
\casepanelh{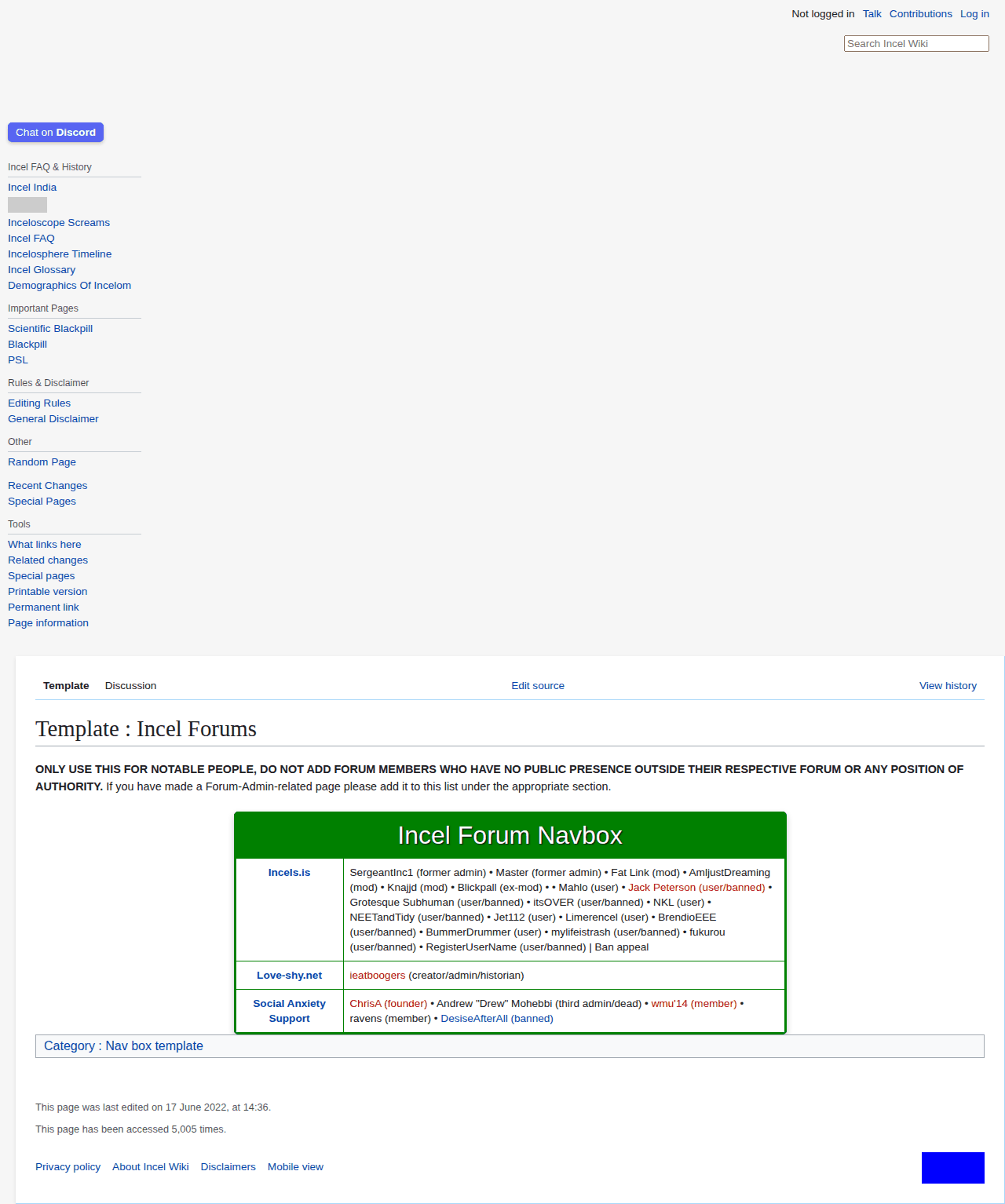}{\modelthirtyfive{} w/o RL}{0.48\linewidth}{1.7in}\hfill
\casepanelh{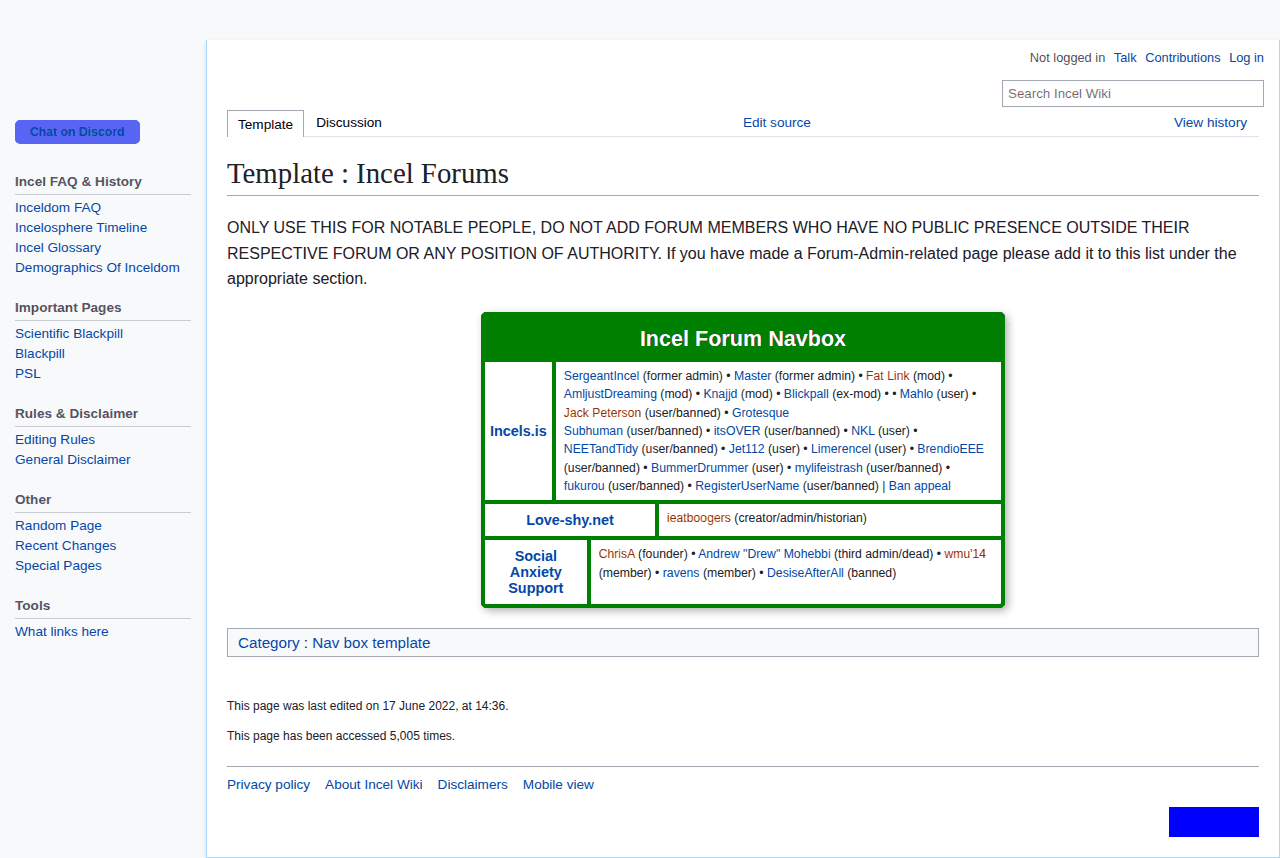}{\modelthirtyfive{}}{0.48\linewidth}{1.7in}
\caption{Front-end generation on Design2Code, where only \modelthirtyfive{} reproduces
the layout.}
\label{fig:case-web-gen-35b}
\end{figure}

\begin{figure}[htbp]
\centering
\casepanel{}{Source}{0.32\linewidth}\hspace{0.03\linewidth}
\casepanel{}{Target}{0.32\linewidth}\\[6pt]
\casepanel{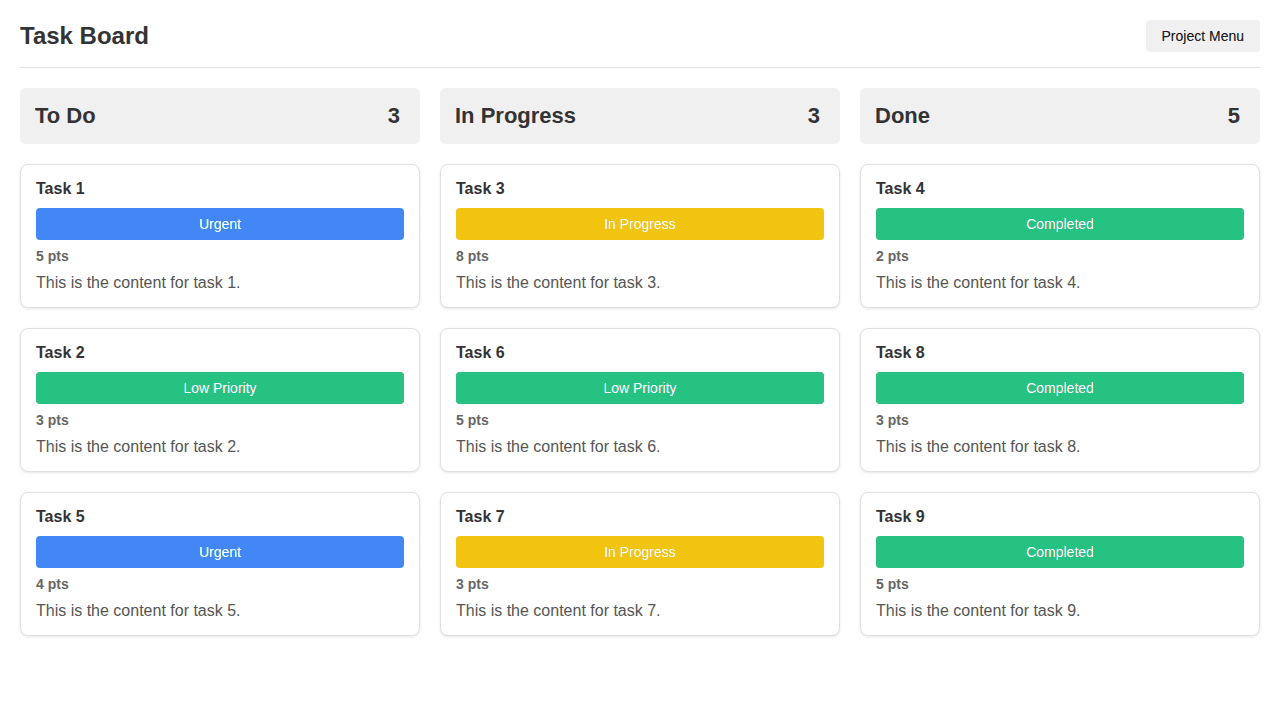}{Qwen3.5-9B}{0.32\linewidth}\hfill
\casepanel{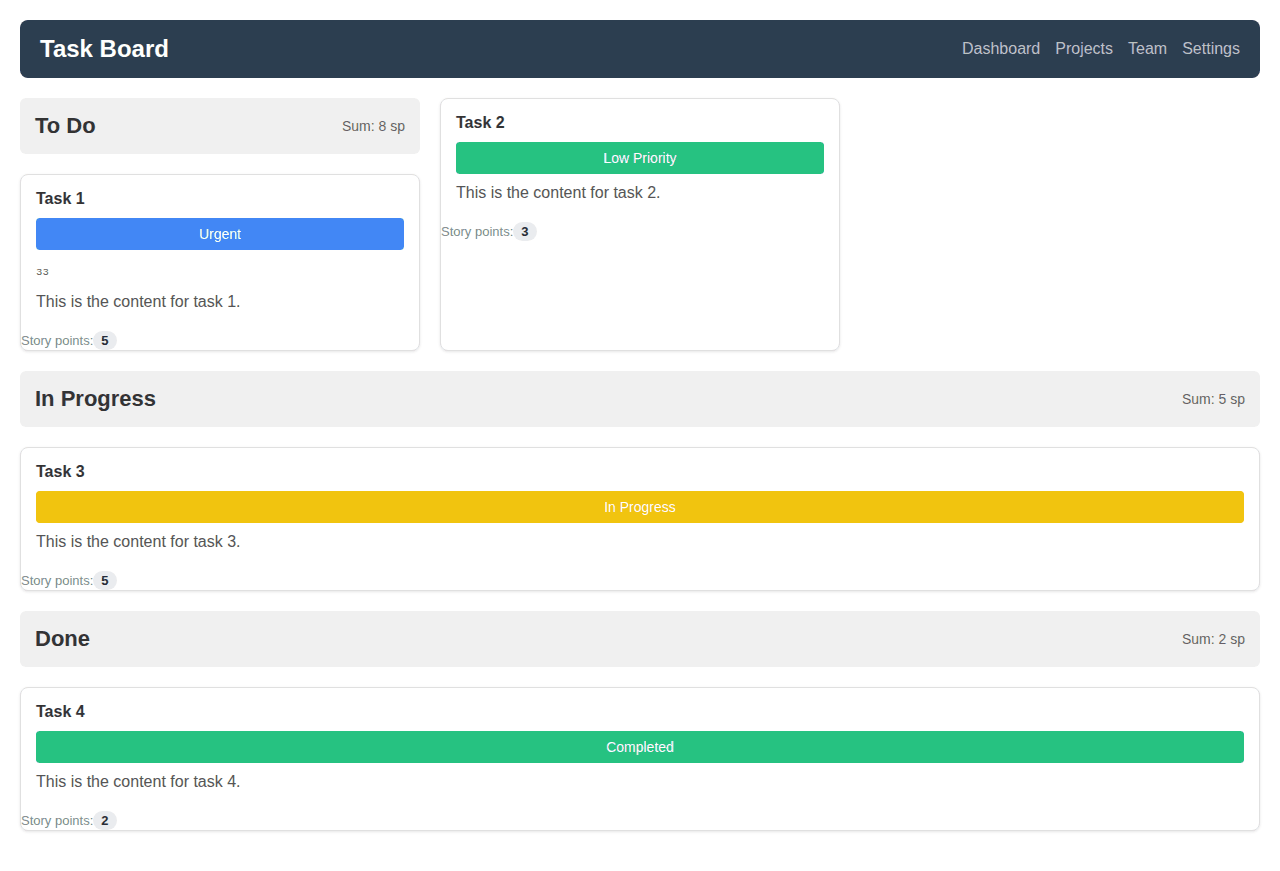}{\modelnine{} w/o RL}{0.32\linewidth}\hfill
\casepanel{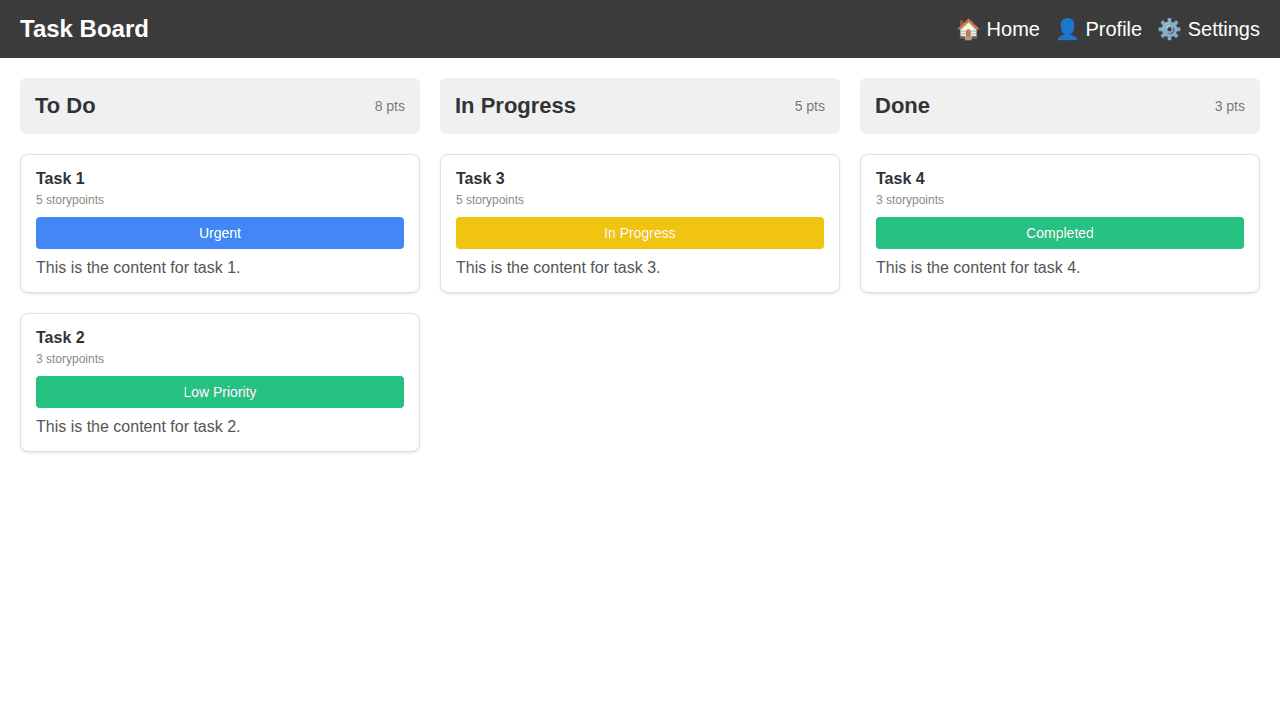}{\modelnine{}}{0.32\linewidth}
\caption{Front-end editing on DesignBench, where only \modelnine{} adds the requested story
points, per-column sums, and header menu while keeping the board.}
\label{fig:case-web-edit-9b}
\end{figure}

\begin{figure}[htbp]
\centering
\casepanel{}{Source}{0.32\linewidth}\hspace{0.03\linewidth}
\casepanel{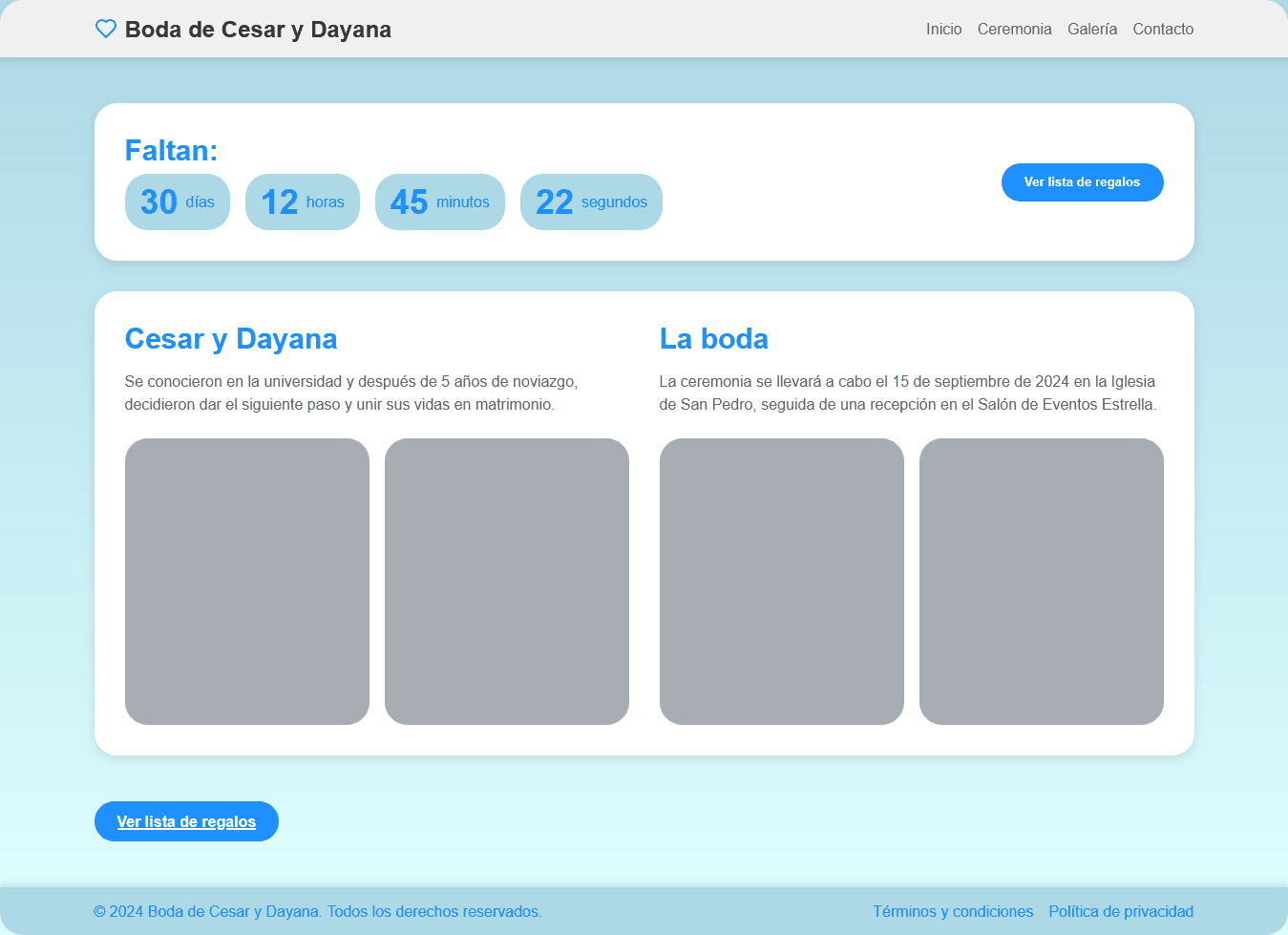}{Target}{0.32\linewidth}\\[6pt]
\casepanel{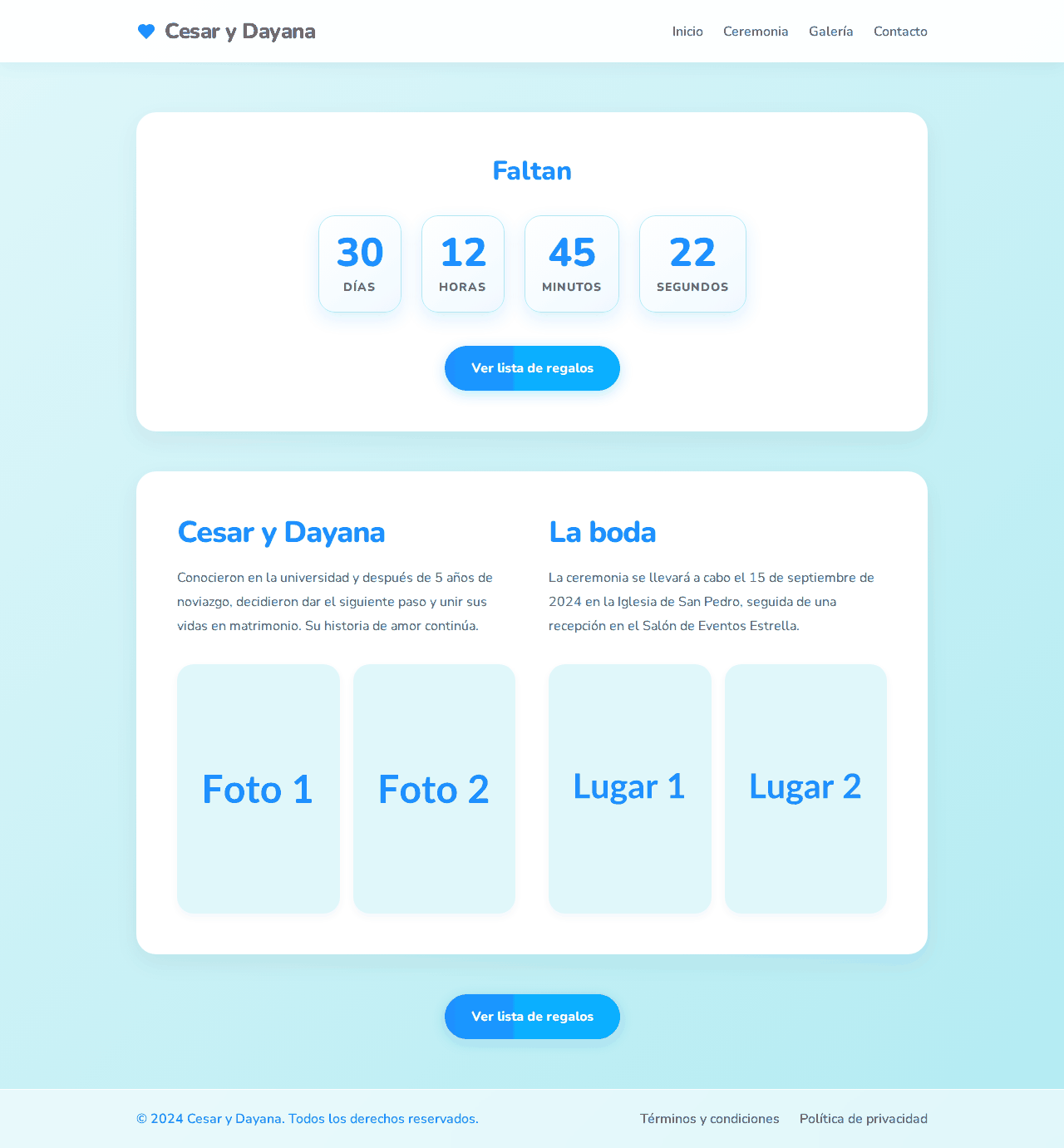}{Qwen3.5-35B-A3B}{0.32\linewidth}\hfill
\casepanel{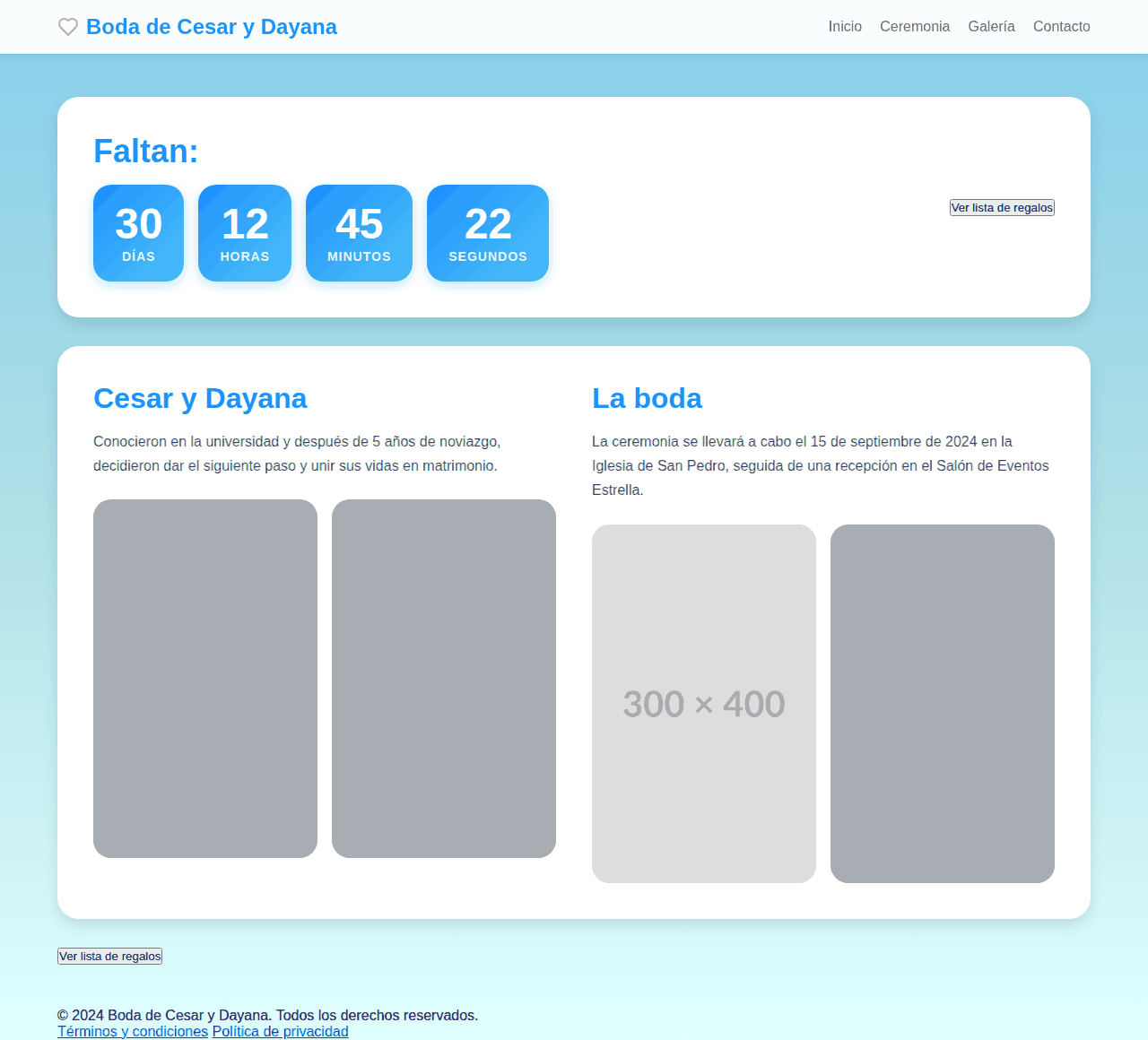}{\modelthirtyfive{} w/o RL}{0.32\linewidth}\hfill
\casepanel{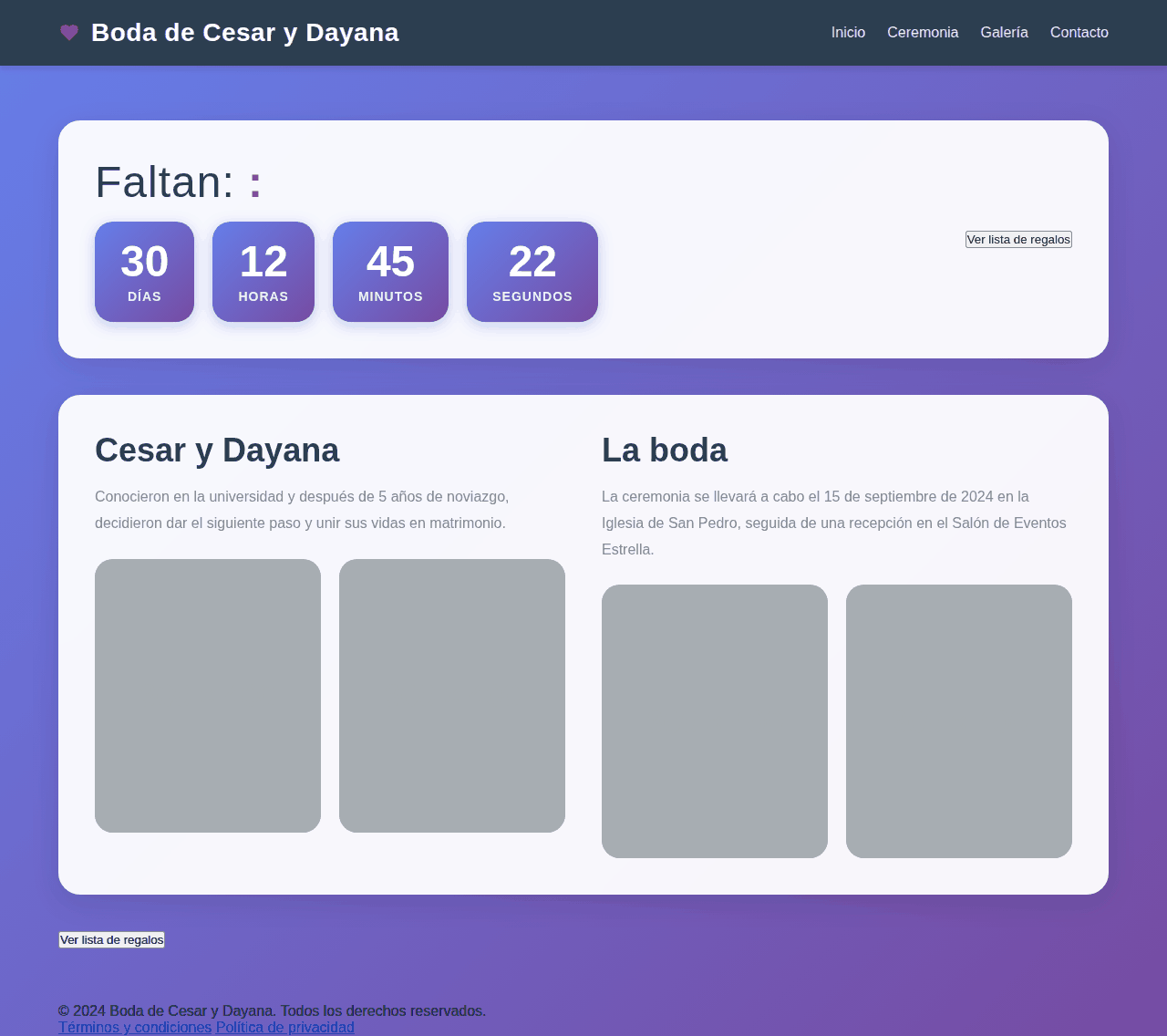}{\modelthirtyfive{}}{0.32\linewidth}
\caption{Front-end editing on DesignBench, where only \modelthirtyfive{} applies the requested
renaming, rounded corners, and countdown restyling.}
\label{fig:case-web-edit-35b}
\end{figure}

Six qualitative cases drawn from VisPlotBench, one per format family. Three
compare \modelnine{} against its Qwen3.5-9B base (\LaTeX{}, SVG, Mermaid); three
compare \modelthirtyfive{} against its SFT-only variant (Vega-Lite, HTML,
Python), isolating what the rubric-based RL stage adds. In every case the
middle panel executes and renders, and the failure is visual. A seventh case,
chart-to-code on ChartMimic, follows \modelnine{} across its training stages
(\cref{fig:case-chartmimic-stages}) and sets the result beside an open and two
proprietary references (\cref{fig:case-chartmimic-refs}). Four front-end cases
close the appendix, two screenshot-to-page generations from Design2Code and two
instruction-driven edits from DesignBench, each across the same training stages
(\cref{fig:case-web-gen-9b,fig:case-web-gen-35b,fig:case-web-edit-9b,fig:case-web-edit-35b}).

\casetriptych{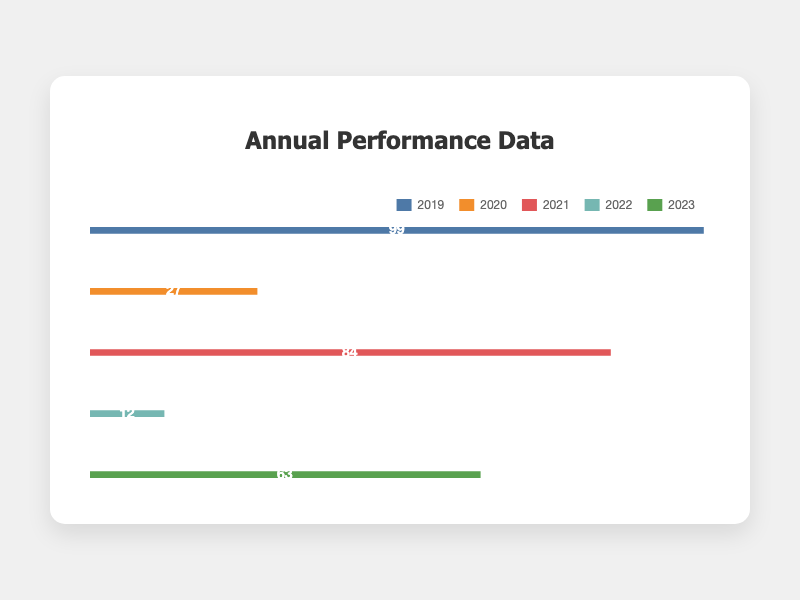}{PrismaCoder-35B w/o RL}{\modelthirtyfive{}}
{HTML bar chart. The SFT-only model renders the bars as hairlines; the RL model
recovers bar geometry and value labels.}
{fig:case-html}

\casetriptych{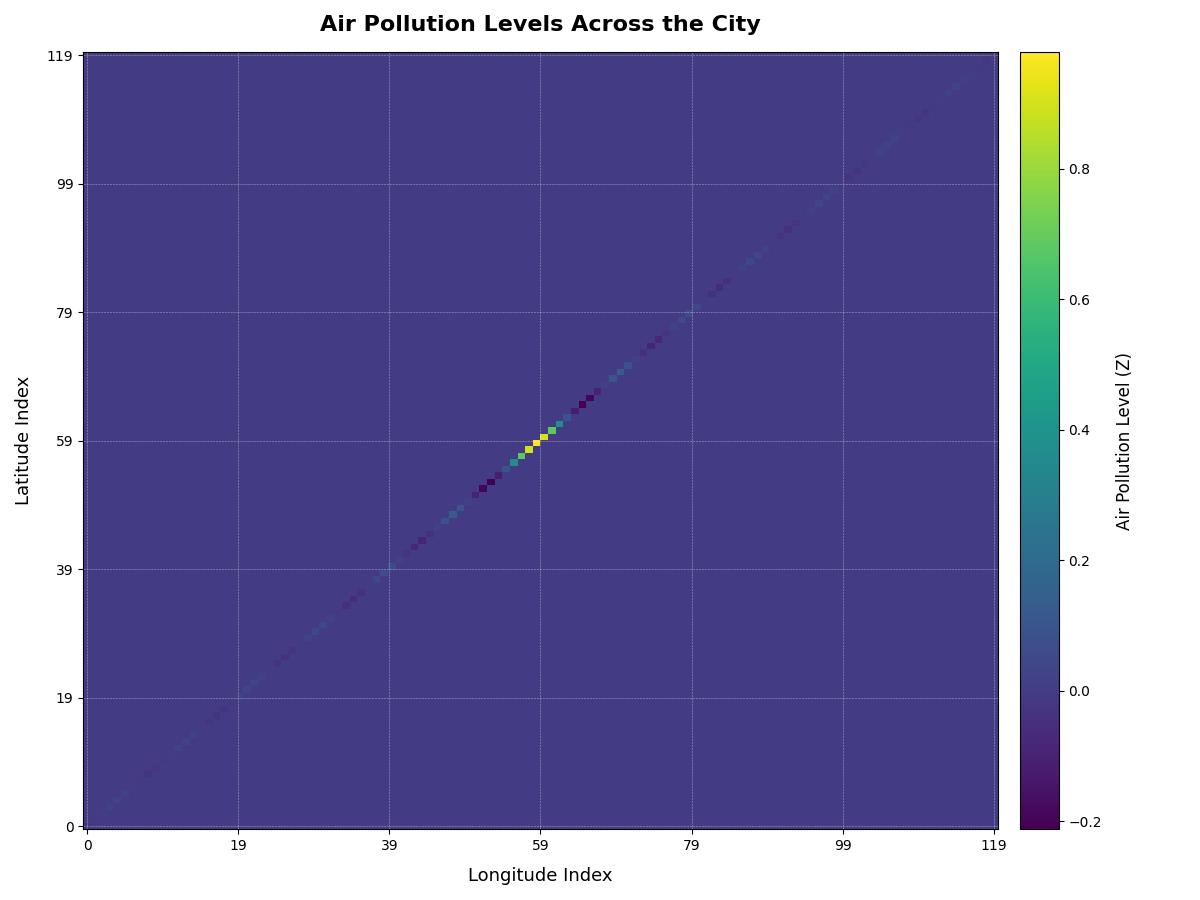}{PrismaCoder-35B w/o RL}{\modelthirtyfive{}}
{Python heatmap. The SFT-only output runs and passes as code yet renders the
radial field onto a diagonal; the RL model restores it.}
{fig:case-python}

\casetriptych{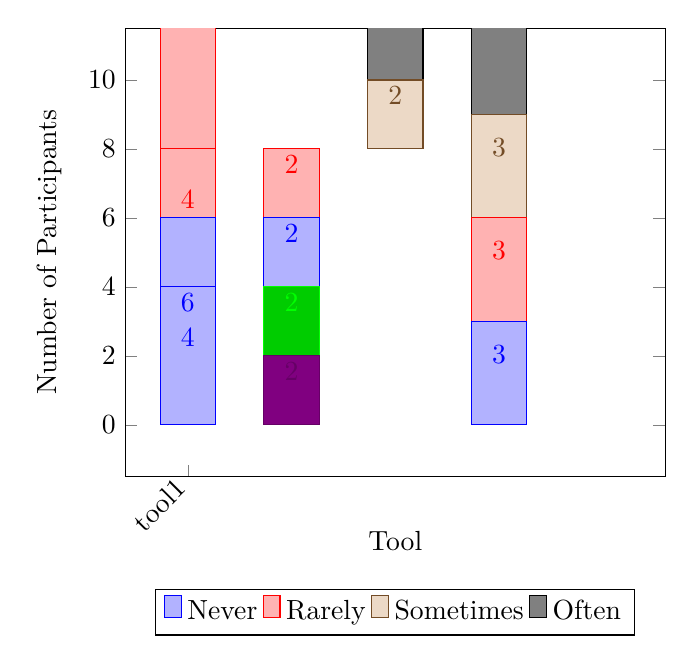}{Qwen3.5-9B}{\modelnine{}}
{\LaTeX{} stacked bars. The base model mis-stacks the categories and drops
bars; \modelnine{} reproduces the per-tool composition.}
{fig:case-latex}

\casetriptych{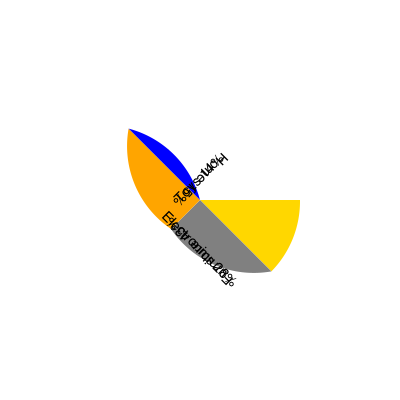}{Qwen3.5-9B}{\modelnine{}}
{SVG pie chart. The base model's sectors collapse into overlapping wedges;
\modelnine{} closes the pie with correct shares and labels.}
{fig:case-svg}

\casetriptych{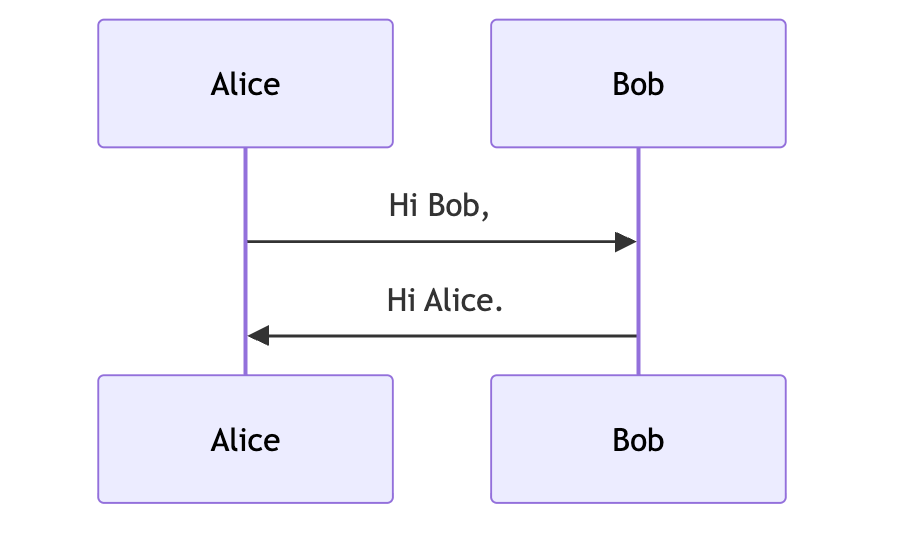}{Qwen3.5-9B}{\modelnine{}}
{Mermaid sequence diagram. The base model draws participants as plain boxes;
\modelnine{} reproduces the actor figures and the message order.}
{fig:case-mermaid}

%% file: Tables/appendix/data-source.tex
\begin{table}[htbp]
\centering
\caption{Data sources and routing statistics.}
\label{tab:data_source_stat}
\resizebox{\linewidth}{!}{
\begin{tabular}{lrrr}
\toprule
\textbf{Data Source} &
\textbf{\# Seed Samples} &
\textbf{\# Regenerated Samples} &
\textbf{\# Synthesized Samples} \\
\midrule
MCD dataset
    & 598,000 & 307,184 & 273,199 \\
GitHub repositories
    & 417,000 & 297,942 & 188,630 \\
WebCode2M (Generate)
    & 243,100 & 182,300 & 215,663 \\
WebCode2M (Edit)
    & 264,541 & 167,810 & 190,882 \\
JanusCoder-800K
    & 109,826 & 80,000 & 29,000 \\
WebSight
    & 1,922,671 & 40,000 & 12,000 \\
ProImageBench
    & 654 & 600 & 80 \\
MMMG
    & 4456 & 2400 & 975 \\
UniSVG
    & 528,591 & 65,100 & 51,100 \\
OmniSVG
    & 255,412 & 37,000 & 10,000 \\
arXiv papers
    & 67,000 & 60,155 & 21,400 \\
PubChem
    & 60,000 & 60,000 & 59,000 \\
Manim video code
    & 41,707 & 25,440 & 16,300 \\
\bottomrule
\end{tabular}
}
\end{table}

%% file: Tables/comparison.tex
\newcommand{\nst}{--}
\begin{table}[t]
\centering
\footnotesize
\setlength{\tabcolsep}{4pt}
\renewcommand{\arraystretch}{1.08}
\caption{Coverage of open multimodal code-intelligence works. \cmark{} = present in the
released training corpus as its paper describes it; \xmark{} = absent from the paper;
\nst{} = not stated.}
\label{tab:comparison}
\resizebox{\linewidth}{!}{%
\begin{tabular}{l >{\columncolor{boxblue}}c c c c c}
\toprule
& \textbf{\ourdata} & JanusCoder & VisCodex & VisCoder2 & VinciCoder \\
\midrule
\multicolumn{6}{l}{\emph{Visual code data types}} \\
Python plotting (text to plot)      & \cmark & \cmark & \nst   & \cmark & \xmark \\
Web UI (screenshot or text to page) & \cmark & \cmark & \cmark & \xmark & \cmark \\
Chart-to-code (image to code)       & \cmark & \cmark & \cmark & \xmark & \cmark \\
SVG                                 & \cmark & \cmark & \xmark & \cmark & \cmark \\
LaTeX / TikZ                        & \cmark & \xmark & \xmark & \cmark & \cmark \\
Mermaid                             & \cmark & \xmark & \xmark & \cmark & \cmark \\
Asymptote                           & \cmark & \xmark & \xmark & \cmark & \cmark \\
LilyPond                            & \cmark & \xmark & \xmark & \cmark & \xmark \\
Vega-Lite                           & \cmark & \xmark & \xmark & \cmark & \xmark \\
JavaScript                          & \cmark & \nst   & \xmark & \cmark & \xmark \\
Code-driven animation (Manim)       & \cmark & \cmark & \xmark & \xmark & \xmark \\
Interactive demos and artifacts     & \cmark & \cmark & \xmark & \xmark & \xmark \\
Chemical-structure code             & \cmark & \xmark & \xmark & \xmark & \cmark \\
Other plotting languages (R, MATLAB, C++, TS) & \xmark & \cmark & \xmark & \cmark & \xmark \\
Wolfram / Mathematica               & \xmark & \cmark & \xmark & \xmark & \xmark \\
\# visual data types covered (of 15) & \textbf{13} & 8 & 2 & 9 & 7 \\
\midrule
\multicolumn{6}{l}{\emph{Other data}} \\
General (non-visual) code           & \xmark & \cmark & \cmark & \cmark & \xmark \\
\midrule
\multicolumn{6}{l}{\emph{Artifacts and method}} \\
Released corpus                     & 2M & 800K & 598K & 679K & 1.6M \\
Open-weight models                  & 9B, 35B-A3B & 7B, 8B, 14B & \nst & 3B to 32B & 7B, 8B \\
Curation signal                     & code + vision rubrics & exec. + judge score & exec. + judge score & exec. + image checks & exec. + linters \\
Per-instruction rubric              & \cmark & \xmark & \xmark & \xmark & \xmark \\
RL stage                            & GRPO, rubric reward & \xmark & \xmark & \xmark & GRPO, visual similarity \\
Rubric as reward                    & \cmark & \xmark & \xmark & \xmark & \xmark \\
Human-agreement study of the score  & \cmark & \cmark & \xmark & \xmark & \xmark \\
\bottomrule
\end{tabular}}
\end{table}

%% file: Tables/appendix/reward_spec.tex
\begin{table}[h]
\centering
\footnotesize
\caption{RL rewards by task family. Rewards lie in $[0,1]$;
unscoreable completions receive $0$.}
\label{tab:reward-spec}
\setlength{\tabcolsep}{4pt}
\begin{tabular}{l l l l}
\toprule
Task family & Spec.\ $\to$ output & Reward Type & Weights Assignment \\
\midrule
Chart generation & text $\to$ Python & Rubrics & $\lambda_c=\lambda_v = 0.5$ \\
Chart reproduction & image $\to$ Python & Rubrics & $\lambda_c=\lambda_v = 0.5$ \\
Webpage generation & text $\to$ HTML & Rubrics & $\lambda_c=0.4, \lambda_v = 0.6$ \\
HTML demonstration & plan $\to$ HTML & Rubrics & $\lambda_c=0.4, \lambda_v = 0.6$ \\
\quad + reference snapshots & plan + images $\to$ HTML & Rubrics & $\lambda_c=0.4, \lambda_v = 0.6$ \\
SVG generation & text $\to$ SVG & Rubrics & $\lambda_c=0.4, \lambda_v = 0.6$ \\
Interface reproduction & image $\to$ HTML & Rubrics+Holistic & $\lambda_c=0.2,\lambda_v = 0.5, \lambda_h = 0.3$ \\
SVG reproduction & image $\to$ SVG & Rubrics+Holistic & $\lambda_c=0.2,\lambda_v = 0.5, \lambda_h = 0.3$ \\
Interface editing & page + instr.\ $\to$ HTML & Rubrics+Holistic & $\lambda_c=0.2,\lambda_v = 0.4, \lambda_h = 0.4$ \\
\bottomrule
\end{tabular}
\end{table}

%% file: Tables/appendix/chartmimic.tex
\setlength{\tabcolsep}{3pt}
\caption{ChartMimic, full metrics for the Direct-1800 splits.}
\label{tab:chartmimic-full}

\begin{tabular}{l cccccccc}
\toprule
& \multicolumn{8}{c}{Direct} \\
\cmidrule(lr){2-9}
Model
& \makecell{Exec.\\Rate}
& Text
& Layout
& Type
& Color
& \makecell{Avg.\\(L)}
& \makecell{GPT\\(H)}
& Overall \\
\midrule

\multicolumn{9}{l}{\textit{Open models}} \\

JanusCoder-7B
& 89.16 & 77.61 & 85.49 & 70.06 & 67.34 & 75.12 & 68.97 & 72.05 \\

JanusCoder-8B
& 80.91 & 73.10 & 81.19 & 65.00 & 69.03 & 76.54 & 69.23 & 72.89 \\

VinciCoder-7B
& 90.88 & 79.91 & 87.99 & 73.72 & 71.59 & 78.30 & 66.09 & 72.20 \\

VinciCoder-8B
& 90.72 & 79.78 & 86.62 & 73.03 & 70.95 & 77.59 & 66.44 & 72.01 \\

\rowcolor{boxblue}
\textbf{\modelnine}
& 95.50
& 88.78
& 92.77
& 74.87
& 69.88
& 81.57
& 73.82
& 77.70 \\

\rowcolor{boxblue}
\textbf{\modelthirtyfive}
& \textbf{96.16}
& \textbf{90.41}
& \textbf{94.00}
& \textbf{76.28}
& \textbf{73.01}
& \textbf{83.42}
& \textbf{80.69}
& \textbf{82.05} \\

\midrule
\multicolumn{9}{l}{\textit{Proprietary}} \\

\rowcolor[gray]{0.92}
\gptfouro
& 92.38
& 81.13
& 86.71
& 72.83
& 64.87
& 76.38
& 72.02
& 74.20 \\

\rowcolor[gray]{0.92}
Gemini 3 Flash
& 95.61
& 90.49
& 92.46
& 78.66
& 78.86
& 85.12
& 83.04
& 84.08 \\

\rowcolor[gray]{0.92}
Gemini 3.1 Pro 
& 95.77
& 81.68
& 82.84
& 73.33
& 73.98
& 77.96
& 81.73
& 79.85 \\

\rowcolor[gray]{0.92} 
Claude Sonnet 4.6 
& 97.11
& 91.89
& 94.17
& 68.53
& 66.35
& 80.23
& 80.73
& 80.48 \\

\rowcolor[gray]{0.92}
Claude Opus 4.6 
& 95.05
& 90.20
& 92.01
& 73.10
& 69.13
& 81.11
& 79.68
& 80.40 \\

\bottomrule
\end{tabular}

%% file: Tables/appendix/pandasplotbench.tex
\setlength{\tabcolsep}{4pt}
\caption{PandasPlotBench, full metrics under the Visual and Task lenses.
}
\label{tab:pandasplot-full}

\begin{tabular}{l c cc cc}
\toprule
& \makecell{Incorr.\\Code$\downarrow$}
& \multicolumn{2}{c}{Mean}
& \multicolumn{2}{c}{Good $\geq$75} \\
\cmidrule(lr){3-4}\cmidrule(lr){5-6}
Model & & Vis. & Task & Vis. & Task \\
\midrule

Qwen3-VL-8B-Instruct
    & 16.00\% & 60.01 & 73.67 & 0.506 & 0.701 \\
Qwen3-VL-30B-A3B-Instruct
    & 11.43\% & 60.05 & 80.58 & 0.555 & 0.806 \\

Qwen3.5-9B
    & 24.57\% & 52.98 & 65.47 & 0.417 & 0.640 \\
Qwen3.5-35B-A3B
    & 15.43\% & 60.36 & 75.17 & 0.509 & 0.743 \\

JanusCoderV-7B
    & 23.43\% & 53.05 & 65.77 & 0.440 & 0.640 \\
JanusCoderV-8B
    & 26.30\% & 56.30 & 70.91 & 0.450 & 0.650 \\

VinciCoder-7B
    & 9.14\% & 25.37 & 51.57 & 0.137 & 0.314 \\
VisCoder2-7B
    & 20.00\% & 59.48 & 67.59 & 0.511 & 0.623 \\

\rowcolor{boxblue}
\textbf{PrismaCoder-9B}
    & 10.29\%
    & 60.54
    & 77.94
    & 0.497
    & 0.766 \\

\rowcolor{boxblue}
\textbf{PrismaCoder-35B}
    & 9.29\% & {62.38} & 79.85 & 0.606 & 0.772 \\

\midrule

\rowcolor[gray]{0.92}
GPT-4o
    & 8.00\% & 67.63 & 82.46 & 0.571 & 0.822 \\
\rowcolor[gray]{0.92}
Gemini 3 Flash
    & 3.43\% & 69.95 & 91.21 & 0.566 & 0.949 \\
\rowcolor[gray]{0.92}
Gemini 3.1 Pro
    & 1.71\% & 78.41 & 94.70 & 0.703 & 0.960 \\
\rowcolor[gray]{0.92}
Claude Sonnet 4.6
    & 3.43\% & 74.59 & 91.93 & 0.686 & 0.914 \\
\rowcolor[gray]{0.92}
Claude Opus 4.6
    & 1.71\% & 76.03 & 94.97 & 0.674 & 0.966 \\

\bottomrule
\end{tabular}

%% file: Tables/appendix/visplotbench_avg.tex
\setlength{\tabcolsep}{4pt}
\caption{VisPlotBench averaged metrics.}
\label{tab:visplot-avg}
\resizebox{\linewidth}{!}{%
\begin{tabular}{l c cc cc}
\toprule
& \makecell{Incorr.\\Code$\downarrow$} & \multicolumn{2}{c}{Mean} & \multicolumn{2}{c}{Good $\geq$75} \\
\cmidrule(lr){3-4}\cmidrule(lr){5-6}
Model & & Vis. & Task & Vis. & Task \\
\midrule
VinciCoder-7B    & 35.90\% & 15.93 & 23.95 & 4.16 & 10.11 \\
VinciCoder-8B    & 36.08\% & 8.84 & 12.55 & 1.50 & 2.80 \\
\rowcolor{boxblue} \textbf{\modelnine}       & 34.70\% & 36.40 & 53.99 & 23.12 & 52.62 \\
\rowcolor{boxblue} \textbf{\modelthirtyfive} & 28.86\% & 45.16 & 62.47 & 32.55 & 61.49 \\
\rowcolor[gray]{0.92} \gptfouro & 32.01\% & 38.19 & 56.01 & 24.75 & 52.75 \\
\bottomrule
\end{tabular}}

%% file: Tables/appendix/unisvg.tex
\setlength{\tabcolsep}{4pt}
\caption{Full UniSVG metrics~\citep{li2025unisvg}.}
\label{tab:unisvg}
\resizebox{\linewidth}{!}{%
\begin{tabular}{l ccc}
\toprule
Model & ISVGen & TSVGen & Final \\
\midrule
JanusCoderV-7B & 61.60 & 66.43 & 61.23 \\
JanusCoderV-8B  & 62.47 & 67.28 & 63.47 \\
VinciCoder-7B & 85.72 & 64.06 & 67.22 \\
VinciCoder-8B & 87.18 & 63.45 & 67.58 \\
\rowcolor{boxblue} \textbf{\modelnine} & 69.64 & 68.49 & 65.31 \\
\rowcolor{boxblue} \textbf{\modelthirtyfive} & 84.14 & 70.72 & 72.96 \\
\rowcolor[gray]{0.92} \gptfouro & 70.98 & 72.80 & 69.60 \\
\bottomrule
\end{tabular}}

%% file: Tables/appendix/visplotbench_languages.tex
\setlength{\tabcolsep}{4pt}
\caption{VisPlotBench, per-language incorrect-code rate. }
\label{tab:visplot-full}
\resizebox{\linewidth}{!}{%
\begin{tabular}{l c cccccccc}
\toprule
\rowcolor{boxredlight}
Model & \makecell{Avg.\\Incorr.$\downarrow$}
& Python & \makecell{Vega-\\Lite} & LilyPond & Mermaid & SVG & \LaTeX & \makecell{Asym-\\ptote} & HTML \\
\midrule
\multicolumn{10}{l}{\emph{Open models}} \\
VinciCoder-7B    & 35.90\% & 10.20\% & 66.67\% & 100.00\% & 7.63\% & 1.54\% & 56.25\% & 21.74\% & 23.15\% \\
VinciCoder-8B    & 36.08\% & 2.55\% & 57.36\% & 100.00\% & 4.58\% & 7.69\% & 58.93\% & 19.57\% & 37.96\% \\
VisCoder2        & 33.26\% & 43.37\% & 16.28\% & 41.82\% & 25.95\% & 12.31\% & 61.61\% & 43.48\% & 21.30\% \\
\rowcolor{boxblue} \textbf{\modelnine}       & 34.70\% & 29.59\% & 13.18\% & 85.45\% & 12.21\% & 1.54\% & 66.96\% & 65.22\% & 19.44\% \\
\rowcolor{boxblue} \textbf{\modelthirtyfive} & 28.86\% & 11.22\% & 3.88\% & 89.09\% & 9.92\% & 0.00\% & 70.54\% & 51.09\% & 11.11\% \\
\midrule
\multicolumn{10}{l}{\emph{Proprietary}} \\
\rowcolor[gray]{0.92} \gptfouro & 32.01\% & 22.45\% & 8.53\% & 40.00\% & 24.43\% & 0.00\% & 61.61\% & 75.00\% & 24.07\% \\
\bottomrule
\end{tabular}}

%% file: Tables/appendix/chemdraw.tex
\setlength{\tabcolsep}{7pt}
\caption{ChemDraw, full metrics on the 128-sample test set. Tani.\,(mean) averages over
RDKit-valid outputs; the other columns count all samples. }
\label{tab:chemdraw}
\begin{tabular}{l ccccc}
\toprule
Model & Valid. & \makecell{Exact\\Match} & \makecell{Tani.\\(mean)} & \makecell{Tani.\\$=1$} & \makecell{Tani.\\$\geq.85$} \\
\midrule
\multicolumn{6}{l}{\textit{Open models}} \\
Qwen3-VL-8B-Instruct & 78.91\% & 25.00\% & 54.79 & 25.00\% & 25.00\% \\
Qwen3-VL-30B-A3B-Instruct & 82.03\% & 26.56\% & 54.55 & 27.34\% & 28.12\% \\
Qwen3.5-9B & 88.28\% & 33.59\% & 66.79 & 37.50\% & 39.84\% \\
Qwen3.5-35B-A3B & 92.97\% & 53.12\% & 79.86 & 57.81\% & 60.16\% \\
\rowcolor{boxblue} \textbf{\modelnine} & 96.88\% & 83.59\% & 95.94 & 87.50\% & 89.06\% \\
\rowcolor{boxblue} \textbf{\modelthirtyfive} & 100.00\% & 92.19\% & 97.56 & 95.31\% & 95.31\% \\
\midrule
\multicolumn{6}{l}{\textit{Proprietary}} \\
\rowcolor[gray]{0.92} \gptfouro & 92.19\% & 32.03\% & 57.80 & 33.59\% & 33.59\% \\
\rowcolor[gray]{0.92} Gemini 3 Flash & 63.28\% & 52.34\% & 81.39 & 53.12\% & 53.91\% \\
\rowcolor[gray]{0.92} Claude Sonnet 4.6 & 88.28\% & 53.91\% & 85.43 & 61.72\% & 64.06\% \\
\rowcolor[gray]{0.92} Claude Opus 4.6 & 96.88\% & 52.34\% & 77.71 & 56.25\% & 57.81\% \\
\rowcolor[gray]{0.92} Kimi K2.5 & 91.41\% & 65.62\% & 91.20 & 75.00\% & 76.56\% \\
\bottomrule
\end{tabular}

%% file: Tables/appendix/webcode2m.tex
\setlength{\tabcolsep}{3pt}
\caption{WebCode2M, full metrics by code length.}
\label{tab:webcode2m-full}
\resizebox{\linewidth}{!}{%
\begin{tabular}{l cccc cccc cccc cccc}
\toprule
& \multicolumn{4}{c}{Short} & \multicolumn{4}{c}{Mid} & \multicolumn{4}{c}{Long} & \multicolumn{4}{c}{Average} \\
\cmidrule(lr){2-5}\cmidrule(lr){6-9}\cmidrule(lr){10-13}\cmidrule(lr){14-17}
Model
& LLM & Visual & CLIP & \makecell{Tree-\\BLEU}
& LLM & Visual & CLIP & \makecell{Tree-\\BLEU}
& LLM & Visual & CLIP & \makecell{Tree-\\BLEU}
& LLM & Visual & CLIP & \makecell{Tree-\\BLEU} \\
\midrule
\multicolumn{17}{l}{\textit{Open models}} \\
JanusCoderV-7B & 40.10 & 72.39 & 76.11 & 29.64 & 43.06 & 74.08 & 76.77 & \underline{37.43} & 39.61 & 72.24 & 75.11 & \textbf{39.37} & 40.92 & 72.90 & 76.00 & \textbf{35.48}  \\
JanusCoderV-8B & 34.60 & 65.74 & 73.14 & \underline{30.37} & 36.17 & 65.41 & 73.46 & 29.88 & 34.74 & 66.71 & 72.53 & 27.66 & 35.17 & 65.95 & 73.04 & 29.30  \\
VinciCoder-7B & 46.01 & 76.76 & 80.43 & 30.00 & 44.34 & 73.57 & 79.38 & \textbf{38.11} & 38.99 & 66.92 & 75.97 & 29.74 & 43.11 & 72.42 & 78.59 & 32.62 \\
VinciCoder-8B & 48.35 & 80.43 & 82.03 & \textbf{31.39} & 48.08 & 79.53 & 81.70 & \textbf{39.48} & 42.05 & 74.83 & 78.63 & \underline{32.57} & 46.16 & 78.26 & 80.79 & \underline{34.48}  \\
\rowcolor{boxblue} \textbf{\modelnine} & \textbf{70.65} & \textbf{90.06} & \textbf{90.29} & 17.23 & \textbf{69.08} & \textbf{90.02} & \textbf{87.76} & 18.65 & \textbf{69.42} & \textbf{91.15} & \textbf{87.32} & 17.23 & \textbf{69.72} & \textbf{90.41} & \textbf{88.46} & 17.70  \\
\rowcolor{boxblue} \textbf{\modelthirtyfive} & \underline{68.52} & \underline{89.58} & \underline{89.93} & 15.78 & \underline{68.12} & \underline{89.54} & \underline{87.21} & 17.77 & \underline{66.69} & \underline{89.87} & \underline{86.19} & 17.18 & \underline{67.78} & \underline{89.66} & \underline{87.78} & 16.91  \\
\midrule
\multicolumn{17}{l}{\textit{Proprietary}} \\
\rowcolor[gray]{0.92} \gptfouro & 57.62 & 83.96 & 82.10 & 11.73 & 53.27 & 83.35 & 79.62 & 10.00 & 53.93 & 83.03 & 78.94 & 8.13 & 54.94 & 83.45 & 80.22 & 9.95  \\
\rowcolor[gray]{0.92} Gemini 3.1 Pro & 71.67 & 89.23 & 88.52 & 16.77 & 70.83 & 89.78 & 87.58 & 16.56 & 70.40 & 90.87 & 86.82 & 14.44 & 70.97 & 89.96 & 87.64 & 15.92  \\
\bottomrule
\end{tabular}}

%% file: Tables/appendix/designbench.tex
\setlength{\tabcolsep}{4pt}
\caption{DesignBench, full metrics for generation and editing.}
\label{tab:designbench-full}
\begin{tabular}{l cccc cccc}
\toprule
& \multicolumn{4}{c}{Generation} & \multicolumn{4}{c}{Editing} \\
\cmidrule(lr){2-5}\cmidrule(lr){6-9}
Model & Judge & CLIP & MAE$\downarrow$ & SSIM & Judge & CMS & MAE$\downarrow$ & SSIM \\
\midrule
\multicolumn{9}{l}{\textit{Open models}} \\
JanusCoderV-7B & 40.37 & 71.34 & 77.05 & 0.5910 & 7.44 & 0.2234 & 30.11 & 0.8399 \\
JanusCoderV-8B & 34.30 & 68.38 & 83.64 & 0.5822 & 7.26 & 0.2346 & 28.57 & 0.8446 \\
VinciCoder-7B & 43.24 & 72.81 & 68.05 & 0.6137 & 1.40 & 0.1686 & 45.33 & 0.8119 \\
VinciCoder-8B & 48.09 & 76.79 & 61.86 & 0.6330 & 1.71 & 0.1514 & 41.83 & 0.8199 \\
\rowcolor{boxblue} \textbf{\modelnine} & \textbf{69.95} & \textbf{84.77} & \underline{42.07} & \underline{0.6956} & \textbf{8.93} & \textbf{0.2791} & \textbf{30.56} & \underline{0.8373} \\
\rowcolor{boxblue} \textbf{\modelthirtyfive} & \underline{68.98} & \underline{84.56} & \textbf{41.97} & \textbf{0.6972} & \underline{8.89} & \underline{0.2601} & \underline{30.67} & \textbf{0.8398} \\
\midrule
\multicolumn{9}{l}{\textit{Proprietary}} \\
\rowcolor[gray]{0.92} \gptfouro & 52.64 & 76.45 & 64.11 & 0.6259 & 8.89 & 0.3567 & 27.97 & 0.8439 \\
\rowcolor[gray]{0.92} Gemini 3.1 Pro & 69.47 & 85.23 & 44.11 & 0.6914 & 9.57 & 0.3751 & 26.13 & 0.8548 \\
\bottomrule
\end{tabular}

%% file: Tables/appendix/design2code.tex
\setlength{\tabcolsep}{3pt}
\caption{Design2Code, full metrics.}
\label{tab:design2code}
\resizebox{\linewidth}{!}{%
\begin{tabular}{l ccccccc}
\toprule
Model & Block & Text & Pos. & Color & CLIP & Avg & \makecell{LLM\\Score} \\
\midrule
\multicolumn{8}{l}{\textit{Open models}} \\
JanusCoderV-7B & 63.88 & 85.40 & 62.74 & 59.11 & 82.94 & 70.81 & 36.49 \\
JanusCoderV-8B & 59.11 & 81.72 & 61.45 & 57.63 & 83.13 & 68.61 & 34.53 \\
VinciCoder-7B & 71.01 & 87.30 & 68.04 & 71.28 & 85.58 & 76.64 & 42.30 \\
VinciCoder-8B & 79.41 & 91.79 & 73.90 & 77.00 & 87.37 & 81.89 & 45.72 \\
\rowcolor{boxblue} \textbf{\modelnine} & \textbf{90.09} & \textbf{95.95} & \textbf{89.30} & \underline{90.00} & \textbf{92.37} & \textbf{91.54} & \textbf{70.51} \\
\rowcolor{boxblue} \textbf{\modelthirtyfive} & \underline{89.45} & \underline{95.47} & \underline{89.14} & \textbf{90.05} & \underline{91.98} & \underline{91.22} & \underline{67.55} \\
\midrule
\multicolumn{8}{l}{\textit{Proprietary}} \\
\rowcolor[gray]{0.92} \gptfouro & 89.50 & 96.91 & 83.67 & 82.27 & 88.80 & 88.23 & 58.48 \\
\rowcolor[gray]{0.92} Gemini 3.1 Pro & 90.55 & 95.21 & 88.58 & 87.69 & 91.46 & 90.70 & 70.90 \\
\bottomrule
\end{tabular}
}

%% file: Tables/appendix/artifactsbench.tex
\setlength{\tabcolsep}{3pt}
\caption{ArtifactsBench, overall and per-domain scores. }
\label{tab:artifactsbench-full}
\begin{tabular}{l c ccccc}
\toprule
Model & Overall
& \makecell{Game\\Dev.} & \makecell{SVG\\Gen.} & \makecell{Web\\Apps} & \makecell{Simu-\\lation} & \makecell{Mgmt.\\Sys.} \\
\midrule
\multicolumn{7}{l}{\textit{Open models}} \\
JanusCoderV-7B & 31.16 & 27.95 & 28.97 & 33.32 & 34.04 & 33.64 \\
VinciCoder-7B  & 11.35 & 10.51 & 12.04 & 12.48 & 11.48 & 11.29 \\
VinciCoder-8B  & 11.97 & 11.06 & 14.94 & 12.34 & 11.27 & 12.10 \\
VisCoder2  & 17.51 & 16.49 & 20.43 & 18.46 & 18.13 & 15.98 \\
\rowcolor{boxblue} \textbf{\modelnine} & 45.28 & 44.49 & 50.95 & 46.22 & 48.23 & 42.27 \\
\rowcolor{boxblue} \textbf{\modelthirtyfive} & 46.01 & 54.33 & 45.00 & 44.51 & 58.53 & 37.09 \\
\midrule
\multicolumn{7}{l}{\textit{Proprietary}} \\
\rowcolor[gray]{0.92} \gptfouro & 33.54 & 36.96 & 39.54 & 39.27 & 35.73 & 35.83 \\
\bottomrule
\end{tabular}

%% file: Tables/appendix/interactscience.tex
\setlength{\tabcolsep}{4pt}
\caption{InteractScience, full metrics for both lenses. }
\label{tab:interactscience}
\resizebox{\linewidth}{!}{%
\begin{tabular}{l ccc c}
\toprule
& \multicolumn{3}{c}{PFT} & VQT \\
\cmidrule(lr){2-4}\cmidrule(lr){5-5}
Model & Overall & Avg. & Perfect & VLM \\
\midrule
JanusCoderV-7B & 17.73\% & 16.91\% & 4.20\% & 27.67 \\
JanusCoderV-8B & 17.60\% & 17.30\% & 4.20\% & 33.32 \\
VinciCoder-7B & 0.40\% & 0.00\% & 0.00\% & 0.00 \\
VinciCoder-8B & 0.40\% & 0.42\% & 0.00\% & 2.33 \\
\rowcolor{boxblue} \textbf{\modelnine}       & 24.80\% & 22.47\% & 7.69\% & 51.87 \\
\rowcolor{boxblue} \textbf{\modelthirtyfive} & 28.00\% & 25.94\% & 6.99\% & 53.54 \\
\rowcolor[gray]{0.92} \gptfouro & 31.07\% & 28.59\% & 10.49\% & 46.01 \\
\bottomrule
\end{tabular}}

%% file: Tables/appendix/backbone_gemma.tex
\begin{table}[H]
\centering
\footnotesize
\setlength{\tabcolsep}{5pt}
\caption{VisPlotBench scores of the Gemma arm under the designated judge and a newer
judge of the same family. \textbf{Best} per column within a split.}
\label{tab:backbone-gemma}
\begin{tabular}{ll ccc ccc}
\toprule
& & \multicolumn{3}{c}{Judge: GPT-4.1} & \multicolumn{3}{c}{Judge: GPT-5.5} \\
\cmidrule(lr){3-5}\cmidrule(lr){6-8}
Split & Model & Err\%$\downarrow$ & Vis$\uparrow$ & Task$\uparrow$ & Err\%$\downarrow$ & Vis$\uparrow$ & Task$\uparrow$ \\
\midrule
\multirow{3}{*}{HTML}
& \gemmaicon Gemma-4 (base) & 5.6 & 46.6 & 78.8 & 5.6 & 45.4 & 64.2 \\
& \quad+ SFT & \textbf{4.6} & 55.6 & 76.2 & \textbf{4.6} & 53.0 & \textbf{77.8} \\
& \quad+ SFT + RL & \textbf{4.6} & \textbf{64.5} & \textbf{84.0} & \textbf{4.6} & \textbf{64.6} & 75.4 \\
\midrule
\multirow{3}{*}{Python}
& \gemmaicon Gemma-4 (base) & \textbf{6.1} & 69.4 & 87.2 & \textbf{6.1} & 66.8 & 85.0 \\
& \quad+ SFT & 7.1 & 73.3 & 89.5 & 7.1 & 71.0 & 87.2 \\
& \quad+ SFT + RL & \textbf{6.1} & \textbf{75.0} & \textbf{91.0} & \textbf{6.1} & \textbf{72.9} & \textbf{87.6} \\
\bottomrule
\end{tabular}
\end{table}

%% file: Tables/appendix/human_study.tex
\begin{table}[H]
\centering
\footnotesize
\caption{Agreement of judge scores with human ratings on 100 samples.
\textbf{Best} per judge.}
\label{tab:human-study}
\setlength{\tabcolsep}{6pt}
\begin{tabular}{l l c c}
\toprule
Judge & Scoring & Krippendorff's $\alpha$ & Kendall's $\tau_b$ \\
\midrule
\multirow{2}{*}{Claude Sonnet 5}  & Direct       & 0.386 & 0.292 \\
                                  & Rubric-based & \textbf{0.889} & \textbf{0.670} \\
\midrule
\multirow{2}{*}{Gemini 3.5 Flash} & Direct       & 0.303 & 0.282 \\
                                  & Rubric-based & \textbf{0.861} & \textbf{0.643} \\
\midrule
\multirow{2}{*}{GPT-5.5}          & Direct       & 0.514 & 0.442 \\
                                  & Rubric-based & \textbf{0.918} & \textbf{0.657} \\
\bottomrule
\end{tabular}
\end{table}

%% file: main.bbl
\begin{thebibliography}{69}
\providecommand{\natexlab}[1]{#1}
\providecommand{\url}[1]{\texttt{#1}}
\expandafter\ifx\csname urlstyle\endcsname\relax
  \providecommand{\doi}[1]{doi: #1}\else
  \providecommand{\doi}{doi: \begingroup \urlstyle{rm}\Url}\fi

\bibitem[{Anthropic}(2026{\natexlab{a}})]{anthropic2026opus46}
{Anthropic}.
\newblock Introducing {Claude} {Opus} 4.6.
\newblock \url{https://www.anthropic.com/news/claude-opus-4-6}, February 2026{\natexlab{a}}.
\newblock Accessed: 2026-06-29.

\bibitem[{Anthropic}(2026{\natexlab{b}})]{anthropic2026sonnet5}
{Anthropic}.
\newblock Introducing {Claude} {Sonnet} 5.
\newblock \url{https://www.anthropic.com/news/claude-sonnet-5}, June 2026{\natexlab{b}}.
\newblock Accessed: 2026-09-23.

\bibitem[Bai et~al.(2025)Bai, Cai, Chen, Chen, Chen, Cheng, Deng, Ding, Gao, Ge, et~al.]{bai2025qwen3}
Shuai Bai, Yuxuan Cai, Ruizhe Chen, Keqin Chen, Xionghui Chen, Zesen Cheng, Lianghao Deng, Wei Ding, Chang Gao, Chunjiang Ge, et~al.
\newblock Qwen3-vl technical report.
\newblock \emph{arXiv preprint arXiv:2511.21631}, 2025.

\bibitem[Chen et~al.(2025{\natexlab{a}})Chen, Liu, Li, Chen, Guo, Cheng, and Yuan]{chen2025interact}
Qiaosheng Chen, Yang Liu, Lei Li, Kai Chen, Qipeng Guo, Gong Cheng, and Fei Yuan.
\newblock Interactscience: Programmatic and visually-grounded evaluation of interactive scientific demonstration code generation, 2025{\natexlab{a}}.
\newblock URL \url{https://arxiv.org/abs/2510.09724}.

\bibitem[Chen et~al.(2025{\natexlab{b}})Chen, Zhang, Huang, Qiu, Zhang, Wen, and Liu]{chen2025symbolicgraphics}
Yamei Chen, Haoquan Zhang, Yangyi Huang, Zeju Qiu, Kaipeng Zhang, Yandong Wen, and Weiyang Liu.
\newblock Symbolic graphics programming with large language models, 2025{\natexlab{b}}.
\newblock URL \url{https://arxiv.org/abs/2509.05208}.

\bibitem[Galimzyanov et~al.(2025)Galimzyanov, Titov, Golubev, and Bogomolov]{pandasplotbench2025timur}
Timur Galimzyanov, Sergey Titov, Yaroslav Golubev, and Egor Bogomolov.
\newblock Drawing pandas: A benchmark for llms in generating plotting code.
\newblock In \emph{2025 IEEE/ACM 22nd International Conference on Mining Software Repositories (MSR)}, pp.\  503--507, 2025.
\newblock \doi{10.1109/MSR66628.2025.00083}.

\bibitem[{Gemini Team}(2025)]{gemini2025gemini3}
{Gemini Team}.
\newblock Gemini 3 pro model card, November 2025.
\newblock URL \url{https://deepmind.google/models/model-cards/gemini-3-1-pro/}.

\bibitem[{Gemini Team}(2026)]{gemini2026gemini35flash}
{Gemini Team}.
\newblock Gemini 3.5 flash model card, May 2026.
\newblock URL \url{https://deepmind.google/models/model-cards/gemini-3-5-flash/}.

\bibitem[{Gemma Team}(2026)]{google2026gemma}
{Gemma Team}.
\newblock Gemma 4 technical report.
\newblock \emph{arXiv preprint arXiv:2607.02770}, 2026.

\bibitem[Gui et~al.(2025)Gui, Li, Wan, Shi, Zhang, Su, Chen, Chen, Wu, Zhou, Jiang, Jin, and Zhang]{gui2025webcode2m}
Yi~Gui, Zhen Li, Yao Wan, Yemin Shi, Hongyu Zhang, Yi~Su, Bohua Chen, Dongping Chen, Siyuan Wu, Xing Zhou, Wenbin Jiang, Hai Jin, and Xiangliang Zhang.
\newblock Webcode2m: A real-world dataset for code generation from webpage designs.
\newblock In \emph{THE WEB CONFERENCE 2025}, 2025.
\newblock URL \url{https://openreview.net/forum?id=aeP5nmlw5B}.

\bibitem[Guo et~al.(2025)Guo, Zheng, Li, Bai, Li, Wang, Zhu, Neubig, Chen, and Yue]{guo2025mammothvl}
Jiawei Guo, Tianyu Zheng, Yizhi Li, Yuelin Bai, Bo~Li, Yubo Wang, King Zhu, Graham Neubig, Wenhu Chen, and Xiang Yue.
\newblock {MA}mmo{TH}-{VL}: Eliciting multimodal reasoning with instruction tuning at scale.
\newblock In \emph{Proceedings of the 63rd Annual Meeting of the Association for Computational Linguistics (Volume 1: Long Papers)}, pp.\  13869--13920, Vienna, Austria, July 2025. Association for Computational Linguistics.
\newblock ISBN 979-8-89176-251-0.
\newblock \doi{10.18653/v1/2025.acl-long.680}.
\newblock URL \url{https://aclanthology.org/2025.acl-long.680/}.

\bibitem[Hunter(2007)]{hunter2007matplotlib}
John~D. Hunter.
\newblock Matplotlib: A 2d graphics environment.
\newblock \emph{Computing in Science \& Engineering}, 9\penalty0 (3):\penalty0 90--95, 2007.
\newblock \doi{10.1109/MCSE.2007.55}.

\bibitem[Jiang et~al.(2026)Jiang, Huang, Wu, Li, Chen, Zhang, and Wei]{jiang2026viscodex}
Lingjie Jiang, Shaohan Huang, Xun Wu, Yixia Li, Guanhua Chen, Dongdong Zhang, and Furu Wei.
\newblock Viscodex: Unified multimodal code generation via merging vision and coding models.
\newblock In \emph{The Fourteenth International Conference on Learning Representations}, 2026.
\newblock URL \url{https://openreview.net/forum?id=hUXzPauNEM}.

\bibitem[Jiang et~al.(2025)Jiang, Zheng, Wan, Han, Wang, Lyu, and Yue]{ScreenCoder}
Yilei Jiang, Yaozhi Zheng, Yuxuan Wan, Jiaming Han, Qunzhong Wang, Michael~R. Lyu, and Xiangyu Yue.
\newblock Screencoder: Advancing visual-to-code generation for front-end automation via modular multimodal agents.
\newblock \emph{CoRR}, abs/2507.22827, 2025.
\newblock \doi{10.48550/ARXIV.2507.22827}.
\newblock URL \url{https://doi.org/10.48550/arXiv.2507.22827}.

\bibitem[Kim et~al.(2025)Kim, Chen, Cheng, Gindulyte, He, He, Li, Shoemaker, Thiessen, Yu, Zaslavsky, Zhang, and Bolton]{kim2025pubchem}
S.~Kim, J.~Chen, T.~Cheng, A.~Gindulyte, J.~He, S.~He, Q.~Li, B.~A. Shoemaker, P.~A. Thiessen, B.~Yu, L.~Zaslavsky, J.~Zhang, and E.~E. Bolton.
\newblock Pubchem 2025 update.
\newblock \emph{Nucleic Acids Research}, 53\penalty0 (D1):\penalty0 D1516--D1525, Jan 2025.
\newblock \doi{10.1093/nar/gkae1059}.

\bibitem[Ku et~al.(2025)Ku, Chong, Leung, Shah, Yu, and Chen]{ku2025theoremexplainagent}
Max Ku, Cheuk~Hei Chong, Jonathan Leung, Krish Shah, Alvin Yu, and Wenhu Chen.
\newblock {T}heorem{E}xplain{A}gent: Towards video-based multimodal explanations for {LLM} theorem understanding.
\newblock In \emph{Proceedings of the 63rd Annual Meeting of the Association for Computational Linguistics (Volume 1: Long Papers)}, pp.\  6663--6684, Vienna, Austria, July 2025. Association for Computational Linguistics.
\newblock ISBN 979-8-89176-251-0.
\newblock \doi{10.18653/v1/2025.acl-long.332}.
\newblock URL \url{https://aclanthology.org/2025.acl-long.332/}.

\bibitem[Lauren{\c{c}}on et~al.(2024)Lauren{\c{c}}on, Tronchon, and Sanh]{laurenccon2024websight}
Hugo Lauren{\c{c}}on, L{\'e}o Tronchon, and Victor Sanh.
\newblock Unlocking the conversion of web screenshots into html code with the websight dataset.
\newblock \emph{arXiv preprint arXiv:2403.09029}, 2024.

\bibitem[Li et~al.(2025)Li, Yu, Wei, Dong, Lin, Yang, Wang, and Hao]{li2025unisvg}
Jinke Li, Jiarui Yu, Chenxing Wei, Hande Dong, Qiang Lin, Liangjing Yang, Zhicai Wang, and Yanbin Hao.
\newblock {UniSVG}: A unified dataset for vector graphic understanding and generation with multimodal large language models.
\newblock In \emph{Proceedings of the 33rd ACM International Conference on Multimedia}, pp.\  13156--13163, 2025.

\bibitem[Li et~al.(2024)Li, Tian, Hu, Luo, Huang, and Ma]{li2024mmcode}
Kaixin Li, Yuchen Tian, Qisheng Hu, Ziyang Luo, Zhiyong Huang, and Jing Ma.
\newblock {MMC}ode: Benchmarking multimodal large language models for code generation with visually rich programming problems.
\newblock In \emph{Findings of the Association for Computational Linguistics: EMNLP 2024}, pp.\  736--783, Miami, Florida, USA, November 2024. Association for Computational Linguistics.
\newblock \doi{10.18653/v1/2024.findings-emnlp.42}.
\newblock URL \url{https://aclanthology.org/2024.findings-emnlp.42/}.

\bibitem[Liu et~al.(2025)Liu, Zeng, Zhang, Wang, Shan, and He]{liu2025perceptionbottleneckvlmschart}
Junteng Liu, Weihao Zeng, Xiwen Zhang, Yijun Wang, Zifei Shan, and Junxian He.
\newblock On the perception bottleneck of vlms for chart understanding, 2025.
\newblock URL \url{https://arxiv.org/abs/2503.18435}.

\bibitem[Liu et~al.(2026)Liu, Ding, Fang, Dai, Yang, Liang, Wang, Chen, Lin, and Zang]{liu2026visual}
Ziyu Liu, Shengyuan Ding, Xinyu Fang, Xuanlang Dai, Penghui Yang, Jianze Liang, Jiaqi Wang, Kai Chen, Dahua Lin, and Yuhang Zang.
\newblock Visual-erm: Reward modeling for visual equivalence.
\newblock \emph{arXiv preprint arXiv:2603.13224}, 2026.

\bibitem[Luo et~al.(2025{\natexlab{a}})Luo, Yuan, Chen, Cai, Yue, Yang, Daha, Li, and Lian]{luo2026mmmg}
Yuxuan Luo, Ryan Yuan, Junwen Chen, Haonan Cai, Ziyi Yue, Yuwei Yang, Fatima~Zohra Daha, Ji~Li, and Zhouhui Lian.
\newblock Mmmg: A massive, multidisciplinary, multi-tier generation benchmark for text-to-image reasoning.
\newblock \emph{Advances in Neural Information Processing Systems}, 38, 2025{\natexlab{a}}.

\bibitem[Luo et~al.(2024)Luo, Xu, Zhao, Sun, Geng, Hu, Tao, Ma, Lin, and Jiang]{luo2024wizard}
Ziyang Luo, Can Xu, Pu~Zhao, Qingfeng Sun, Xiubo Geng, Wenxiang Hu, Chongyang Tao, Jing Ma, Qingwei Lin, and Daxin Jiang.
\newblock Wizardcoder: Empowering code large language models with evol-instruct.
\newblock In B.~Kim, Y.~Yue, S.~Chaudhuri, K.~Fragkiadaki, M.~Khan, and Y.~Sun (eds.), \emph{International Conference on Learning Representations}, volume 2024, pp.\  27168--27188, 2024.
\newblock URL \url{https://proceedings.iclr.cc/paper_files/paper/2024/file/72eba29737f9c3a5a4ce8cdb7b667145-Paper-Conference.pdf}.

\bibitem[Luo et~al.(2025{\natexlab{b}})Luo, Li, Lin, Tian, Kankanhalli, and Ma]{luo2025tree}
Ziyang Luo, Kaixin Li, Hongzhan Lin, Yuchen Tian, Mohan Kankanhalli, and Jing Ma.
\newblock Tree-of-evolution: Tree-structured instruction evolution for code generation in large language models.
\newblock In \emph{Proceedings of the 63rd Annual Meeting of the Association for Computational Linguistics (Volume 1: Long Papers)}, pp.\  297--316, Vienna, Austria, July 2025{\natexlab{b}}. Association for Computational Linguistics.
\newblock ISBN 979-8-89176-251-0.
\newblock \doi{10.18653/v1/2025.acl-long.14}.
\newblock URL \url{https://aclanthology.org/2025.acl-long.14/}.

\bibitem[Ma et~al.(2025)Ma, Zhang, Zhao, Song, Wang, Sui, and Luo]{ma2025stabilizing}
Wenhan Ma, Hailin Zhang, Liang Zhao, Yifan Song, Yudong Wang, Zhifang Sui, and Fuli Luo.
\newblock Stabilizing moe reinforcement learning by aligning training and inference routers.
\newblock \emph{arXiv preprint arXiv:2510.11370}, 2025.

\bibitem[Masry et~al.(2022)Masry, Long, Tan, Joty, and Hoque]{masry2022chartqa}
Ahmed Masry, Do~Xuan Long, Jia~Qing Tan, Shafiq Joty, and Enamul Hoque.
\newblock {C}hart{QA}: A benchmark for question answering about charts with visual and logical reasoning.
\newblock In \emph{Findings of the Association for Computational Linguistics: ACL 2022}, pp.\  2263--2279, Dublin, Ireland, May 2022. Association for Computational Linguistics.
\newblock \doi{10.18653/v1/2022.findings-acl.177}.
\newblock URL \url{https://aclanthology.org/2022.findings-acl.177/}.

\bibitem[Ni et~al.(2025{\natexlab{a}})Ni, Yang, Zhang, Li, Lin, Lin, Wang, Wang, Liu, Zhang, et~al.]{ni2025proimage}
Minheng Ni, Zhengyuan Yang, Yaowen Zhang, Linjie Li, Chung-Ching Lin, Kevin Lin, Zhendong Wang, Xiaofei Wang, Shujie Liu, Lei Zhang, et~al.
\newblock Proimage-bench: Rubric-based evaluation for professional image generation.
\newblock \emph{arXiv preprint arXiv:2512.12220}, 2025{\natexlab{a}}.

\bibitem[Ni et~al.(2025{\natexlab{b}})Ni, Nie, Zou, Yue, and Chen]{ni2025viscoder}
Yuansheng Ni, Ping Nie, Kai Zou, Xiang Yue, and Wenhu Chen.
\newblock Viscoder: Fine-tuning llms for executable python visualization code generation, 2025{\natexlab{b}}.
\newblock URL \url{https://arxiv.org/abs/2506.03930}.

\bibitem[Ni et~al.(2026)Ni, Cai, Chen, Liang, Lyu, Deng, Zou, Nie, Yuan, Yue, and Chen]{ni2026viscoder}
Yuansheng Ni, Songcheng Cai, Xiangchao Chen, Jiarong Liang, Zhiheng Lyu, Jiaqi Deng, Kai Zou, Ping Nie, Fei Yuan, Xiang Yue, and Wenhu Chen.
\newblock Viscoder2: Building multi-language visualization coding agents.
\newblock In \emph{The Fourteenth International Conference on Learning Representations}, 2026.
\newblock URL \url{https://openreview.net/forum?id=4zoMnmZzh4}.

\bibitem[Nishina \& Matsui(2024)Nishina and Matsui]{nishina2024svgeditbench}
Kunato Nishina and Yusuke Matsui.
\newblock Svgeditbench: A benchmark dataset for quantitative assessment of llm's svg editing capabilities, 2024.
\newblock URL \url{https://arxiv.org/abs/2404.13710}.

\bibitem[{OpenAI}(2025)]{openai2025gpt41}
{OpenAI}.
\newblock Introducing {GPT-4.1} in the {API}.
\newblock \url{https://openai.com/index/gpt-4-1/}, April 2025.
\newblock Accessed: 2026-09-22.

\bibitem[{OpenAI}(2026)]{openai2026gpt55}
{OpenAI}.
\newblock {GPT-5.5} system card.
\newblock \url{https://openai.com/index/gpt-5-5-system-card/}, April 2026.
\newblock Accessed: 2026-09-22.

\bibitem[Qiu et~al.(2025)Qiu, Liu, Feng, Liu, Xiao, Collins, Tenenbaum, Weller, Black, and Sch{\"o}lkopf]{qiu2025can}
Zeju Qiu, Weiyang Liu, Haiwen Feng, Zhen Liu, Tim~Z. Xiao, Katherine~M. Collins, Joshua~B. Tenenbaum, Adrian Weller, Michael~J. Black, and Bernhard Sch{\"o}lkopf.
\newblock Can large language models understand symbolic graphics programs?
\newblock In \emph{The Thirteenth International Conference on Learning Representations}, 2025.
\newblock URL \url{https://openreview.net/forum?id=Yk87CwhBDx}.

\bibitem[{Qwen Team}(2026)]{qwen3_5}
{Qwen Team}.
\newblock {Qwen3.5}: Towards native multimodal agents, February 2026.
\newblock URL \url{https://qwen.ai/blog?id=qwen3.5}.

\bibitem[Shao et~al.(2024)Shao, Wang, Zhu, Xu, Song, Bi, Zhang, Zhang, Li, Wu, et~al.]{shao2024deepseekmath}
Zhihong Shao, Peiyi Wang, Qihao Zhu, Runxin Xu, Junxiao Song, Xiao Bi, Haowei Zhang, Mingchuan Zhang, YK~Li, Yang Wu, et~al.
\newblock Deepseekmath: Pushing the limits of mathematical reasoning in open language models.
\newblock \emph{arXiv preprint arXiv:2402.03300}, 2024.

\bibitem[Si et~al.(2025)Si, Zhang, Li, Yang, Liu, and Yang]{si2025design2code}
Chenglei Si, Yanzhe Zhang, Ryan Li, Zhengyuan Yang, Ruibo Liu, and Diyi Yang.
\newblock {D}esign2{C}ode: Benchmarking multimodal code generation for automated front-end engineering.
\newblock In \emph{Proceedings of the 2025 Conference of the Nations of the Americas Chapter of the Association for Computational Linguistics: Human Language Technologies (Volume 1: Long Papers)}, pp.\  3956--3974, Albuquerque, New Mexico, April 2025. Association for Computational Linguistics.
\newblock ISBN 979-8-89176-189-6.
\newblock \doi{10.18653/v1/2025.naacl-long.199}.
\newblock URL \url{https://aclanthology.org/2025.naacl-long.199/}.

\bibitem[Sun et~al.(2025{\natexlab{a}})Sun, Wang, Gu, Li, and Cheng]{sun2025fullfront}
Haoyu Sun, Huichen~Will Wang, Jiawei Gu, Linjie Li, and Yu~Cheng.
\newblock Fullfront: Benchmarking mllms across the full front-end engineering workflow, 2025{\natexlab{a}}.
\newblock URL \url{https://arxiv.org/abs/2505.17399}.

\bibitem[Sun et~al.(2024)Sun, Chen, Xu, Cheng, Ma, Yin, Wang, Han, Zhu, Yuan, et~al.]{sun2024survey}
Qiushi Sun, Zhirui Chen, Fangzhi Xu, Kanzhi Cheng, Chang Ma, Zhangyue Yin, Jianing Wang, Chengcheng Han, Renyu Zhu, Shuai Yuan, et~al.
\newblock A survey of neural code intelligence: Paradigms, advances and beyond.
\newblock \emph{arXiv preprint arXiv:2403.14734}, 2024.

\bibitem[Sun et~al.(2025{\natexlab{b}})Sun, Cheng, Ding, Jin, Wang, Xu, Wu, Jia, Chen, Liu, Kao, Li, He, Qiao, and Wu]{sun2025osgenesis}
Qiushi Sun, Kanzhi Cheng, Zichen Ding, Chuanyang Jin, Yian Wang, Fangzhi Xu, Zhenyu Wu, Chengyou Jia, Liheng Chen, Zhoumianze Liu, Ben Kao, Guohao Li, Junxian He, Yu~Qiao, and Zhiyong Wu.
\newblock {OS}-genesis: Automating {GUI} agent trajectory construction via reverse task synthesis.
\newblock In \emph{Proceedings of the 63rd Annual Meeting of the Association for Computational Linguistics (Volume 1: Long Papers)}, pp.\  5555--5579, Vienna, Austria, July 2025{\natexlab{b}}. Association for Computational Linguistics.
\newblock ISBN 979-8-89176-251-0.
\newblock \doi{10.18653/v1/2025.acl-long.277}.
\newblock URL \url{https://aclanthology.org/2025.acl-long.277/}.

\bibitem[Sun et~al.(2026{\natexlab{a}})Sun, Gong, Li, Guo, and Yuan]{sun2026codeevo}
Qiushi Sun, Jingyang Gong, Lei Li, Qipeng Guo, and Fei Yuan.
\newblock {C}ode{E}vo: Interaction-driven synthesis of code-centric data through hybrid and iterative feedback.
\newblock In \emph{Proceedings of the 64th Annual Meeting of the {A}ssociation for {C}omputational {L}inguistics (Volume 1: Long Papers)}, pp.\  9668--9687, San Diego, California, United States, July 2026{\natexlab{a}}. Association for Computational Linguistics.
\newblock ISBN 979-8-89176-390-6.
\newblock URL \url{https://aclanthology.org/2026.acl-long.438/}.

\bibitem[Sun et~al.(2026{\natexlab{b}})Sun, Gong, Liu, Chen, Li, Chen, Guo, Kao, and Yuan]{sun2026januscoder}
Qiushi Sun, Jingyang Gong, Yang Liu, Qiaosheng Chen, Lei Li, Kai Chen, Qipeng Guo, Ben Kao, and Fei Yuan.
\newblock Januscoder: Towards a foundational visual-programmatic interface for code intelligence.
\newblock In \emph{The Fourteenth International Conference on Learning Representations}, 2026{\natexlab{b}}.
\newblock URL \url{https://openreview.net/forum?id=N4BB09TXad}.

\bibitem[Wang et~al.(2025)Wang, Zhou, Xu, Cheng, Zuo, Tian, Song, Lu, Hu, and Liu]{wang2025codevision}
Hanbin Wang, Xiaoxuan Zhou, Zhipeng Xu, Keyuan Cheng, Yuxin Zuo, Kai Tian, Jingwei Song, Junting Lu, Wenhui Hu, and Xueyang Liu.
\newblock Code-vision: Evaluating multimodal llms logic understanding and code generation capabilities, 2025.
\newblock URL \url{https://arxiv.org/abs/2502.11829}.

\bibitem[Wang et~al.(2024)Wang, Xia, He, Chen, Liu, Zhu, Liang, Wu, Liu, Malladi, Chevalier, Arora, and Chen]{wang2024charxiv}
Zirui Wang, Mengzhou Xia, Luxi He, Howard Chen, Yitao Liu, Richard Zhu, Kaiqu Liang, Xindi Wu, Haotian Liu, Sadhika Malladi, Alexis Chevalier, Sanjeev Arora, and Danqi Chen.
\newblock Charxiv: Charting gaps in realistic chart understanding in multimodal {LLM}s.
\newblock In \emph{The Thirty-eight Conference on Neural Information Processing Systems Datasets and Benchmarks Track}, 2024.
\newblock URL \url{https://openreview.net/forum?id=cy8mq7QYae}.

\bibitem[Wei et~al.(2024)Wei, Wang, Liu, Ding, and Zhang]{wei2024magicoder}
Yuxiang Wei, Zhe Wang, Jiawei Liu, Yifeng Ding, and Lingming Zhang.
\newblock Magicoder: empowering code generation with oss-instruct.
\newblock In \emph{Proceedings of the 41st International Conference on Machine Learning}, ICML'24. JMLR.org, 2024.

\bibitem[Wu et~al.(2025)Wu, Liang, Ge, Guo, Lu, Wang, Shan, and Luo]{wu2025plot2code}
Chengyue Wu, Zhixuan Liang, Yixiao Ge, Qiushan Guo, Zeyu Lu, Jiahao Wang, Ying Shan, and Ping Luo.
\newblock {P}lot2{C}ode: A comprehensive benchmark for evaluating multi-modal large language models in code generation from scientific plots.
\newblock In \emph{Findings of the Association for Computational Linguistics: NAACL 2025}, pp.\  3006--3028, Albuquerque, New Mexico, April 2025. Association for Computational Linguistics.
\newblock ISBN 979-8-89176-195-7.
\newblock \doi{10.18653/v1/2025.findings-naacl.164}.
\newblock URL \url{https://aclanthology.org/2025.findings-naacl.164/}.

\bibitem[Xia et~al.(2025)Xia, Zhang, Ye, Yan, Liu, Zhou, Chen, Ye, Dou, Shi, Yan, and Qiao]{xia2025chartx}
Renqiu Xia, Bo~Zhang, Hancheng Ye, Xiangchao Yan, Qi~Liu, Hongbin Zhou, Zijun Chen, Peng Ye, Min Dou, Botian Shi, Junchi Yan, and Yu~Qiao.
\newblock Chartx \& chartvlm: A versatile benchmark and foundation model for complicated chart reasoning, 2025.
\newblock URL \url{https://arxiv.org/abs/2402.12185}.

\bibitem[Xiao et~al.(2025)Xiao, Wang, Lam, Wan, Liu, Huo, and Lyu]{xiao2025designbench}
Jingyu Xiao, Ming Wang, Man~Ho Lam, Yuxuan Wan, Junliang Liu, Yintong Huo, and Michael~R. Lyu.
\newblock Designbench: A comprehensive benchmark for mllm-based front-end code generation, 2025.
\newblock URL \url{https://arxiv.org/abs/2506.06251}.

\bibitem[Xu et~al.(2024)Xu, Wu, Sun, Ren, Yuan, Yuan, Lin, Qiao, and Liu]{xu2024symbol}
Fangzhi Xu, Zhiyong Wu, Qiushi Sun, Siyu Ren, Fei Yuan, Shuai Yuan, Qika Lin, Yu~Qiao, and Jun Liu.
\newblock Symbol-{LLM}: Towards foundational symbol-centric interface for large language models.
\newblock In \emph{Proceedings of the 62nd Annual Meeting of the Association for Computational Linguistics (Volume 1: Long Papers)}, pp.\  13091--13116, Bangkok, Thailand, August 2024. Association for Computational Linguistics.
\newblock \doi{10.18653/v1/2024.acl-long.707}.
\newblock URL \url{https://aclanthology.org/2024.acl-long.707/}.

\bibitem[Xu et~al.(2025{\natexlab{a}})Xu, Sun, Cheng, Liu, Qiao, and Wu]{xu2025interactive}
Fangzhi Xu, Qiushi Sun, Kanzhi Cheng, Jun Liu, Yu~Qiao, and Zhiyong Wu.
\newblock Interactive evolution: A neural-symbolic self-training framework for large language models.
\newblock In \emph{Proceedings of the 63rd Annual Meeting of the Association for Computational Linguistics (Volume 1: Long Papers)}, pp.\  12975--12993, Vienna, Austria, July 2025{\natexlab{a}}. Association for Computational Linguistics.
\newblock ISBN 979-8-89176-251-0.
\newblock \doi{10.18653/v1/2025.acl-long.635}.
\newblock URL \url{https://aclanthology.org/2025.acl-long.635/}.

\bibitem[Xu et~al.(2025{\natexlab{b}})Xu, Liu, Yin, Zhou, and Poovendran]{xu2025kodcode}
Zhangchen Xu, Yang Liu, Yueqin Yin, Mingyuan Zhou, and Radha Poovendran.
\newblock {K}od{C}ode: A diverse, challenging, and verifiable synthetic dataset for coding.
\newblock In Wanxiang Che, Joyce Nabende, Ekaterina Shutova, and Mohammad~Taher Pilehvar (eds.), \emph{Findings of the Association for Computational Linguistics: ACL 2025}, pp.\  6980--7008, Vienna, Austria, July 2025{\natexlab{b}}. Association for Computational Linguistics.
\newblock ISBN 979-8-89176-256-5.
\newblock \doi{10.18653/v1/2025.findings-acl.365}.
\newblock URL \url{https://aclanthology.org/2025.findings-acl.365/}.

\bibitem[Yang et~al.(2025{\natexlab{a}})Yang, Shi, Liu, Shui, Wang, Jing, XU, Zhu, Li, Zhang, Liu, Nie, Cai, and Yang]{yang2025chartmimic}
Cheng Yang, Chufan Shi, Yaxin Liu, Bo~Shui, Junjie Wang, Mohan Jing, Linran XU, Xinyu Zhu, Siheng Li, Yuxiang Zhang, Gongye Liu, Xiaomei Nie, Deng Cai, and Yujiu Yang.
\newblock Chartmimic: Evaluating {LMM}'s cross-modal reasoning capability via chart-to-code generation.
\newblock In \emph{The Thirteenth International Conference on Learning Representations}, 2025{\natexlab{a}}.
\newblock URL \url{https://openreview.net/forum?id=sGpCzsfd1K}.

\bibitem[Yang et~al.(2025{\natexlab{b}})Yang, Cheng, Chen, Zeng, Zhang, Wang, Yu, Ma, and Jiang]{yang2025omnisvg}
Yiying Yang, Wei Cheng, Sijin Chen, Xianfang Zeng, Jiaxu Zhang, Liao Wang, Gang Yu, Xinjun Ma, and Yu-Gang Jiang.
\newblock Omnisvg: A unified scalable vector graphics generation model.
\newblock \emph{arXiv preprint arxiv:2504.06263}, 2025{\natexlab{b}}.

\bibitem[Yang et~al.(2024)Yang, Zhou, Wang, Cong, Han, Yan, Liu, Tan, Liu, Yu, Liu, Shi, and Sun]{yang2024matplotagent}
Zhiyu Yang, Zihan Zhou, Shuo Wang, Xin Cong, Xu~Han, Yukun Yan, Zhenghao Liu, Zhixing Tan, Pengyuan Liu, Dong Yu, Zhiyuan Liu, Xiaodong Shi, and Maosong Sun.
\newblock {M}at{P}lot{A}gent: Method and evaluation for {LLM}-based agentic scientific data visualization.
\newblock In \emph{Findings of the Association for Computational Linguistics: ACL 2024}, pp.\  11789--11804, Bangkok, Thailand, August 2024. Association for Computational Linguistics.
\newblock \doi{10.18653/v1/2024.findings-acl.701}.
\newblock URL \url{https://aclanthology.org/2024.findings-acl.701/}.

\bibitem[Ye et~al.(2024)Ye, Kim, Kim, Hwang, Kim, Jo, Thorne, Kim, and Seo]{ye2024iclrw-flask}
Seonghyeon Ye, Doyoung Kim, Sungdong Kim, Hyeonbin Hwang, Seungone Kim, Yongrae Jo, James Thorne, Juho Kim, and Minjoon Seo.
\newblock Flask: Fine-grained language model evaluation based on alignment skill sets.
\newblock In B.~Kim, Y.~Yue, S.~Chaudhuri, K.~Fragkiadaki, M.~Khan, and Y.~Sun (eds.), \emph{International Conference on Learning Representations}, volume 2024, pp.\  55361--55414, 2024.
\newblock URL \url{https://proceedings.iclr.cc/paper_files/paper/2024/file/f41b4a6b202adcd8e150a9d4f124d8f6-Paper-Conference.pdf}.

\bibitem[Yun et~al.(2024)Yun, Lin, Thushara, Bhat, Wang, Jiang, Deng, Wang, Tao, Li, Li, Nakov, Baldwin, Liu, Xing, Liang, and Shen]{yun2024webcode}
Sukmin Yun, Haokun Lin, Rusiru Thushara, Mohammad~Qazim Bhat, Yongxin Wang, Zutao Jiang, Mingkai Deng, Jinhong Wang, Tianhua Tao, Junbo Li, Haonan Li, Preslav Nakov, Timothy Baldwin, Zhengzhong Liu, Eric~P. Xing, Xiaodan Liang, and Zhiqiang Shen.
\newblock Web2code: A large-scale webpage-to-code dataset and evaluation framework for multimodal {LLM}s.
\newblock In \emph{The Thirty-eight Conference on Neural Information Processing Systems Datasets and Benchmarks Track}, 2024.
\newblock URL \url{https://openreview.net/forum?id=hFVpqkRRH1}.

\bibitem[Zan et~al.(2023)Zan, Chen, Zhang, Lu, Wu, Guan, Yongji, and Lou]{zan2023large}
Daoguang Zan, Bei Chen, Fengji Zhang, Dianjie Lu, Bingchao Wu, Bei Guan, Wang Yongji, and Jian-Guang Lou.
\newblock Large language models meet {NL}2{C}ode: A survey.
\newblock In \emph{Proceedings of the 61st Annual Meeting of the Association for Computational Linguistics (Volume 1: Long Papers)}, pp.\  7443--7464, Toronto, Canada, July 2023. Association for Computational Linguistics.
\newblock \doi{10.18653/v1/2023.acl-long.411}.
\newblock URL \url{https://aclanthology.org/2023.acl-long.411/}.

\bibitem[Zhang et~al.(2025)Zhang, Li, Xu, Liu, Liu, Hu, Wu, Huang, Li, Yi, Xiong, Zhu, Zhang, Jiang, Zhang, Xu, Zhai, He, Li, Zhao, Zhang, Tan, Guo, Pang, Ruan, Zhang, Wang, Xu, Yin, Zhou, Zhou, and Lian]{zhang2025artifactsbench}
Chenchen Zhang, Yuhang Li, Can Xu, Jiaheng Liu, Ao~Liu, Shihui Hu, Dengpeng Wu, Guanhua Huang, Kejiao Li, Qi~Yi, Ruibin Xiong, Haotian Zhu, Yuanxing Zhang, Yuhao Jiang, Yue Zhang, Zenan Xu, Bohui Zhai, Guoxiang He, Hebin Li, Jie Zhao, Le~Zhang, Lingyun Tan, Pengyu Guo, Xianshu Pang, Yang Ruan, Zhifeng Zhang, Zhonghu Wang, Ziyan Xu, Zuopu Yin, Wiggin Zhou, Chayse Zhou, and Fengzong Lian.
\newblock Artifactsbench: Bridging the visual-interactive gap in llm code generation evaluation, 2025.
\newblock URL \url{https://arxiv.org/abs/2507.04952}.

\bibitem[Zhang et~al.(2023)Zhang, Chen, Zhang, Keung, Liu, Zan, Mao, Lou, and Chen]{zhang2023repocoder}
Fengji Zhang, Bei Chen, Yue Zhang, Jacky Keung, Jin Liu, Daoguang Zan, Yi~Mao, Jian-Guang Lou, and Weizhu Chen.
\newblock {R}epo{C}oder: Repository-level code completion through iterative retrieval and generation.
\newblock In \emph{Proceedings of the 2023 Conference on Empirical Methods in Natural Language Processing}, pp.\  2471--2484, Singapore, December 2023. Association for Computational Linguistics.
\newblock \doi{10.18653/v1/2023.emnlp-main.151}.
\newblock URL \url{https://aclanthology.org/2023.emnlp-main.151/}.

\bibitem[Zhang et~al.(2024{\natexlab{a}})Zhang, Wu, Bai, Lin, Li, Yu, Wang, Chen, and Keung]{zhang2024humanevalv}
Fengji Zhang, Linquan Wu, Huiyu Bai, Guancheng Lin, Xiao Li, Xiao Yu, Yue Wang, Bei Chen, and Jacky Keung.
\newblock Humaneval-v: Benchmarking high-level visual reasoning with complex diagrams in coding tasks.
\newblock \emph{arXiv preprint arXiv:2410.12381}, 2024{\natexlab{a}}.

\bibitem[Zhang et~al.(2024{\natexlab{b}})Zhang, Hu, Xu, Yan, Xu, Jin, Zhang, and Huang]{zhang2024tinychart}
Liang Zhang, Anwen Hu, Haiyang Xu, Ming Yan, Yichen Xu, Qin Jin, Ji~Zhang, and Fei Huang.
\newblock {T}iny{C}hart: Efficient chart understanding with program-of-thoughts learning and visual token merging.
\newblock In \emph{Proceedings of the 2024 Conference on Empirical Methods in Natural Language Processing}, pp.\  1882--1898, Miami, Florida, USA, November 2024{\natexlab{b}}. Association for Computational Linguistics.
\newblock \doi{10.18653/v1/2024.emnlp-main.112}.
\newblock URL \url{https://aclanthology.org/2024.emnlp-main.112/}.

\bibitem[Zhang et~al.(2024{\natexlab{c}})Zhang, Ma, and Vosoughi]{zhang2024academicvis}
Zhehao Zhang, Weicheng Ma, and Soroush Vosoughi.
\newblock Is {GPT}-4{V} (ision) all you need for automating academic data visualization? exploring vision-language models' capability in reproducing academic charts.
\newblock In \emph{Findings of the Association for Computational Linguistics: EMNLP 2024}, pp.\  8271--8288, Miami, Florida, USA, November 2024{\natexlab{c}}. Association for Computational Linguistics.
\newblock \doi{10.18653/v1/2024.findings-emnlp.485}.
\newblock URL \url{https://aclanthology.org/2024.findings-emnlp.485/}.

\bibitem[Zhao et~al.(2025{\natexlab{a}})Zhao, Jiang, Zeng, Chen, Qiu, Huang, Zhong, Zheng, Cao, and Ma]{zhao2025vincicoderunifyingmultimodalcode}
Xuanle Zhao, Deyang Jiang, Zhixiong Zeng, Lei Chen, Haibo Qiu, Jing Huang, Yufeng Zhong, Liming Zheng, Yilin Cao, and Lin Ma.
\newblock Vincicoder: Unifying multimodal code generation via coarse-to-fine visual reinforcement learning, 2025{\natexlab{a}}.
\newblock URL \url{https://arxiv.org/abs/2511.00391}.

\bibitem[Zhao et~al.(2025{\natexlab{b}})Zhao, Liu, Haoyue, Luo, Zeng, Li, Shi, and Chen]{zhao2025chartedit}
Xuanle Zhao, Xuexin Liu, Yang Haoyue, Xianzhen Luo, Fanhu Zeng, Jianling Li, Qi~Shi, and Chi Chen.
\newblock {C}hart{E}dit: How far are {MLLM}s from automating chart analysis? evaluating {MLLM}s' capability via chart editing.
\newblock In Wanxiang Che, Joyce Nabende, Ekaterina Shutova, and Mohammad~Taher Pilehvar (eds.), \emph{Findings of the Association for Computational Linguistics: ACL 2025}, pp.\  3616--3630, Vienna, Austria, July 2025{\natexlab{b}}. Association for Computational Linguistics.
\newblock ISBN 979-8-89176-256-5.
\newblock \doi{10.18653/v1/2025.findings-acl.185}.
\newblock URL \url{https://aclanthology.org/2025.findings-acl.185/}.

\bibitem[Zhao et~al.(2025{\natexlab{c}})Zhao, Luo, Shi, Chen, Wang, Liu, and Sun]{zhao2025chartcoder}
Xuanle Zhao, Xianzhen Luo, Qi~Shi, Chi Chen, Shuo Wang, Zhiyuan Liu, and Maosong Sun.
\newblock {C}hart{C}oder: Advancing multimodal large language model for chart-to-code generation.
\newblock In \emph{Proceedings of the 63rd Annual Meeting of the Association for Computational Linguistics (Volume 1: Long Papers)}, pp.\  7333--7348, Vienna, Austria, July 2025{\natexlab{c}}. Association for Computational Linguistics.
\newblock ISBN 979-8-89176-251-0.
\newblock \doi{10.18653/v1/2025.acl-long.363}.
\newblock URL \url{https://aclanthology.org/2025.acl-long.363/}.

\bibitem[Zhao et~al.(2026)Zhao, Sun, Xiao, Liu, Yang, Chen, Luo, Huang, Zhong, Chen, Fu, Wei, Bi, Jiang, Qiu, Yang, Shi, Hu, and Zeng]{zhao2026beyond}
Xuanle Zhao, Qiushi Sun, Jingyu Xiao, Xuexin Liu, Haoyue Yang, Qiaosheng Chen, Xianzhen Luo, Jing Huang, Yufeng Zhong, Lei Chen, Shuai Fu, Zhenlin Wei, Jinhe Bi, Lei Jiang, Haibo Qiu, Siqi Yang, Peng Shi, Jian Hu, and Zhixiong Zeng.
\newblock Beyond {NL}2code: A structured survey of multimodal code intelligence.
\newblock \emph{Transactions on Machine Learning Research}, 2026.
\newblock ISSN 2835-8856.
\newblock URL \url{https://openreview.net/forum?id=pn2sdu3Vrf}.

\bibitem[Zheng et~al.(2025)Zheng, Liu, Li, Chen, Yu, Gao, Dang, Liu, Men, Yang, Zhou, and Lin]{gspo}
Chujie Zheng, Shixuan Liu, Mingze Li, Xiong-Hui Chen, Bowen Yu, Chang Gao, Kai Dang, Yuqiong Liu, Rui Men, An~Yang, Jingren Zhou, and Junyang Lin.
\newblock Group sequence policy optimization.
\newblock \emph{arXiv preprint arXiv:2507.18071}, 2025.

\bibitem[Zheng et~al.(2026)Zheng, Zhong, Wang, Dai, Liu, Chu, Lv, Torr, and Lin]{zheng2026code2world}
Yuhao Zheng, Li'an Zhong, Yi~Wang, Rui Dai, Kaikui Liu, Xiangxiang Chu, Linyuan Lv, Philip Torr, and Kevin~Qinghong Lin.
\newblock Code2world: A gui world model via renderable code generation.
\newblock \emph{arXiv preprint arXiv:2602.09856}, 2026.

\bibitem[Zhu et~al.(2025)Zhu, Zhang, Zhao, Ding, Liu, Liu, Wang, Liu, and Li]{zhu2025frontendbench}
Hongda Zhu, Yiwen Zhang, Bing Zhao, Jingzhe Ding, Siyao Liu, Tong Liu, Dandan Wang, Yanan Liu, and Zhaojian Li.
\newblock Frontendbench: A benchmark for evaluating llms on front-end development via automatic evaluation, 2025.
\newblock URL \url{https://arxiv.org/abs/2506.13832}.

\bibitem[Zhuo et~al.(2025)Zhuo, Jin, Liu, Jiang, Liu, Gong, Bishnoi, Mishra, Suppa, Ziems, Utpala, Xu, Song, Li, Cao, Liu, Liu, Abdurakhmanova, Yu, Jia, Yao, Hamilton, Shridhar, Vu, Wang, Liu, Wang, Liu, Hui, Risdal, Khaliq, Sood, Xing, Ahmad, Grundy, Lo, Zhu, Du, Scholak, and von Werra]{zhuo2025bigcodearena}
Terry~Yue Zhuo, Xiaolong Jin, Hange Liu, Juyong Jiang, Tianyang Liu, Chen Gong, Bhupesh Bishnoi, Vaisakhi Mishra, Marek Suppa, Noah Ziems, Saiteja Utpala, Ming Xu, Guangyu Song, Kaixin Li, Yuhan Cao, Bo~Liu, Zheng Liu, Sabina Abdurakhmanova, Wenhao Yu, Mengzhao Jia, Jihan Yao, Kenneth Hamilton, Kumar Shridhar, Minh~Chien Vu, Dingmin Wang, Jiawei Liu, Zijian Wang, Qian Liu, Binyuan Hui, Meg Risdal, Ahsen Khaliq, Atin Sood, Zhenchang Xing, Wasi~Uddin Ahmad, John Grundy, David Lo, Banghua Zhu, Xiaoning Du, Torsten Scholak, and Leandro von Werra.
\newblock Bigcodearena: Unveiling more reliable human preferences in code generation via execution, 2025.
\newblock URL \url{https://arxiv.org/abs/2510.08697}.

\end{thebibliography}
